\documentclass[acmtog,authorversion]{acmart}  
\acmSubmissionID{papers\_1519}

\usepackage{booktabs} 
\usepackage{graphicx}
\usepackage{booktabs}
\usepackage{wrapfig}
\usepackage{tikz}
\usetikzlibrary{spy}

\usepackage{multirow}
\usepackage{colortbl}
\usepackage{bm}
\usepackage{multirow}
\usepackage[dvipsnames]{xcolor}

\usepackage{colortbl}
\usepackage{newunicodechar}
\newunicodechar{↔}{\ensuremath{\leftrightarrow}}
\usepackage{amsmath,bm}
\usepackage{microtype}
\usepackage{xspace}
\usepackage[export]{adjustbox}

\usepackage{amsmath}

\DeclareMathOperator*{\argmin}{arg\,min}

\newcommand{\methodname}{Marigold\,\,V2\xspace}

\usepackage{pifont}%
\newcommand{\xmark}{\ding{55}}%
\newcommand{\cmark}{\ding{51}}%

\usepackage{subcaption}
\usepackage{enumitem}

\newcommand{\nocompete}[1]{\textcolor{gray}{#1}}

\definecolor{colorfirst}{rgb}{.866,.945, 0.831}
\definecolor{colorsecond}{rgb}{1, 0.98, 0.83}
\definecolor{colorthird}{rgb}{0.76, 0.87, 0.92}
\definecolor{colorcite}{rgb}{0.212, 0.490, 0.741}
\newcommand{\cellfirst}{\cellcolor{colorfirst}}
\newcommand{\cellsecond}{\cellcolor{colorsecond}}
\newcommand{\cellthird}{\cellcolor{colorthird}}

\usepackage[ruled]{algorithm2e} 

\SetAlFnt{\small}
\SetAlCapFnt{\small}
\SetAlCapNameFnt{\small}
\SetAlCapHSkip{0pt}

\setcopyright{cc}
\setcctype{by}
\acmJournal{TOG}
\acmYear{2026} 
\acmVolume{45} 
\acmNumber{6} 
\acmArticle{204}
\acmMonth{12} 
\acmDOI{10.1145/3842528}

\begin{document}
\title[Marigold V2]{Marigold V2:\\ Revisiting Diffusion Transformers for Monocular Depth Estimation}

\author{Igor Pavlovic}
\authornote{%
    Equal contribution \hfill
    $^\dagger$ Internship at HUAWEI Bayer Lab \hfill
    \textsuperscript{\S}
    Project lead
}
\authornotemark[2]
\orcid{0009-0001-6088-0316}
\affiliation{%
 \institution{EPFL, HUAWEI Bayer Lab}
 \country{Switzerland}
}

\author{Thiemo Wandel}
\authornotemark[1]
\orcid{0009-0003-0692-5653}
\affiliation{%
 \institution{HUAWEI Bayer Lab}
 \country{Switzerland}
}

\author{Anton Obukhov}
\authornotemark[4]
\orcid{0000-0002-5337-583X}
\affiliation{%
 \institution{HUAWEI Bayer Lab}
 \country{Switzerland}
}

\author{Luca Bartolomei}
\orcid{0000-0002-5509-437X}
\affiliation{%
 \institution{University of Bologna}
 \country{Italy}
}

\author{Andrey Davydov}
\orcid{0009-0007-9384-0058}
\affiliation{%
 \institution{HUAWEI Bayer Lab}
 \country{Switzerland}
}

\author{Fabio Tosi}
\orcid{0000-0002-6276-5282}
\affiliation{%
 \institution{University of Bologna}
 \country{Italy}
}

\author{Matteo Poggi}
\orcid{0000-0002-3337-2236}
\affiliation{%
 \institution{University of Bologna}
 \country{Italy}
}

\author{Sabine Süsstrunk}
\orcid{0000-0002-0441-6068}
\affiliation{%
 \institution{EPFL}
 \country{Switzerland}
}

\author{Dengxin Dai}
\orcid{0000-0001-5440-9678}
\affiliation{%
 \institution{HUAWEI Bayer Lab}
 \country{Switzerland}
}

\begin{teaserfigure}
  \includegraphics[trim={0 0 0 0}, clip, width=\linewidth]{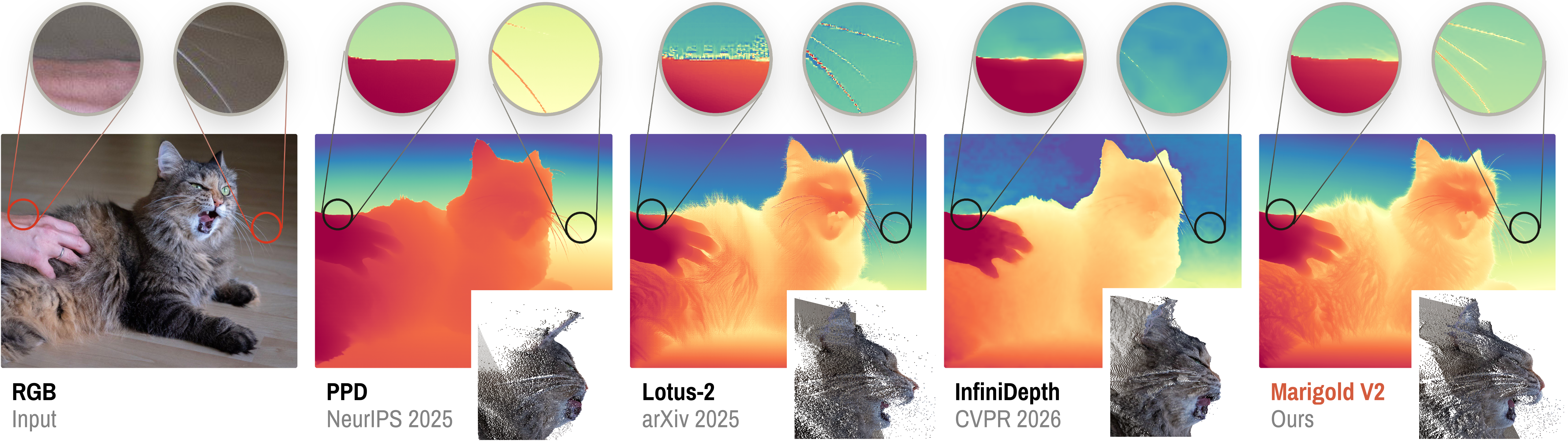}
  \caption{%
\textbf{We present \methodname}, a model and a cost-effective fine-tuning protocol that repurposes an open-source image-editing diffusion transformer (Qwen-\linebreak Image-Edit) into a state-of-the-art monocular depth estimator in less than a week on a single 32GB GPU. 
Our protocol extends Marigold and its DiT follow-ups via novel Sinkhorn and representation alignment losses to sharpen details and improve geometry, while remaining affordable for individual practitioners and small labs.
Beyond quantitative accuracy, \methodname excels qualitatively, as shown above: it faithfully reproduces sharp edges, fur, and hair-thin details, surpassing recent detail-oriented methods such as Pixel-Perfect Depth (PPD) and InfiniDepth, all while retaining the efficiency of a VAE-based model.
  }
  \label{fig:teaser}
  \Description[Marigold V2 depth maps with sharp fine details versus prior methods]{Qualitative comparison of monocular depth estimation. Input photos are shown alongside depth maps from Marigold V2, Pixel-Perfect Depth, and InfiniDepth. Marigold V2 preserves fur, foliage, and hair-thin edges crisply, while the other methods blur or lose them.}
\end{teaserfigure}

\begin{abstract}
Monocular depth estimation is a ubiquitous yet highly ill-posed computer vision task, with downstream applications in scene reconstruction, computational photography, and robotics, among others. 
Despite the field's maturity, recent models still struggle to generalize to out-of-distribution inputs and to produce sharp and detailed depth maps.

In this paper, we revisit Marigold, a set of techniques for repurposing modern image generation and editing models, powered by the diffusion transformer (DiT) architecture, into state-of-the-art monocular depth estimators. 
Our recipes target single-step inference from pretrained multi-step flow-matching models, with quantization where needed, preserving model capacity while remaining cheap to run.
We analyze the artifacts of na\"ive training and identify two effective remedies: aligning the model's internal representations with semantic features extracted from ground-truth, and adopting a 2-stage fine-tuning protocol built around a novel Sinkhorn-based loss. 
The results are crisper, cleaner depth maps that generalize well out-of-distribution, with 16–26\% improvement in AbsRel over the previous best on KITTI and ETH3D. 
Qualitatively, our model resolves fur, foliage, and hair-thin edges that have eluded prior models. 
Furthermore, Marigold V2 achieves state-of-the-art results when applied to other dense regression tasks, such as surface normals estimation and intrinsic image decomposition.
Project website: \href{https://hf.co/spaces/huawei-bayerlab/marigold-v2-web}{https://hf.co/spaces/huawei-bayerlab/marigold-v2-web}.

\end{abstract}

%
%


%
%


\maketitle

\section{Introduction}

Monocular depth estimation, which aims at recovering per-pixel depth from a single image, is a fundamental problem in computer vision and computational photography with far-reaching implications for graphics and visual computing.
Accurate depth maps underpin a broad spectrum of applications central to the graphics community, including image-based rendering and novel view synthesis~\cite{kangle2021dsnerf,safadoust2024BMVC}, bokeh simulation and computational refocusing~\cite{Peng2022BokehMe}, portrait relighting and matting~\cite{yang2021S3Net}, as well as geometry-aware image editing and controllable generation~\cite{zhang2023adding, hu2024animate}.
Beyond 2D effects, monocular depth serves as the entry point for single-image 3D reconstruction and scene lifting~\cite{huang20242d, long2024wonder3d, jiang2026dimer}, enabling downstream tasks such as object insertion, augmented reality compositing, and 3D content creation from casually captured photographs.

The core challenge is one of inherent ambiguity: a single 2D image is consistent with infinitely many 3D scene configurations, and resolving this ambiguity requires reasoning about scene structure, material properties, and lighting that goes far beyond low-level appearance cues.
Early geometry-based methods that rely on multi-view constraints, photometric consistency, or hand-crafted shape priors collapse in the unconstrained single-image setting, while deep learning made it possible to cast the problem as a regression task, guided by appearance features, learned through supervised training over annotated datasets~\cite{eigen2014depth, fu2018deep,yuan2022neural}.  
With the steady increase in the amount of data available for training, more and more accurate models have emerged over the years \cite{yang2024depth, yang2024depth2,wang2025moge,wang2025moge2}, although they are inevitably bound to the depth distribution coverage of such training data. As a consequence, these models suffer significant drops in accuracy in corner cases %
that are underrepresented in the training distribution (e.g., non-Lambertian surfaces or adverse weather conditions). 

In parallel, advances in generative diffusion models~\cite{rombach2022high,bfl2024flux,wu2025qwen} unveiled an alternative paradigm for depth estimation and visual understanding. From the billion-scale data used for training, these models learn a geometrically consistent representation of the world, pivotal in making the generated images more and more realistic. 
Following this intuition, a family of depth estimation approaches, concurrent to the aforementioned \textit{discriminative} models and derived from generative diffusion models, emerged~\cite{ke2024repurposing,ke2025marigold}, showing impressive results %
despite the very limited amount of depth-annotated training data used for this repurposing~\cite{fu2024geowizard, he2025lotus1, he2025lotus2, garcia2025fine, zhaodiception}. %
Nevertheless, several limitations of these dif\-fu\-sion-based models remain unresolved: %
the loss of fine-grained details and oversmoothed boundaries in the predicted depth maps (or \textit{flying pixels} when projected into point clouds), most prominently. %

In this paper, we present \textbf{\methodname}, a model and a cost-effective fine-tuning protocol that repurposes an open-source image-editing diffusion transformer (Qwen-Image-Edit~\cite{wu2025qwen}) into a state-of-the-art monocular depth estimator.
Our approach extends the Marigold family~\cite{ke2024repurposing, ke2025marigold} and its DiT follow-ups to the image-editing paradigm, while introducing a principled solution to the fine-grained detail loss and oversmoothed boundary problem typical of dif\-fu\-sion-based depth estimators that use VAEs. 
Our work introduces three main novel contributions:

\begin{itemize}[leftmargin=*]
    \setlength\itemsep{2pt}
    \item[---] \textbf{Marigold V2 protocol}. A lightweight fine-tuning protocol to convert an image-editing DiT into a monocular depth estimator or other dense modality regressor. Fine-tuning our model requires a single consumer GPU, a modestly-sized dataset, and a few days of training, made possible by 4-bit quantization with QLoRA~\cite{dettmers2023qlora,Zakarin_2026_CVPR}.
    \item[---] \textbf{iREPA-depth.} We revisit representation alignment \cite{singh2025irepa} and apply it unconventionally in \textit{Stage 1} of training, aligning towards semantic features extracted from the \textit{ground-truth depth map} rather than from RGB. This supplies semantic and geometric information, eases convergence, and improves visual quality. 
    \item[---] \textbf{SinkLoss.} We introduce a novel Sinkhorn matching-based objective, coined \textit{SinkLoss}, to improve edge sharpness while preserving semantic details -- including fur, hair, and thin structures, as shown in Figs.~\ref{fig:teaser}, \ref{fig:appendix:qualitative_comparison:a}, and \ref{fig:appendix:qualitative_comparison:b}. 
    The tuning with SinkLoss is \textit{Stage 2} of our training pipeline.
\end{itemize}

Our \methodname model achieves state-of-the-art results in depth estimation on standard benchmarks, outperforming the latest dif\-fu\-sion-based alternatives \cite{he2025lotus2,xu2026pixel} and other competitors~\cite{yu2026infinidepth}.
Furthermore, the \methodname recipe adapts readily to other dense regression tasks: depth completion, see-through depth, surface normal estimation, and intrinsic image decomposition, achieving state-of-the-art results on each and demonstrating its broad applicability in computational photography.

\section{Related Work}

\subsection{Discriminative Monocular Depth Estimation}
Estimating depth from a single image is inherently ill-posed, yet deep learning has established it as a credible alternative to traditional approaches. 
Early works operated in single domains trained with ground truth~\cite{eigen2014depth,fu2018deep,lee2019big,yuan2022neural}, self-supervision~\cite{monodepth17,monodepth2,Poggi_CVPR_2020,zhao2022monovit}, or proxy annotations~\cite{tosi2019learning,zhao2023gasmono}. A second generation achieved zero-shot cross-dataset generalization by learning affine-invariant depth over mixed datasets~\cite{eftekhar2021omnidata,ranftl2020towards,ranftl2021vision}, followed by a third generation built on Vision Transformers~\cite{dosovitskiyimage,oquabdinov2} and million-scale data, advancing affine-invariant~\cite{yang2024depth,yang2024depth2,lin2025depth}, universal metric~\cite{piccinelli2024unidepth,piccinelli2025unik3d,ganesan2026unidac,wang2025moge,wang2025moge2}, and video depth~\cite{piccinelli2026velodepth,video_depth_anything}.

{\emergencystretch=2em 
Despite impressive results, discriminative models such as MoGe~\cite{wang2025moge,wang2025moge2}, $\pi^3$~\cite{wang2025pi}, and Depth Anything~\cite{yang2024depth,yang2024depth2} face two fundamental limitations: ground-truth depth data remains scarce and million-scale, far outpaced by the billion-scale data available to generative models; and sensor noise along with poor handling of non-Lambertian, transparent, or reflective surfaces limits annotation quality, a weakness inherited by trained models.
\par}

\subsection{Generative Priors for Monocular Depth Estimation}

Generative models trained on orders of magnitude more data encapsulate richer world knowledge, spurring interest in repurposing them as dense depth predictors. The approaches fall into three families. The first preserves the multi-step diffusion paradigm~\cite{ke2024repurposing,fu2024geowizard,he2025lotus1,gui2025depthfm}, suffering from high inference latency and high sensitivity to noise initialization. The second trades quality for speed by recasting the backbone as a single-pass feed-forward network~\cite{garcia2025fine,he2025lotus1,ke2025marigold}.
The third departs from pure fine-tuning: some methods feed a coarse modality estimate as auxiliary input~\cite{zhang2024betterdepth,ye2024stablenormal}, while others extend the VAE to broader output modalities~\cite{xu2025matters,Krishnan_2025_ICCV}.

Most of the early frameworks were built on Stable Diffusion's convolutional U-Net~\cite{rombach2022high,ronneberger2015u}, fine-tuned on synthetic datasets with pixel-perfect depth. As the generative community migrated from U-Nets to Diffusion Transformers (DiTs)~\cite{peebles2023scalable} -- through PixArt-$\alpha$~\cite{chen2024pixart}, Stable Diffusion 3~\cite{esser2024scaling}, FLUX~\cite{bfl2024flux}, and Qwen~\cite{wu2025qwen} -- repurposing became costlier due to the higher complexity of DiTs. 
Training from scratch requires tens of GPUs~\cite{Le_2025_CVPR}, and LoRA~\cite{hu2022lora} does not always reduce this burden: DICEPTION~\cite{zhaodiception} needs 96 GPU-days,
and Lotus-2~\cite{he2025lotus2} uses 
8 GPUs. 
Vision Banana~\cite{gabeur2026image} further demonstrates that instruction-tuning can unlock existing geometric understanding in pre-trained generators~\cite{liu2026opensourceimageeditingmodels}. Nevertheless, supervised fine-tuning remains the dominant approach -- and the one we adopt here.

\subsection{Representation Alignment for Depth Estimation}

Regularizing diffusion-based depth estimators with features from pretrained visual encoders has emerged as an effective strategy to bridge the gap between generative and discriminative representations.
REPA~\cite{repa} and iREPA~\cite{singh2025irepa} showed that such a regularization can improve semantic fidelity and training convergence of diffusion models. In monocular depth estimation, this principle has recently been adapted to inject semantic information into dif\-fu\-sion-based depth prediction. DepthMaster~\cite{song2026depthmaster} aligns intermediate U-Net features with DINOv2 representations extracted from the input image, while Pixel-Perfect Depth~\cite{xu2026pixel} incorporates semantic representations from vision foundation models directly into the diffusion process via a Semantics-Prompted DiT. In both cases, alignment targets features extracted from the RGB input. 
In contrast, our iREPA-depth variant draws its alignment target from a frozen DINOv3 encoder applied to the ground-truth depth map rather than the input image, providing a more direct supervisory signal for geometric reconstruction, and without introducing inference-time dependencies. It brings semantic details into depth map estimates, improves visual quality, and facilitates overall training convergence.

\subsection{Handling Depth Quality and Artifacts}

Beyond aggregate accuracy, the practical utility of depth maps depends critically on faithful boundary reconstruction and fine-grained surface detail -- qualities that directly impact novel view synthesis, 3D reconstruction, and computational photography.
Prior work has addressed this in isolation: edge-aware losses~\cite{yang2022monocular}, architectural refinements~\cite{depth-pro}, and diffu\-sion-based post-processing improve sharpness, yet without reliably preserving fine-grained detail. SharpDepth~\cite{pham2025sharpdepth} distills boundary sharpness from generative models into a discriminative backbone, yet predictions remain over-smoothed at depth edges. InfiniDepth~\cite{yu2026infinidepth} enables arbitrary-resolution queries via neural implicit fields. Lotus-2~\cite{he2025lotus2} mitigates detail loss through a predictor-sharpener design, at the cost of multi-step inference. Pixel-Perfect Depth~\cite{xu2026pixel} performs diffusion in pixel space to reduce flying pixel artifacts, but does not recover fine-grained detail.
Critically, no existing approach jointly addresses detail preservation and robust supervision under noisy or ambiguous ground-truth within a single-step VAE-based generative framework -- the two limitations our method is designed to address.

\begin{figure}[!t]
  \centering
    \includegraphics[width=\linewidth]{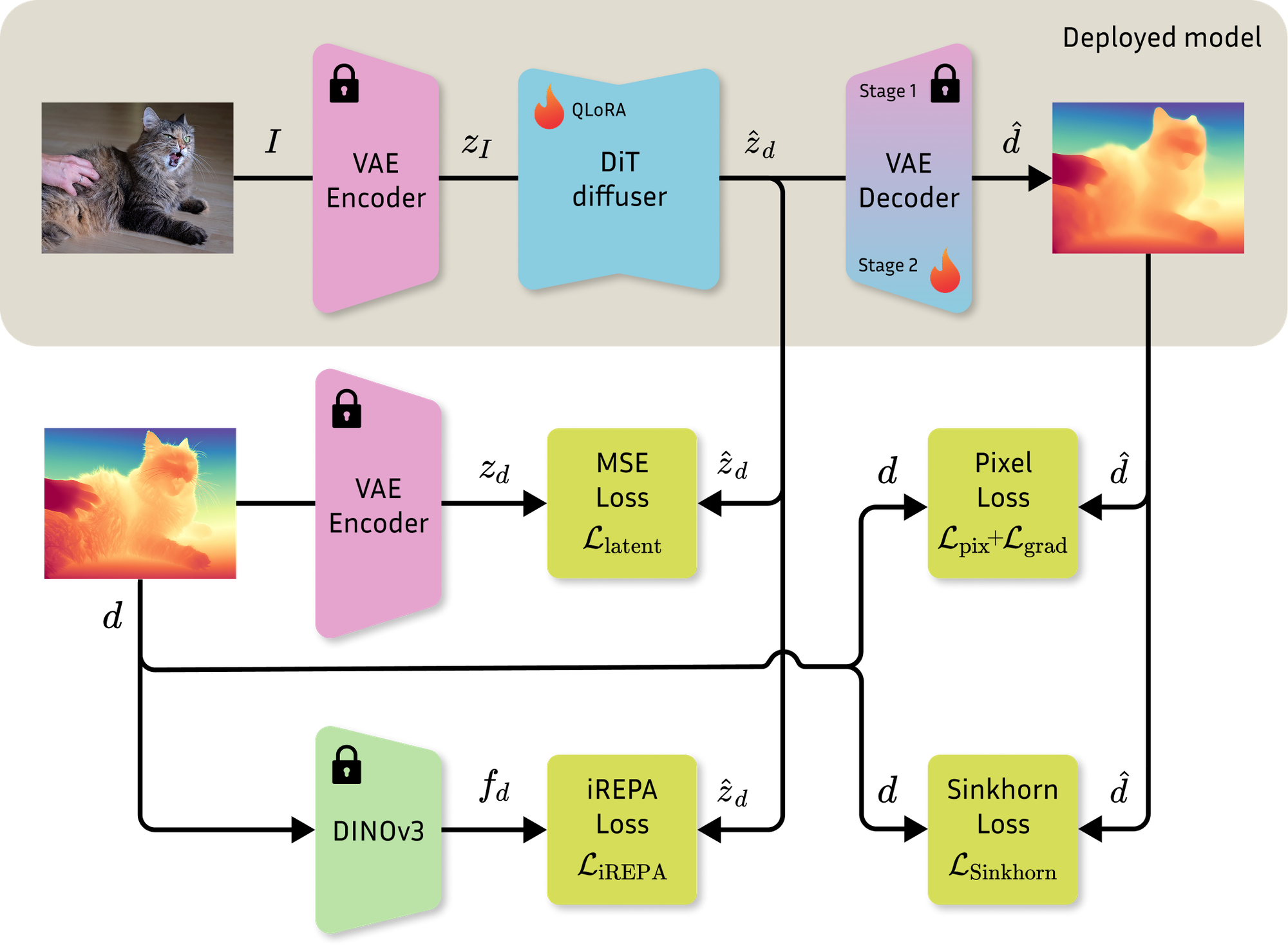}
  \caption{%
\textbf{%
\methodname training protocol.
}
During Stage 1, only QLoRA adapter weights are fine-tuned, regularized by iREPA-depth, which already produces a strong model.
Stage 2 refines it further by adding SinkLoss and unfreezing the VAE decoder. 
At inference, the deployed model requires just one forward pass through the VAE and the DiT.
}
  \label{fig:method}
  \Description[Architectural diagram illustrating our two-stage training protocol]{Architectural diagram illustrating our two-stage training protocol, featuring a frozen VAE Encoder, DiT trained with QLoRA efficient training, and a VAE Decoder, frozen during stage 1 but trained during stage 2. There are multiple loss functions, including Sinkhorn and iREPA.}
\end{figure}

\section{Method}

We propose a dif\-fu\-sion-based monocular depth estimator that adapts a pretrained image-editing diffusion transformer to predict high-quality affine-invariant depth maps from a single RGB image. 
Mari\-gold V1~\cite{ke2024repurposing,ke2025marigold} 
established this direction as an effective alternative to discriminative predictors, showing that pretrained diffusion models carry strong semantic and geometric priors for dense prediction.
Building on this line of work, \methodname repurposes Qwen-Image-Edit-2509~\cite{wu2025qwen} for monocular depth estimation and trains it to directly transform RGB latents into normalized depth latents. Unlike prior dif\-fu\-sion-based depth estimators that mainly rely on latent-space supervision, we additionally introduce pixel-space and semantic feature losses to improve local reconstruction quality and preserve fine geometric details.

Our training follows a two-stage procedure, as depicted in Fig.~\ref{fig:method}. In~\textit{Stage 1}, we adapt the pretrained image-editing DiT to monocular depth estimation using latent rectified-flow supervision together with pixel-space and semantic feature losses. This stage establishes a strong affine-invariant depth predictor with accurate global structure and improved local reconstruction quality. In~\textit{Stage 2}, we further fine-tune the resulting model using the proposed SinkLoss. This second stage is designed to refine fine details and reduce the effect of ambiguous or noisy ground-truth pixels, especially around transparent, thin, or indiscernible structures.

\subsection{Depth Normalization}

Given an RGB image \(I\) and its metric ground-truth depth \(D\), we first convert the target depth into an affine-invariant log-depth representation. Specifically, we convert the metric depth \(D\) into an affine-invariant normalized log-depth target $d$ as:
\begin{equation}
    d
    =
    2
    \left(
    \frac{
        \log(D+\epsilon) - d_{2}
    }{
        d_{98} - d_{2}
    }
    - \frac{1}{2}
    \right),
    \label{eq:log_depth_normalization}
\end{equation}
where \(d_i\) denotes the \(i\)-th percentile of \(\log(D+\epsilon)\), computed over valid pixels. Equivalently, \(d_2\) and \(d_{98}\) correspond to the \(2\%\) and \(98\%\) quantiles used for robust clipping. Values outside this interval are clipped before the linear mapping to \([-1,1]\). This representation removes the global scale and shift ambiguity of monocular depth estimation while preserving relative scene geometry. To make the target compatible with the RGB image-editing backbone, we encode the normalized depth map as a grayscale RGB image by replicating the same normalized depth value across the three color channels.

\subsection{Diffusion Transformer Adaptation}

We initialize our model from Qwen-Image-Edit-2509~\cite{wu2025qwen} and adapt it for monocular depth estimation using parameter-efficient fine-tuning. To make training memory-efficient, we apply 4-bit quantization to the pretrained DiT weights and fine-tune rank-128 QLoRA adapters. Following the Lotus-2 rectified-flow formulation, we train the model to directly transform RGB latents into grayscale depth latents using a single forward pass.

Let \(\mathcal{E}_{\mathrm{vae}}\) and \(\mathcal{D}_{\mathrm{vae}}\) denote the pretrained VAE encoder and decoder. We encode the RGB image \(I\) and normalized depth target $d$ as
\begin{equation}
    z_I = \mathcal{E}_{\mathrm{vae}}(I),
    \qquad
    z_d = \mathcal{E}_{\mathrm{vae}}(d).
\end{equation}
We define the target velocity as \(v = z_I - z_d\) and train the DiT \(f_\theta\), conditioned on \(z_I\) and a fixed timestep \(t=0.5\), to regress it:
\begin{equation}
    \mathcal{L}_{\mathrm{latent}}
    =
    \left\|
    f_\theta(z_{I},t) - v
    \right\|_2^2 .
\end{equation}
Since \(t\) is fixed, the rectified-flow parameterization reduces to a direct latent regression with no trajectory to integrate.
At inference, we obtain the depth latent in a single forward pass,
\begin{equation}
    \hat{z}_d = z_I - f_\theta(z_I,t),
\end{equation}
and decode it as \(\hat{d}=\mathcal{D}_{\mathrm{vae}}(\hat{z}_d)\).

Unlike previous dif\-fu\-sion-based monocular depth methods that supervise primarily in latent space, we additionally apply direct image-space reconstruction losses. In particular, we use an \(L_1\) reconstruction loss and an \(L_1\) spatial gradient loss:
\begin{align}
    &\mathcal{L}_{\mathrm{pix}} = 
    \|\hat{d} - d\|_1 \\
    &\mathcal{L}_{\mathrm{grad}}
    =
    \|\nabla_x \hat{d} - \nabla_x d\|_1
    +
    \|\nabla_y \hat{d} - \nabla_y d\|_1 .
\end{align}

\newcommand{\qualheight}{0.14\textheight}
\newcommand{\qualhspace}{0.01\linewidth}

\newcommand{\zoomethimg}[4]{%
\begin{tikzpicture}[%
  spy using outlines={%
    rectangle,%
    magnification=#4,%
    size=2.0cm,%
    connect spies=false,%
    black,%
    thick%
  }%
]
  \node[inner sep=0pt] (img) {
    \includegraphics[height=\qualheight,keepaspectratio]{#1}
  };
  \spy on (#2) in node at (#3);
\end{tikzpicture}%
}

\begin{figure}[t]
  \centering
  \setlength{\parindent}{0pt}%
  \setlength{\fboxsep}{0pt}%
  \resizebox{\columnwidth}{!}{%
    \adjustbox{valign=t}{\includegraphics[height=\qualheight,keepaspectratio]{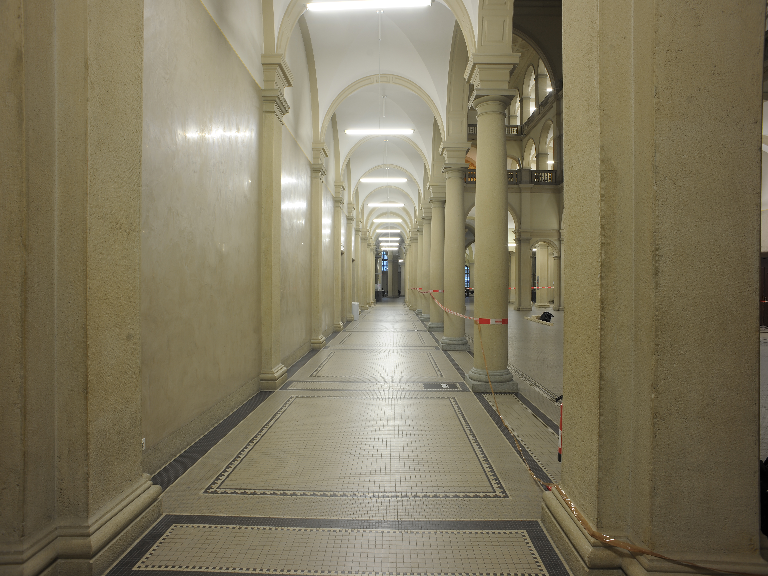}}%
    \hspace{\qualhspace}%
    \adjustbox{valign=t}{\zoomethimg{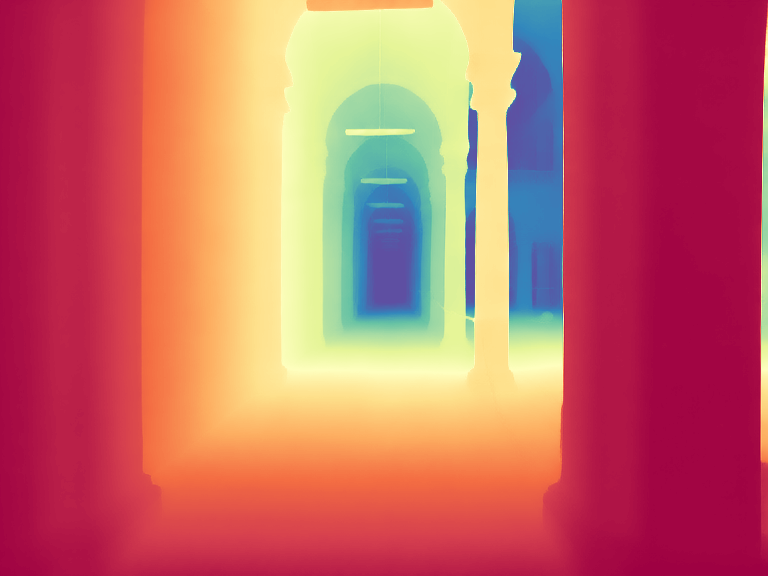}{0.7,0.8}{-1.0,-0.5}{3.0}}%
  }%
  \\[0.03em]
  \resizebox{\columnwidth}{!}{%
    \adjustbox{valign=t}{\zoomethimg{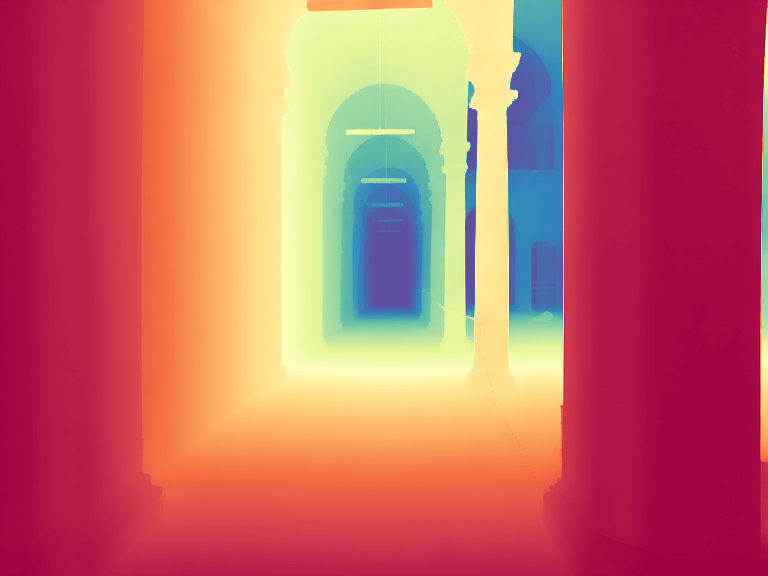}{0.7,0.8}{-1.0,-0.5}{3.0}}%
    \hspace{\qualhspace}%
    \adjustbox{valign=t}{\zoomethimg{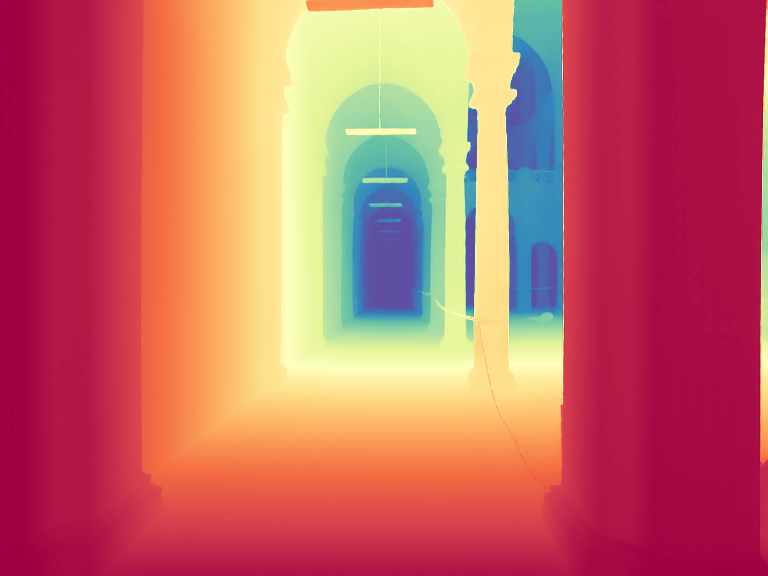}{0.7,0.8}{-1.0,-0.5}{3.0}}%
  }%

  \caption{
  \textbf{Impact of semantic feature losses.}
  Input image, baseline prediction, with LPIPS loss, with iREPA-depth loss. The iREPA-depth variant removes artifacts most efficiently and improves fine details. 
  }
  \label{fig:irepa_qualitative}
  \Description[An RGB image and three depth predictions arranged in a two-by-two grid.]{An RGB image of a hall with pillars alongside baseline, LPIPS, and iREPA depth predictions. The iREPA prediction has sharper structural details.}
\end{figure}

\subsection{Semantic Feature Regularization with iREPA}

To improve reconstruction quality in semantically dense regions, we introduce a variant of iREPA feature alignment loss~\cite{singh2025irepa}. While pixel-level losses encourage accurate local reconstruction, they do not explicitly enforce consistency at higher levels of visual structure. As a result, predictions may still lose fine details in cluttered regions such as foliage, bushes, and repeated object patterns. We therefore regularize the internal DiT representations using pretrained visual features, encouraging the model to also preserve semantically-meaningful structure in the prediction. %

We compare iREPA regularization using DINOv3~\cite{simeoni2025dinov3} features extracted from RGB images and from ground-truth depth maps. While both variants improve over the baseline, features extracted from the depth map provide better AbsRel and $\delta_1$ performance, suggesting that the feature extraction in the depth domain provides more relevant information for geometric reconstruction than RGB-derived features. 
Qualitatively, iREPA improves the semantic and structural consistency of the predicted depth maps in visually dense regions, as shown in Fig.~\ref{fig:irepa_qualitative}.
A direct perceptual loss (LPIPS~\cite{zhang2018unreasonable}) between the prediction and ground truth does not deliver the same improvement in dense regions.

The Stage-1 DiT training objective is therefore:
\begin{equation}
\label{eq:stage1}
    \mathcal{L}_{\mathrm{DiT}}
    =
    \lambda_{\mathrm{latent}}\mathcal{L}_{\mathrm{latent}}
    +
    \lambda_{\mathrm{pix}}\mathcal{L}_{\mathrm{pix}}
    +
    \lambda_{\mathrm{grad}}\mathcal{L}_{\mathrm{grad}}
    +
    \lambda_{\mathrm{iREPA}}\mathcal{L}_{\mathrm{iREPA}} .
\end{equation}

\begin{figure}[t]
  \centering
  \begin{subfigure}{0.495\linewidth}
    \includegraphics[width=\linewidth]{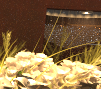}
    \caption{HyperSim RGB Example}
    \label{fig:hypersim_issue:a}
  \end{subfigure}
  \hfill
  \begin{subfigure}{0.495\linewidth}
    \includegraphics[width=\linewidth]{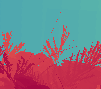}
    \caption{HyperSim Ground Truth}
    \label{fig:hypersim_issue:b}
  \end{subfigure}
  \caption{\textbf{A crop of a HyperSim sample in native resolution.} Thin structures appear semi-transparent in RGB and are not well-defined in ground-truth depth due to stochastic sampling within the rendering pipeline. }
  \label{fig:hypersim_issue:triple}
  \Description[An image crop from HyperSim dataset, highlighting ambiguities in correspondence of thin objects on both RGB and ground truth depth.]{An image crop from HyperSim dataset, highlighting ambiguities in correspondence of thin objects on both RGB and ground truth depth, nearly disappearing on color images and being assigned to inaccurate depth values in the ground truth.}
\end{figure}

\begin{figure}[t]
  \centering
\includegraphics[width=\linewidth]{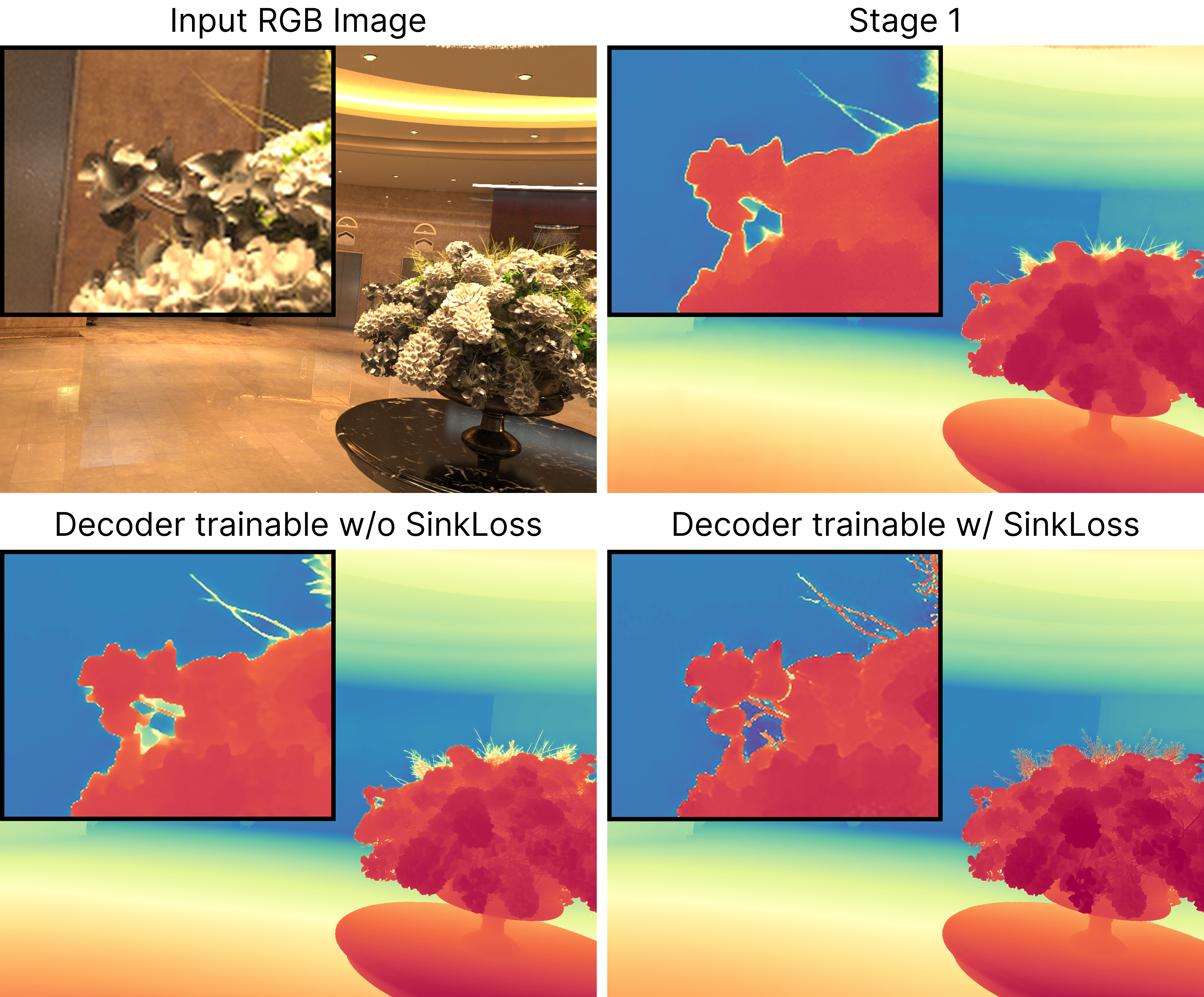}
\caption{\textbf{Qualitative impact of SinkLoss and VAE decoder unfreezing.}
The Stage-1 checkpoint is compared against two continued-training variants with an unfrozen VAE decoder, with and without SinkLoss. SinkLoss substantially reduces flying pixels while preserving fine detail.}
  \label{fig:sinkloss_impact}
  \Description[Example figure showing recovery of details when using our stage 2 training]{Qualitative comparison of depth map results showing an input image followed by three training stages: Stage 1, Stage 2 with an unfrozen VAE but no SinkLoss, and Stage 2 with both an unfrozen VAE and SinkLoss, highlighting the reduction of artifacts.}
\end{figure}

\subsection{Stage-2 Refinement with SinkLoss}

In addition to the global scale ambiguity inherent to monocular depth estimation, transparent and very thin objects introduce a further source of ambiguity. As illustrated in Fig.~\ref{fig:hypersim_issue:triple}, even in high-quality synthetic datasets like HyperSim~\cite{roberts2021hypersim}, the depth ground truth of thin objects is often noisy. To achieve the best AbsRel and $\delta_1$ scores, the predictions would have to align perfectly with the ground truth \emph{and} reproduce that noise at exactly the same pixel locations. The underlying rendering pipeline, however, uses V-Ray with quasi-Monte Carlo sampling, so it is essentially random whether a transparent or edge pixel is assigned a foreground or a background depth value. 
This makes perfect AbsRel effectively unattainable and undesirable as a target, because many downstream tasks benefit from a cohesive depth map over the foreground object with a sharp transition to the background.

To address local ambiguities of noisy ground-truth supervision that are not well handled by strict pixel-wise supervision, we continue fine-tuning the depth estimator during Stage 2 using a novel SinkLoss.
Instead of supervising each pixel directly against its corresponding ground-truth pixel, we tile the image into non-overlapping $K\!\times\!K$ blocks and, within each block, use Sinkhorn--Knopp matching between the $K^2$ predicted depths and the $K^2$ ground-truth depths to obtain a soft one-to-one assignment.
This only requires the network to produce the same set of depth values as the ground truth within each block (up to permutation), without strict spatial alignment.

\paragraph{Formulation.}
Within each non-overlapping $K\!\times\!K$ block we build a
cost matrix $\mathbf{C}\in\mathbb{R}^{K^2 \times K^2}$ between the $K^2$
predicted depths $\{\hat{d}_i\}$ and the $K^2$ ground-truth depths
$\{d_j\}$,
\begin{equation}
    C_{ij} \;=\; \bigl|\,\hat{d}_i - d_j\,\bigr|.
\end{equation}
Invalid pixels (mask $m_i{=}0$) are excluded by replacing the cost of every pair
that touches one, $\tilde{C}_{ij} = C_{ij}$ if $m_i m_j = 1$ and
$\tilde{C}_{ij} = B \gg \max_{ij} C_{ij}$ otherwise, i.e.\ the invalid pixel's
row \emph{and} its column are penalized. A block has as many penalized rows as
penalized columns, so for large $B/\tau$ the optimal plan matches them to one
another: no invalid ground-truth pixel supervises a prediction, and the valid
pixels are left with uniform marginals.
We then compute a soft assignment $\mathbf{M}$ by entropy-regularized optimal
transport,
\begin{equation}
    \mathbf{M} \;=\; \argmin_{\mathbf{M}\in\mathcal{U}}\;
        \langle \mathbf{M},\,\tilde{\mathbf{C}}\rangle \;-\; \tau\,H(\mathbf{M}),
    \label{eq:sinkhorn-ot}
\end{equation}
where $\mathcal{U}=\{\mathbf{M}\in\mathbb{R}_{\ge 0}^{K^2\times K^2} :
\mathbf{M}\mathbf{1}=\mathbf{1}/K^2,\,\mathbf{M}^{\!\top}\mathbf{1}=\mathbf{1}/K^2\}$
is the transport polytope with uniform marginals and
$H(\mathbf{M})=-\sum_{ij} M_{ij}\log M_{ij}$ is the Shannon entropy.
$\mathbf{M}$ is obtained by Sinkhorn--Knopp
iterations~\cite{sinkhorn1967concerning,cuturi2013sinkhorn} on
$\mathbf{G}=\exp(-\tilde{\mathbf{C}}/\tau)$, which we run in the log domain for
numerical stability.   %
The SinkLoss is the transport cost over the valid pairs,
\begin{equation}
    \mathcal{L}_{\text{SinkLoss}} \;=\;
    \frac{\sum_{ij} m_i m_j\, M_{ij}\,C_{ij}}{\sum_{ij} m_i m_j\, M_{ij}},
\end{equation}
averaged over blocks with valid ground truth.
We use $K=5$, $\tau=0.1$, $B=10^{6}$ and $5$ Sinkhorn iterations.
Applying this loss at Stage 2 results in fewer flying-pixel artifacts while preserving fine details in the depth map, as demonstrated in Fig.~\ref{fig:sinkloss_impact}.

\begin{table*}[t]
  \centering
  \caption{\textbf{Comparison of zero-shot affine-invariant monocular depth estimators on NYUv2, KITTI, ETH3D, ScanNet, and DIODE.} We report AbsRel~($\downarrow$) and $\delta_1$ ($\uparrow$). \textbf{Bold} indicates the best result and \underline{underlining} the second-best result for each metric. Methods trained on more than 5M images are given in \nocompete{\textbf{gray}} for reference and are excluded from ranking. Results for InfiniDepth, Lotus-2, FE2E, and our method are reproduced under the same evaluation protocol, while the remaining results are taken from the PPD paper. $^\dagger$$\pi^3$ includes ScanNet in its training data; its ScanNet results are therefore not zero-shot.
  }
    \label{tab:depth_comparison_lotus_fe2e}
  \resizebox{\linewidth}{!}{%
  \begin{tabular}{@{}lccccccccccc}
  \toprule
  \multirow{2}{*}{Method} & \multirow{2}{*}{\shortstack[c]{Training \\ Data $\downarrow$}} & \multicolumn{2}{c}{NYUv2} & \multicolumn{2}{c}{KITTI} & \multicolumn{2}{c}{ETH3D} & \multicolumn{2}{c}{ScanNet} & \multicolumn{2}{c}{DIODE} \\
  & & AbsRel $\downarrow$ & $\delta_1$ $\uparrow$ & AbsRel $\downarrow$ & $\delta_1$ $\uparrow$ & AbsRel $\downarrow$ & $\delta_1$ $\uparrow$ & AbsRel $\downarrow$ & $\delta_1$ $\uparrow$ & AbsRel $\downarrow$ & $\delta_1$ $\uparrow$ \\
  \midrule
  \nocompete{\textbf{Omnidata} {\scriptsize (ICCV 2021)~\cite{eftekhar2021omnidata}}}               & \nocompete{12.2M} & \nocompete{7.4}  & \nocompete{94.5} & \nocompete{14.9} & \nocompete{83.5} & \nocompete{16.6} & \nocompete{77.8} & \nocompete{7.5}  & \nocompete{93.6} & \nocompete{-}    & \nocompete{-}    \\
    \nocompete{\textbf{DepthAnything\ V2} (62M) {\scriptsize (NeurIPS 2024)~\cite{yang2024depth2}}}
    & \nocompete{62.6M} & \nocompete{4.5}  & \nocompete{97.9} & \nocompete{7.4}  & \nocompete{94.6} & \nocompete{13.1} & \nocompete{86.5} & \nocompete{6.5}  & \nocompete{97.2} & \nocompete{6.6}  & \nocompete{95.2} \\
  \nocompete{\textbf{MoGe} {\scriptsize (CVPR 2025)~\cite{wang2025moge}}} & \nocompete{9M} & \nocompete{3.1} & \nocompete{98.3} & \nocompete{7.2} & \nocompete{95.1} & \nocompete{10.0} & \nocompete{89.6} & \nocompete{3.3} & \nocompete{98.3} & \nocompete{29.0} & \nocompete{70.0} \\
  \nocompete{\textbf{MoGe-2} {\scriptsize (NeurIPS 2025)~\cite{wang2025moge2}}} & \nocompete{8.9M} & \nocompete{3.1} & \nocompete{98.4} & \nocompete{6.9} & \nocompete{89.9} & \nocompete{5.0} & \nocompete{96.2} & \nocompete{3.1} & \nocompete{98.3} & \nocompete{9.6} & \nocompete{84.5} \\
  \nocompete{\textbf{$\pi^3$}$^\dagger$ {\scriptsize (ICLR 2026)~\cite{wang2025pi}}} & \nocompete{>9M} & \nocompete{2.9} & \nocompete{98.7} & \nocompete{5.9} & \nocompete{96.7} & \nocompete{2.9} & \nocompete{98.9} & \nocompete{2.1}$^\dagger$ & \nocompete{99.1}$^\dagger$ & \nocompete{4.4} & \nocompete{96.8} \\
  \midrule
\textbf{DiverseDepth} {\scriptsize (arXiv 2020)~\cite{yin2020diversedepth}}           & 320K & 11.7 & 87.5 & 19.0 & 70.4 & 22.8 & 69.4 & 10.9 & 88.2 & -    & -    \\
\textbf{MiDaS} {\scriptsize (TPAMI 2022)~\cite{ranftl2020towards}}                  & 2M & 11.1 & 88.5 & 23.6 & 63.0 & 18.4 & 75.2 & 12.1 & 84.6 & -    & -    \\
\textbf{LeReS} {\scriptsize (CVPR 2021)~\cite{yin2021learning}}                 & 354K & 9.0  & 91.6 & 14.9 & 78.4 & 17.1 & 77.7 & 9.1  & 91.7 & -    & -    \\
\textbf{DPT} {\scriptsize (ICCV 2021)~\cite{ranftl2021vision}}                    & 1.4M & 9.8  & 90.3 & 10.0 & 90.1 & 7.8  & 94.6 & 8.2  & 93.4 & -    & -    \\
\textbf{HDN} {\scriptsize (NeurIPS 2022)~\cite{zhang2022hierarchical}}                     & 300K & 6.9  & 94.8 & 11.5 & 86.7 & 12.1 & 83.3 & 8.0  & 93.9 & -    & -    \\
\textbf{DepthAnything\ V2 (54K)} {\scriptsize (NeurIPS 2024)~\cite{yang2024depth2}}
    & 54K & 5.4  & 97.2 & 8.6  & 92.8 & 12.3 & 88.4 & -    & -    & 8.8  & 93.7 \\
\textbf{Marigold Depth V1} {\scriptsize (CVPR 2024)~\cite{ke2024repurposing,ke2025marigold}}              & 74K & 5.5  & 96.4 & 9.9  & 91.6 & 6.5  & 96.0 & 6.4  & 95.1 & 10.0 & 90.7 \\
\textbf{GeoWizard} {\scriptsize (ECCV 2024)~\cite{fu2024geowizard}}              & 280K & 5.2  & 96.6 & 9.7  & 92.1 & 6.4  & 96.1 & 6.1  & 95.3 & 12.0 & 89.8 \\
\textbf{DepthFM} {\scriptsize (AAAI 2025)~\cite{gui2025depthfm}}                & 63K & 5.5  & 96.3 & 8.9  & 91.3 & 5.8  & 96.2 & 6.3  & 95.4 & -    & -    \\
\textbf{GenPercept} {\scriptsize (ICLR 2025)~\cite{xu2025matters}}             & 90K & 5.2  & 96.6 & 9.4  & 92.3 & 6.6  & 95.7 & 5.6  & 96.5 & -    & -    \\
\textbf{Lotus} {\scriptsize (ICLR 2025)~\cite{he2025lotus1}}                  & 59K & 5.4  & 96.8 & 8.5  & 92.2 & 5.9  & 97.0 & 5.9  & 95.7 & 9.8  & 92.4 \\
\textbf{Lotus\mbox{-}2} {\scriptsize (arXiv 2025)~\cite{he2025lotus2}} & 59K & \cellsecond{\underline{3.7}} & \cellthird{97.6} & \cellthird{6.7} & 94.1 & \cellthird{4.1} & \cellthird{98.6} & \cellsecond{\underline{4.0}} & \cellthird{97.2} & 6.5 & 95.5 \\
\textbf{PPD (512)} {\scriptsize (NeurIPS 2025)~\cite{xu2026pixel}}              & 54K & 4.3  & 97.4 & 8.0  & 93.1 & 4.5  & 97.7 & 4.5  & \cellsecond{\underline{97.3}} & 7.0  & 95.5 \\
\textbf{PPD (1024)} {\scriptsize (NeurIPS 2025)~\cite{xu2026pixel}}             & 125K & 4.1  & \cellsecond{\underline{97.7}} & 7.0  & \cellthird{95.5} & 4.3  & 98.0 & 4.6  & \cellthird{97.2} & 6.8  & \cellthird{95.9} \\
\textbf{InfiniDepth} {\scriptsize (CVPR 2026)~\cite{yu2026infinidepth}} & 3.1M & 4.3 & \cellthird{97.6} & 8.7 & 92.3 & 6.1 & 95.4 & 4.7 & 96.9 & \cellthird{6.4} & 95.8 \\
\textbf{DepthMaster} {\scriptsize (TCSVT 2026)~\cite{song2026depthmaster}} & 74K & 4.8 & 97.0 & 9.1 & 91.1 & 5.4 & 97.4 & 5.8 & 95.6 & 8.0 & 94.3 \\ 
\textbf{FE2E} {\scriptsize (CVPR 2026)~\cite{wang2026fe2e}} & 71K & \cellthird{3.8} & \cellthird{97.6} & \cellsecond{\underline{6.5}} & \cellsecond{\underline{96.0}} & \cellsecond{\underline{3.8}} & \cellsecond{\underline{98.7}} & \cellthird{4.3} & 97.1 & \cellsecond{\underline{5.6}} & \cellsecond{\underline{96.4}} \\
\midrule
\textbf{\methodname (depth, ours)} & 74K & \cellfirst{\textbf{3.6}} & \cellfirst{\textbf{98.0}} & \cellfirst{\textbf{5.4}} & \cellfirst{\textbf{97.4}} & \cellfirst{\textbf{2.8}} & \cellfirst{\textbf{99.2}} & \cellfirst{\textbf{3.7}} & \cellfirst{\textbf{97.9}} & \cellfirst{\textbf{5.2}} & \cellfirst{\textbf{97.1}} \\

\bottomrule
  \end{tabular}  
  }
\end{table*}

\section{Experiments}

\label{sec:experiments}

\subsection{Implementation Details}\label{sec:implementation}

We build our method on top of the Qwen-Image-Edit-2509 model and fine-tune its DiT backbone using QLoRA. Specifically, we quantize the pretrained model weights to 4-bit precision, and we train rank-$128$ LoRA adapters. Batch size is set to 1 in all training runs, and gradient clipping is used to stabilize optimization and reduce the influence of poor-quality or noisy training samples. This setup substantially reduces memory usage while preserving the representational capacity needed for adapting the image-editing backbone to monocular depth estimation.

Following Marigold V1, we train on a deliberately compact mixture of HyperSim~\cite{roberts2021hypersim} and vKITTI~\cite{gaidon2016virtual} datasets. 
For HyperSim, we filter out samples with more than $0.1\%$ invalid depth pixels. 
Considering the different aspect ratios of the two datasets, vKITTI samples are processed at a resolution of $1216 \times 352$, while HyperSim samples are processed at $768 \times 512$. 
Each training batch is constructed by sampling from HyperSim and vKITTI with probabilities of $90\%$ and $10\%$, respectively.

All training experiments are performed on a single 32GB GPU.
Ablation experiments are trained for $30{,}000$ optimization steps. Each ablation run requires approximately one day to complete. We then perform Stage-2 SinkLoss refinement by continuing fine-tuning from the Stage-1 checkpoint for an additional $30{,}000$ steps, which also takes approximately one day under the same setup.

For the final Stage-1 model, we extend the training to $160{,}000$ optimization steps using the same single-GPU setup. We set 
$\lambda_{\mathrm{latent}}{=}1.0$, 
$\lambda_{\mathrm{pix}}{=}1.0$, 
$\lambda_{\mathrm{grad}}{=}5.0$, and 
$\lambda_{\mathrm{iREPA}}{=}0.2$. This final Stage 1 training run requires slightly more than five days to complete on the same GPU. For Stage 2, we set $\lambda_{\mathrm{iREPA}}{=}0.2$ and $\lambda_{\mathrm{SinkLoss}}{=}1.0$.

\begin{table}[t]
  \centering
  \caption{\textbf{Edge-aware evaluation of affine-invariant monocular depth estimators on the HyperSim test set.} We report SEE3, SEE5 and SEE7 ($\downarrow$). \textbf{Bold} indicates the best result and \underline{underlining} the second-best result for each metric, determined from unrounded values. 
  }
  \resizebox{0.9\linewidth}{!}{%
  \begin{tabular}{@{}lcccccccccc@{}}
  \toprule
  Method & SEE3 $(\downarrow)$ & SEE5 $(\downarrow)$ & SEE7 $(\downarrow)$ \\
  \midrule
  \textbf{PPD} {\scriptsize (NeurIPS 2025)~\cite{xu2026pixel}} & \underline{0.404} & \underline{0.385} & \underline{0.371} \\
  \textbf{InfiniDepth} {\scriptsize (CVPR 2026)~\cite{yu2026infinidepth}} & 0.470 & 0.451 & 0.436 \\
  \midrule
  \bf \methodname (depth, ours) & \textbf{0.352} & \textbf{0.333} & \textbf{0.320}\\
    \bottomrule
  \end{tabular}  
  }
  \label{tab:see_metrics}
\end{table}

\begin{table*}[t]
\centering
\caption{\textbf{Ablation study on Stage 1.}
AbsRel and $\delta_1$ comparison across datasets using different Stage 1 loss configurations.
}
\resizebox{0.8\linewidth}{!}{
\begin{tabular}{@{}cccccccccccccc@{}}
\toprule

\multirow{2}{*}{$\mathcal{L}_{\mathrm{latent}}$} & 
$\mathcal{L}_{\mathrm{pix}}$ &
\multirow{2}{*}{$\mathcal{L}_{\mathrm{iREPA}}^{\mathrm{rgb}}$} &
\multirow{2}{*}{$\mathcal{L}_{\mathrm{iREPA}}^{\mathrm{depth}}$} & 
\multicolumn{2}{c}{NYUv2} &
\multicolumn{2}{c}{KITTI} & \multicolumn{2}{c}{ETH3D} & \multicolumn{2}{c}{ScanNet} & \multicolumn{2}{c}{DIODE}
\\

&
$\mathcal{L}_{\mathrm{grad}}$ &
&
&
AbsRel $\downarrow$
&
$\delta_1$ $\uparrow$
&
AbsRel $\downarrow$
&
$\delta_1$ $\uparrow$
&
AbsRel $\downarrow$
&
$\delta_1$ $\uparrow$
&
AbsRel $\downarrow$
&
$\delta_1$ $\uparrow$
&
AbsRel $\downarrow$
&
$\delta_1$ $\uparrow$

\\

\midrule

\multicolumn{14}{l}{\textit{Stage 1 ablations, 30K training steps}} \\

\cmark & \xmark & \xmark & \xmark &
4.53 & 97.60 &
7.26 & 94.94 &
4.25 & 98.25 &
4.59 & 97.41 &
6.82 & 95.81
\\

\cmark & \cmark & \xmark & \xmark &
4.70 & \underline{97.90} &
7.84 & 95.62 &
4.32 & 98.69 &
4.50 & \textbf{97.91} &
6.22 & 96.01
\\

\cmark & \cmark & \cmark & \xmark &
\underline{4.50} & 97.81 &
\underline{6.89} & \underline{96.22} &
\underline{3.64} & \underline{98.86} &
\underline{4.38} & \underline{97.65} &
\underline{5.72} & \underline{96.54}
\\

\cmark & \cmark & \xmark & \cmark &
\textbf{4.36} & \textbf{98.01} &
\textbf{6.72} & \textbf{96.38} &
\textbf{3.62} & \textbf{98.90} &
\textbf{4.22} & \textbf{97.91} &
\textbf{5.55} & \textbf{96.82}
\\

\arrayrulecolor{gray}\midrule\arrayrulecolor{black}

\multicolumn{14}{l}{\textit{Stage 1, 160K training steps}} \\

\cmark & \cmark & \xmark & \xmark &
\textbf{3.62} & \underline{98.01} &
\textbf{5.27} & \underline{97.39} &
\underline{2.84} & \underline{99.08} &
\underline{3.84} & \textbf{97.88} &
\underline{5.16} & \textbf{97.07}
\\

\cmark & \cmark & \xmark & \cmark &
\underline{3.68} & \textbf{98.03} &
\underline{5.30} & \textbf{97.43} &
\textbf{2.68} & \textbf{99.18} &
\textbf{3.81} & \underline{97.82} &
\textbf{5.02} & \underline{97.02}
\\

\bottomrule
\end{tabular}
}

\label{tab:stage1_abalations}

\end{table*}

\begin{figure*}[ht!]
  \centering
    \includegraphics[width=\linewidth]{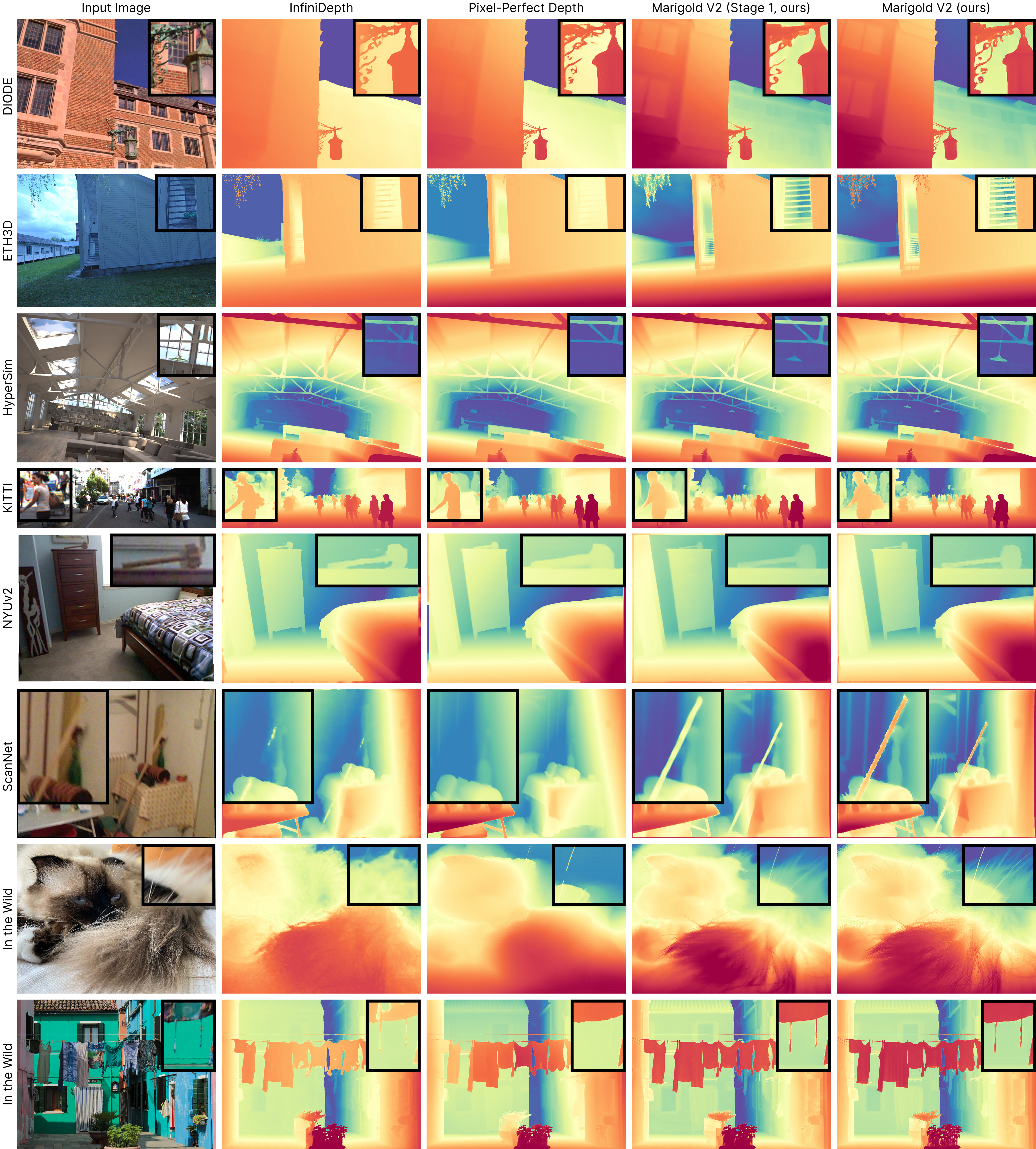}
  \caption{\textbf{Qualitative comparison of state-of-the-art relative depth estimation methods across multiple datasets.} \methodname trained with iREPA-depth and SinkLoss preserves fine details while keeping flying pixels to a mi\-ni\-mum. Pixel-Perfect Depth produces fewer flying pixels overall but loses substantial detail. All methods use the same inference resolution without any resizing. Last row courtesy of Franco Sulli via Pexels. }
  \label{fig:appendix:qualitative_comparison:a}
  \Description[Qualitative comparison grid]{Qualitative comparison grid showing relative depth estimation across eight diverse samples, comparing the proposed model against Infinidepth, Pixel-Perfect-Depth, and its own Stage 1 baseline.}
\end{figure*}

\begin{table}[thb]
  \centering
\caption{\textbf{Impact of depth parameterizations.}
  Average AbsRel and $\delta_1$ across test sets 
  under a common training and evaluation configuration.}
  \label{tab:sup:stage1_abalations}
  \resizebox{0.65\linewidth}{!}{
  \setlength{\tabcolsep}{18pt}
  \begin{tabular}{@{}lcc@{}}
  \toprule
    Representation & AbsRel $\downarrow$ & $\delta_1$ $\uparrow$ \\
  \midrule
    Linear Depth     & \underline{5.04} & 97.10 \\
    Disparity & 5.28 & \underline{97.15} \\
    Log Depth & \textbf{4.72} & \textbf{97.71} \\
  \bottomrule
  \end{tabular}
  }
\end{table}

\subsection{Experimental Setup}

We follow the evaluation protocol of Pixel-Perfect Depth~\cite{xu2026pixel}, aligning predictions to the ground-truth metric depth with a robust RANSAC procedure before computing metrics. This removes global scale and shift mismatches, focusing the evaluation on scene geometry, local structure, and depth discontinuities.

\subsection{Zero-Shot Affine Depth Estimation}

We assess the zero-shot generalization ability of our method on KITTI~\cite{geiger2012kitti}, ETH3D~\cite{schoeps2017eth3d}, ScanNet ~\cite{dai2017scannet}, NYUv2~\cite{silberman2012nyu}, and DIODE~\cite{vasiljevic2019diode}. For each dataset, predictions are aligned to the ground truth following the protocol described above and evaluated using the standard AbsRel and $\delta_1$ metrics. AbsRel measures the mean absolute relative error, while $\delta_1$ reports the percentage of valid pixels whose predicted depth is within a factor of $1.25$ of the ground-truth depth. We run inference at the native image resolution for all datasets except ETH3D, where we use an input resolution of $1008 \times 672$ and upsample the predictions to $2048 \times 1360$ before alignment and evaluation, following the evaluation procedure of Pixel-Perfect Depth~\cite{xu2026pixel}.

The quantitative results are reported in Tab.~\ref{tab:depth_comparison_lotus_fe2e}. We compare our Stage-2 checkpoint against both discriminative and generative monocular depth estimation methods, including Depth Anything V2, Marigold V1, GeoWizard, DepthFM, GenPercept, Lotus, Pixel-Perfect Depth, DepthMaster, Lotus-2, and FE2E. Our method achieves the best zero-shot performance across all evaluated datasets among methods trained on comparable data and obtains especially strong results on the challenging ETH3D, where our best model achieves an AbsRel of $2.8$, substantially improving over the strongest baseline result of $3.8$.
Fig.~\ref{fig:appendix:qualitative_comparison:a} compares the methods qualitatively
across these benchmarks.

\subsection{Edge-Aware Evaluation}

In addition to standard depth accuracy metrics, we evaluate the effectiveness of \methodname in terms of preservation of fine 
geometric details and the reduction of flying-pixel artifacts near depth discontinuities. 
For this purpose, we use the Soft Edge Error (SEE$_k$), computed on object boundaries in the synthetic HyperSim dataset. We identify these as pixels whose disparity change with respect to their neighbors exceeds a fixed threshold. Following prior work \cite{chen2019over_smoothing, tosi2021smd}, SEE$_k$ is defined for each such pixel as the minimum absolute error between the predicted disparity value and any ground-truth disparity value within a local $k \times k$ patch centered on that pixel, with $k \in \{3,5,7\}$.  %

Tab.~\ref{tab:see_metrics} compares \methodname against state-of-the-art methods spe\-ci\-fically designed to preserve fine details and suppress flying pixels, where our model achieves the best results across all metrics.

\subsection{Analysis and Ablation Studies}

We ablate our design choices and analyze how well the recipe transfers.
First, we analyze the choices used during the first stage of DiT fine-tuning.
We evaluate the impact of adding the iREPA feature regularization loss, computed using DINOv3 features extracted either from the RGB input image or ground-truth depth maps. 
Next we compare the effect of various depth range compression representations on the quantitative metrics and motivate log-depth as the optimal one.
Additionally, we study the effect of varying the DiT backbone and the effectiveness of SinkLoss under such a change.
We further analyze the effect of additional training steps and the impact of SinkLoss on the second stage.

\begin{table}[t]
\centering
\caption{\textbf{SinkLoss with other diffusion backbones.} SinkLoss transfers to Stable Diffusion and FLUX.2. %
AbsRel is averaged across five test datasets, while Soft Edge Error (SEE) metrics are computed on the HyperSim test set.}
\label{tab:sd15_backbone}
\resizebox{0.95\linewidth}{!}{
\begin{tabular}{@{}lcccccc@{}}
\toprule
Model & $\mathcal{L}_\mathrm{SinkLoss}$ & AbsRel $\downarrow$ & SEE$_3$ $\downarrow$ & SEE$_5$ $\downarrow$ & SEE$_7$ $\downarrow$ \\
\midrule
\multirow{2}{*}{\shortstack[l]{\textbf{Stable Diffusion V1.5}\\ \small{\cite{rombach2022high}}}} & \xmark & 7.51 & 0.553 & 0.531 & 0.514  \\
 & \cmark & \textbf{7.12} & \textbf{0.485} & \textbf{0.464} & \textbf{0.449} \\
\arrayrulecolor{gray}\midrule\arrayrulecolor{black}
\multirow{2}{*}{{\shortstack[l]{\textbf{FLUX.2 klein} \\ \small{\cite{bfl2024flux2}}}}} & \xmark & 4.88 & 0.491 & 0.472 & 0.457  \\
 & \cmark & \textbf{4.82} & \textbf{0.377} & \textbf{0.359} & \textbf{0.345} \\
\arrayrulecolor{gray}\midrule\arrayrulecolor{black}
\multirow{2}{*}{{\shortstack[l]{\textbf{Qwen-Image-Edit-2509} \\[-1pt] \small{\cite{wu2025qwen}}}}} &  \xmark & \textbf{4.10} & 0.449 & 0.429 &  0.414 \\
 & \cmark & 4.12 & \textbf{0.352} & \textbf{0.333} & \textbf{0.320} \\
\bottomrule
\end{tabular}
}
\end{table}

\textbf{Stage-1 Ablation.} Tab.~\ref{tab:stage1_abalations} reports results concerning the first fine-tuning stage. In the controlled 30K-step setting (top), both iREPA variants improve over the baseline without feature regularization, with iREPA-depth providing the strongest overall performance across the evaluated datasets. We also observe that adding pixel-space supervision on top of the latent MSE loss has only a limited effect on AbsRel, but consistently improves $\delta_1$ on all datasets. In the visualizations, pixel-space losses, and iREPA-depth in particular, further reduce artifacts common to DiT-based depth estimators. 

As the number of training steps increases to 160K (bottom), the impact of iREPA becomes less pronounced in terms of AbsRel and $\delta_1$. Nevertheless, even after 160K training steps, the qualitative improvements introduced by iREPA remain visible, as shown in Fig.~\ref{fig:irepa_qualitative}. This suggests that while longer training can reduce the quantitative gap between variants, feature regularization still improves the visual quality of the predicted depth maps, particularly in regions with fine structures and complex geometry.
iREPA-depth thus accelerates convergence in the cost-efficient \methodname training protocol.

{\emergencystretch=2em
\textbf{Log-Depth Parameterization.} Previous work, including FE2E~\cite{wang2026fe2e} and DepthFM~\cite{gui2025depthfm} showed that log-depth prediction can improve depth estimation accuracy over uniform-depth representations. Recent methods such as InfiniDepth~\cite{yu2026infinidepth} and Pixel-Perfect Depth~\cite{xu2026pixel} also adopt log-depth as their prediction representation.

As shown in Tab.~\ref{tab:sup:stage1_abalations}, log-depth improves both AbsRel and $\delta_1$.
This behavior can be explained by the close relationship between log-depth error and the relative depth error at the core of these metrics. Let $\epsilon=(x_{\mathrm{pred}}-x_{\mathrm{gt}})/x_{\mathrm{gt}}$; for $|\epsilon|\ll1$,
\begin{equation}
\log x_{\mathrm{pred}}-\log x_{\mathrm{gt}}
=\log\left(1+\frac{x_{\mathrm{pred}} - x_{\mathrm{gt}}}{x_{gt}}\right)\approx\frac{x_{\mathrm{pred}}-x_{\mathrm{gt}}}{x_{\mathrm{gt}}}=\epsilon.
\label{eq:log_relative_error_compact}
\end{equation}
Hence, an $L_1$ penalty on log-depth coincides with the per-pixel \mbox{AbsRel} error to first order.

}

\begin{figure}[t]
  \centering
  \includegraphics[trim={0 3 35 0}, clip, width=\linewidth]{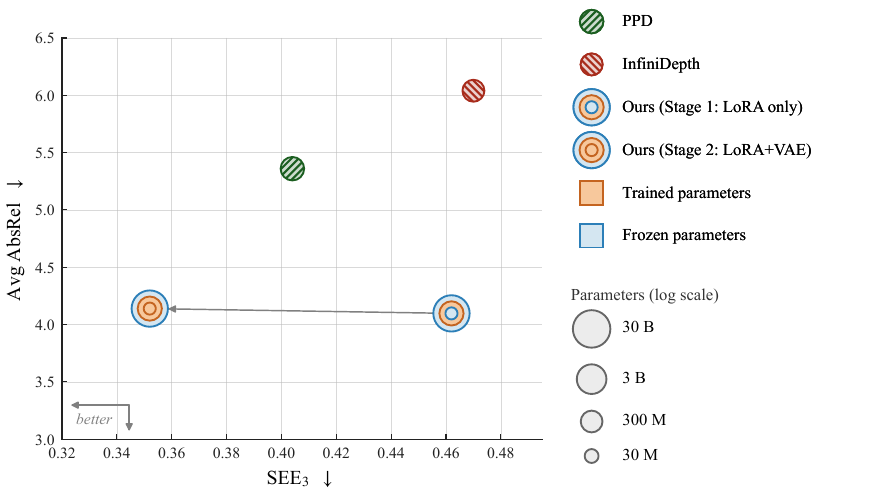}
  \caption{\textbf{%
The impact of SinkLoss on evaluation metrics.
}
SinkLoss greatly improves fine details and edge sharpness, as measured by SEE$_k$ metrics, without affecting standard metrics (AbsRel or $\delta_1$).
With SinkLoss enabled during Stage 2 of our protocol, \methodname outperforms PPD on SEE$_3$.
}
  \label{fig:stage2_abalation}
  \Description[Plot showing improvements on SEE metric when using SinkLoss]{Plot showing improvements on SEE metric when using SinkLoss. SEE metric is improved, but at the same time, the Average AbsRel metric remains the same.}
\end{figure}

\textbf{Backbone Transfer Ablation.} To disentangle the effect of the proposed training recipe from the model prior of the Qwen image-editing backbone, we additionally apply the same losses to Stable Diffusion V1.5, following a setup similar to Marigold V1.1 and MarigoldE2E, as well as to FLUX.2 klein backbone. 
For each backbone, the baseline retains the single-step log-depth formulation with latent MSE, iREPA, and pixel-space losses; SinkLoss is then enabled on top.
The resulting boundary-sensitive SEE metrics on HyperSim are reported in Tab.~\ref{tab:sd15_backbone}, highlighting that the proposed losses also generalize beyond the Qwen backbone. Furthermore, iREPA yields similar qualitative improvements in challenging high-frequency regions, such as dense foliage and thin structures, while SinkLoss improves boundary-sensitive metrics. These results indicate that the gains come from the training recipe rather than from design choices specific to the Qwen backbone.

\begin{table}[t]
  \centering
    \caption{\textbf{Latency and memory benchmarks.} Comparison between different models across resolutions on a single 32GB GPU.}
  \label{tab:latency_comparison}
  \resizebox{\columnwidth}{!}{%
  \begin{tabular}{@{}lcccccc@{}}
  \toprule
  Model& \multicolumn{2}{c}{$1024 \times 1024$} & \multicolumn{2}{c}{$2048 \times 2048$} \\
  & Lat. (s) & Mem. (GB) & Lat. (s) & Mem. (GB) \\
  \midrule
  \textbf{InfiniDepth} \scriptsize{(CVPR 2026) \cite{yu2026infinidepth}}  & 0.2 & 1.9 & 1.2 & 3.4 \\
  \textbf{PPD} \scriptsize{(NeurIPS 2025) \cite{xu2026pixel}} & 1.4 & 5.6 & OOM & OOM \\
  \textbf{Lotus-2} \scriptsize{(w/o sharpener)} & 1.1 & 26.1 & OOM & OOM \\
  \textbf{Lotus-2}  \scriptsize{(arXiv 2025) \cite{he2025lotus2}} & 8.9 & 26.1 & OOM & OOM \\
  \textbf{FE2E}  \scriptsize{(CVPR 2026) \cite{wang2026fe2e}} & 3.9 & 27.5 & OOM & OOM \\
  \midrule
    \textbf{\methodname (depth, ours)} & 1.9 & 16.9 & 9.6 & 29.3 \\
  \bottomrule
  \end{tabular}}
\end{table}

\begin{figure*}[ht!]
  \centering
\includegraphics[width=0.88\linewidth]{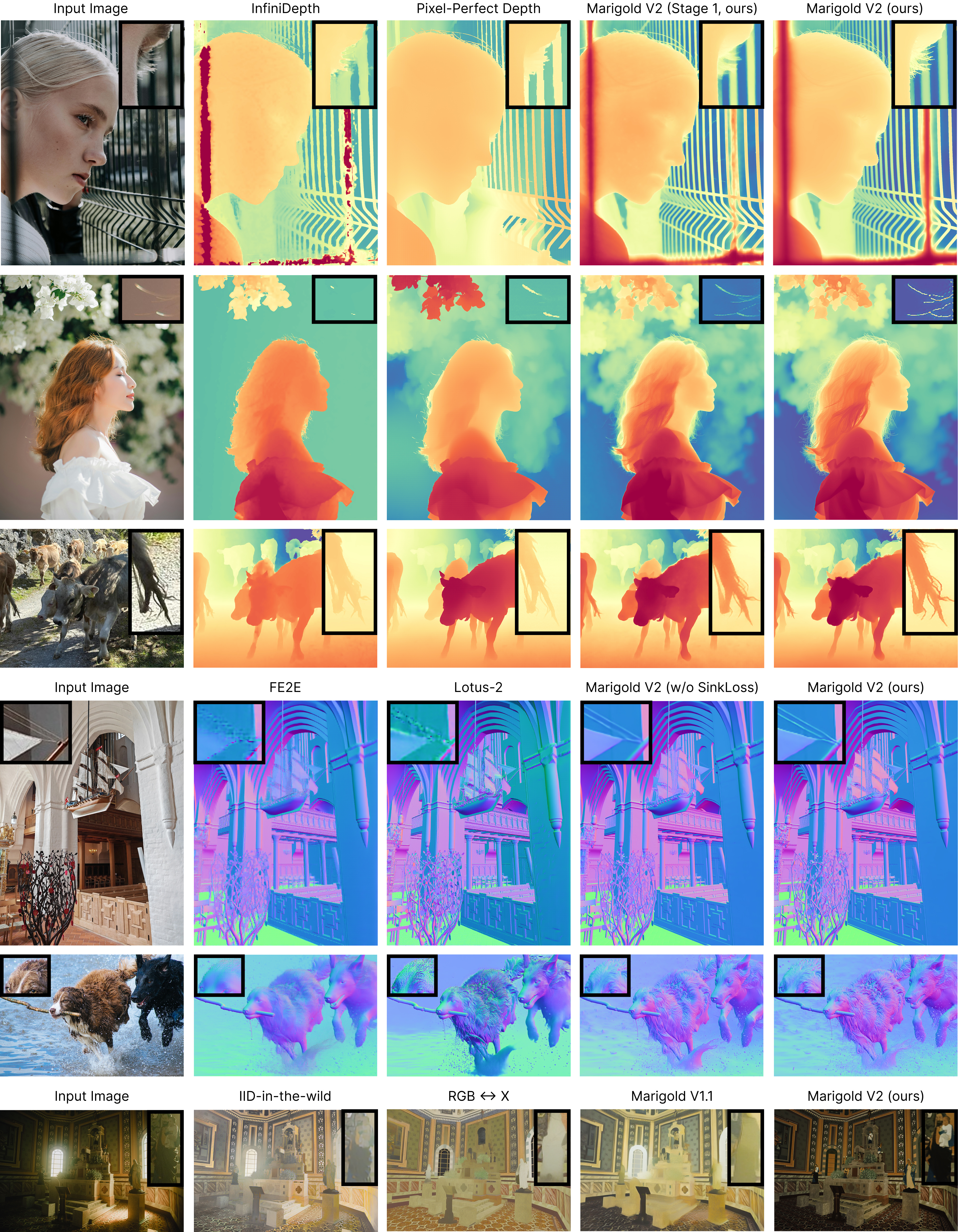}

    \caption{\textbf{Qualitative comparison on in-the-wild images.} \methodname preserves fine details better than the baselines across depth, surface normals, and albedo estimation. Rows 1, 2, and 6 are courtesy of cottonbro studio, VANNGO Ng, and Lucas Oliveira via Pexels, respectively. 
    } 
  \label{fig:appendix:qualitative_comparison:b}
  \Description[Qualitative comparison grid]{Qualitative comparison grid showing relative depth estimation across five diverse samples, comparing the proposed model against Infinidepth, Pixel-Perfect-Depth, and its own Stage 1 baseline.}
\end{figure*}

\begin{table*}[ht]
    \centering
    \caption{\textbf{Zero-shot metric depth completion evaluation}
    using
    the sparse-guidance protocol of Marigold-DC~\cite{marigolddc} at native ground-truth resolution. NYUv2 and iBims are evaluated on their full evaluation sets; KITTI-DC and DDAD use evenly-sampled subsets ($n{=}150$). 
     \textbf{Bold} indicates the best result and \underline{underlining} the second-best result for each metric.
    }
    \label{table:dc}
    \resizebox{\textwidth}{!}{%
    \begin{tabular}{@{}lcccccccccccccccc}
    \toprule
    & \multicolumn{4}{c}{iBims-1} & \multicolumn{4}{c}{NYUv2} & \multicolumn{4}{c}{KITTI-DC ($n{=}150$)} & \multicolumn{4}{c}{DDAD ($n{=}150$)} \\
    \cmidrule(lr){2-5}\cmidrule(lr){6-9}\cmidrule(lr){10-13}\cmidrule(lr){14-17}
    Method & MAE & RMSE & AbsRel & $\delta_1$ & MAE & RMSE & AbsRel & $\delta_1$ & MAE & RMSE & AbsRel & $\delta_1$ & MAE & RMSE & AbsRel & $\delta_1$ \\
    \midrule
    \textbf{Marigold-DC} {\scriptsize (ICCV 2025)~\cite{marigolddc}} & 0.059 & 0.189 & 0.016 & 0.989 & 0.061 & 0.152 & \cellthird{0.021} & 0.987 & 0.598 & 1.729 & 0.034 & 0.989 & 3.247 & 8.236 & 0.120 & 0.884 \\
    \textbf{Marigold-SSD} {\scriptsize (CVPRW 2026)~\cite{marigoldssd}} & 0.060 & 0.185 & 0.016 & 0.990 & 0.069 & 0.162 & 0.025 & 0.986 & 0.456 & 1.527 & 0.026 & \cellthird{0.992} & 2.080 & 6.585 & 0.072 & 0.949 \\
    \textbf{CAPA} {\scriptsize (arXiv 2026)~\cite{capa}} & \cellfirst{\textbf{0.030}} & \cellthird{0.135} & \cellfirst{\textbf{0.008}} & \cellsecond{\underline{0.994}} & \cellfirst{\textbf{0.044}} & \cellthird{0.117} & \cellfirst{\textbf{0.015}} & \cellthird{0.992} & 0.341 & \cellsecond{\underline{1.382}} & \cellsecond{\underline{0.016}} & \cellfirst{\textbf{0.994}} & \cellthird{1.294} & \cellsecond{\underline{6.007}} & \cellsecond{\underline{0.032}} & \cellsecond{\underline{0.978}} \\
    \textbf{LDCM} {\scriptsize (ICLR 2026)~\cite{LDCM}} & \cellthird{0.038} & 0.152 & \cellthird{0.010} & \cellthird{0.993} & \cellthird{0.048} & 0.126 & \cellsecond{\underline{0.016}} & 0.991 & \cellsecond{\underline{0.324}} & 1.452 & \cellfirst{\textbf{0.015}} & \cellsecond{\underline{0.993}} & \cellfirst{\textbf{1.097}} & \cellthird{6.012} & \cellfirst{\textbf{0.022}} & \cellfirst{\textbf{0.981}} \\
    \midrule
    \textbf{\methodname (test-time LoRA, ours)} & 0.042 & 0.160 & 0.012 & \cellthird{0.993} & \cellsecond{\underline{0.045}} & \cellsecond{\underline{0.112}} & \cellfirst{\textbf{0.015}} & \cellsecond{\underline{0.993}} & 0.349 & 1.472 & \cellthird{0.017} & \cellfirst{\textbf{0.994}} & 1.549 & 6.242 & 0.046 & 0.972 \\
    \textbf{\quad + high-resolution inference} & \cellsecond{\underline{0.034}} & \cellsecond{\underline{0.127}} & \cellthird{0.010} & \cellfirst{\textbf{0.995}} & \cellfirst{\textbf{0.044}} & \cellfirst{\textbf{0.110}} & \cellfirst{\textbf{0.015}} & \cellfirst{\textbf{0.994}} & \cellthird{0.340} & \cellthird{1.397} & \cellthird{0.017} & \cellfirst{\textbf{0.994}} & 1.465 & 6.072 & 0.043 & 0.974 \\
    \textbf{\quad + tiled local adaptation} & \cellfirst{\textbf{0.030}} & \cellfirst{\textbf{0.122}} & \cellsecond{\underline{0.009}} & \cellfirst{\textbf{0.995}} & \cellfirst{\textbf{0.044}} & \cellfirst{\textbf{0.110}} & \cellfirst{\textbf{0.015}} & \cellfirst{\textbf{0.994}} & \cellfirst{\textbf{0.318}} & \cellfirst{\textbf{1.333}} & \cellsecond{\underline{0.016}} & \cellfirst{\textbf{0.994}} & \cellsecond{\underline{1.226}} & \cellfirst{\textbf{5.323}} & \cellthird{0.038} & \cellthird{0.976} \\
    \bottomrule
    \end{tabular}%
    }
\end{table*}

\begin{table}[t]
    \centering
    \caption{\textbf{See-Through depth estimation.} Evaluation on the full LayeredDepth-Syn $l8$ validation set.}
    \label{tab:layereddepth-l8-results}
    \resizebox{0.7\linewidth}{!}{
    \begin{tabular}{@{}lcc@{}}
        \toprule
        Checkpoint & AbsRel $\downarrow$ & $\delta_1$ $\uparrow$ \\
        \midrule
        \methodname (depth, ours)          & 13.66 & 83.96 \\
        \methodname (see-through depth, ours) & \textbf{8.17} & \textbf{92.65} \\
        \bottomrule
    \end{tabular}
    }
\end{table}

\begin{figure}[t]
  \centering
  \makebox[\linewidth][l]{%
    \includegraphics[width=0.32666\linewidth]{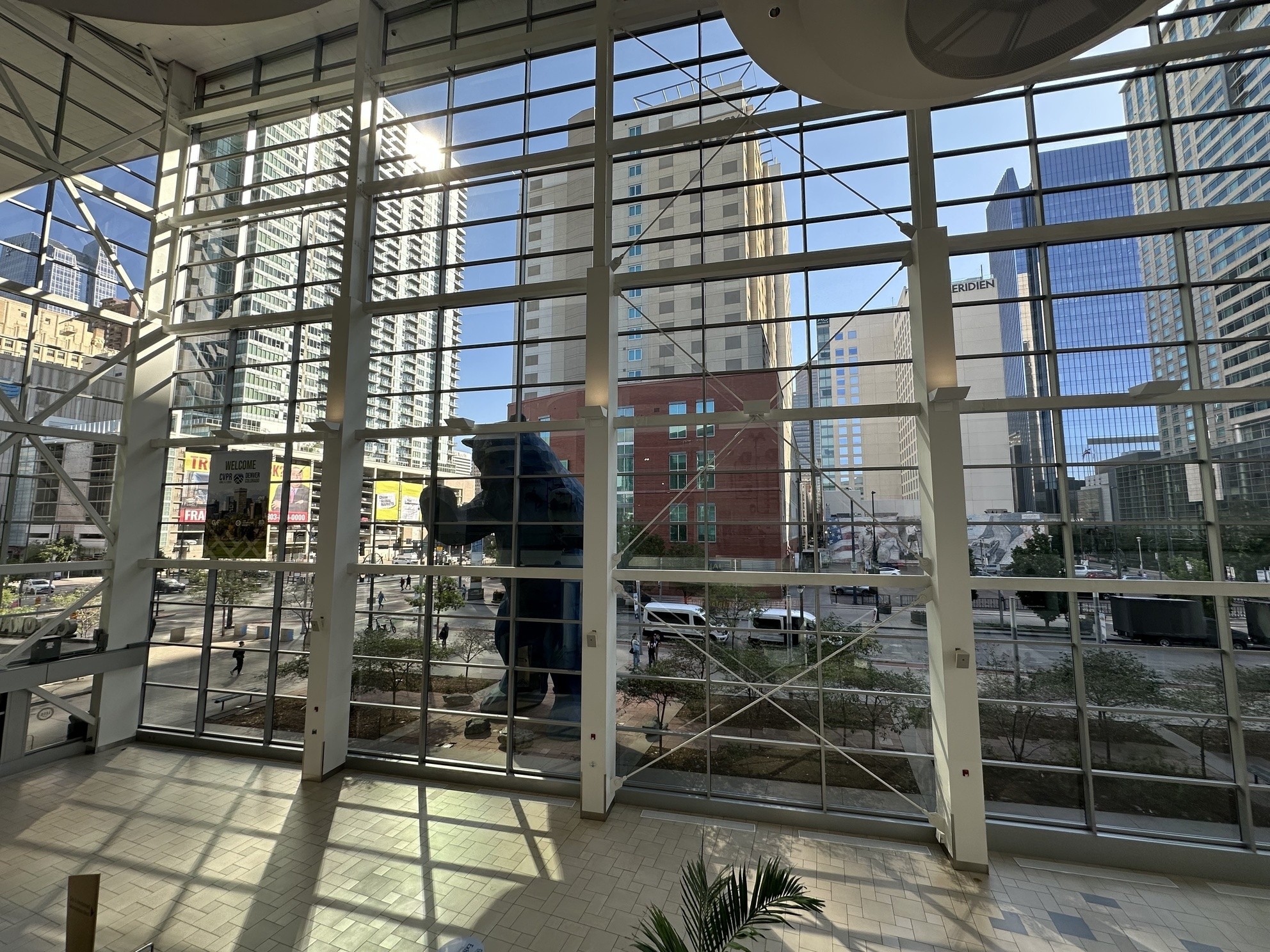}%
    \hspace{0.01\linewidth}%
    \includegraphics[width=0.32666\linewidth]{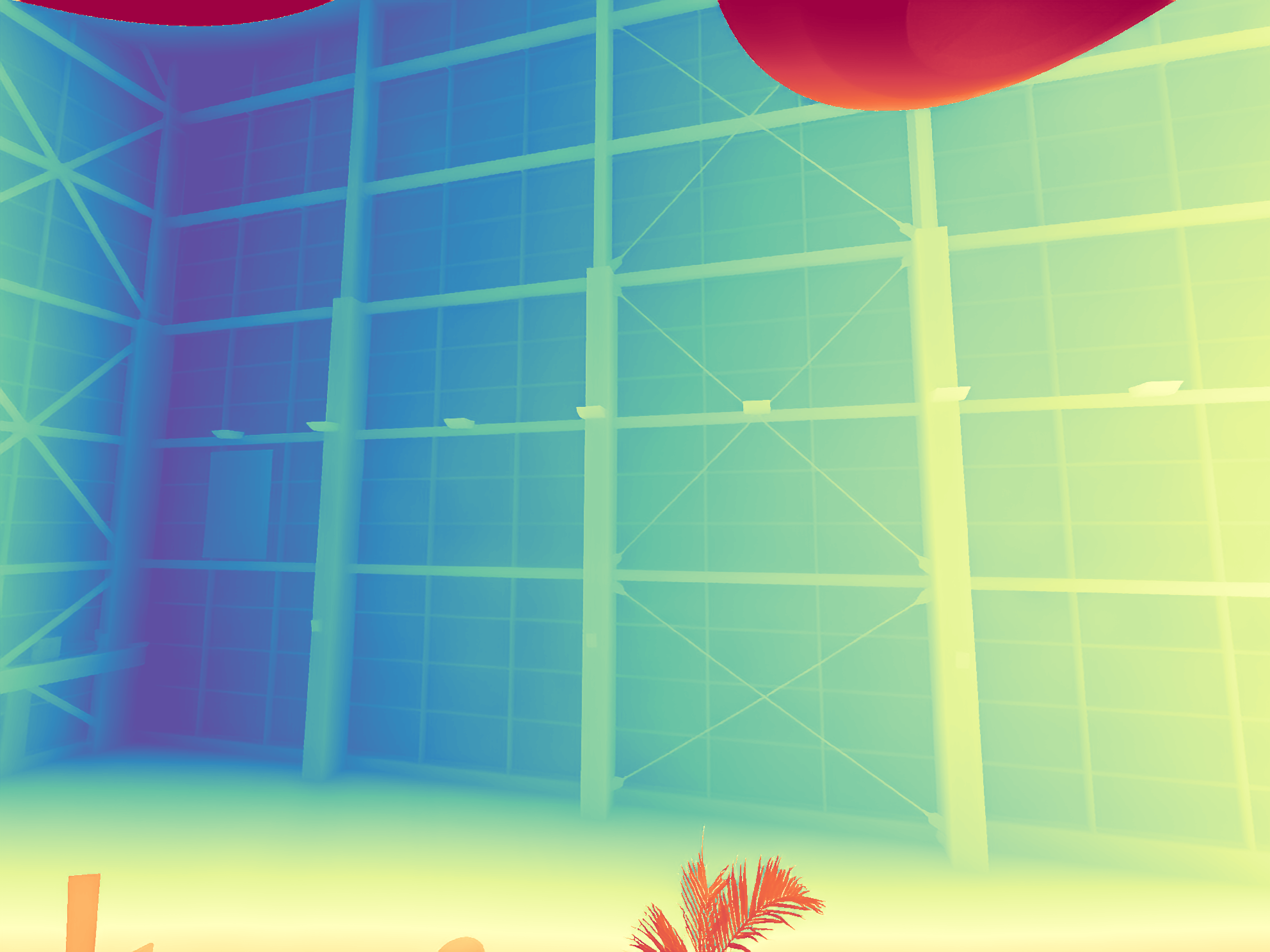}%
    \hspace{0.01\linewidth}%
    \includegraphics[width=0.32666\linewidth]{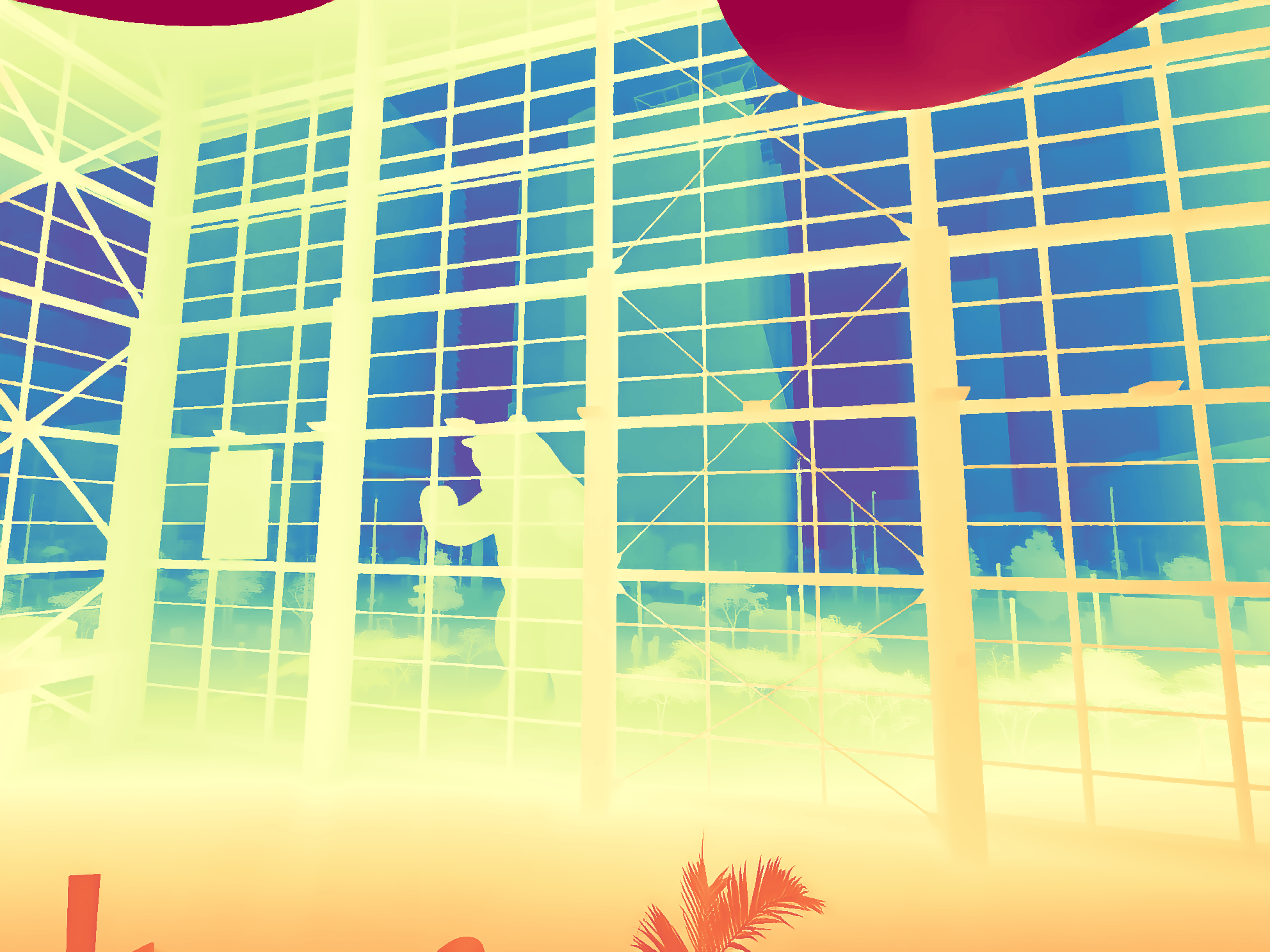}%
  }\\[0.01\linewidth]
  \makebox[\linewidth][l]{%
    \includegraphics[width=0.32666\linewidth]{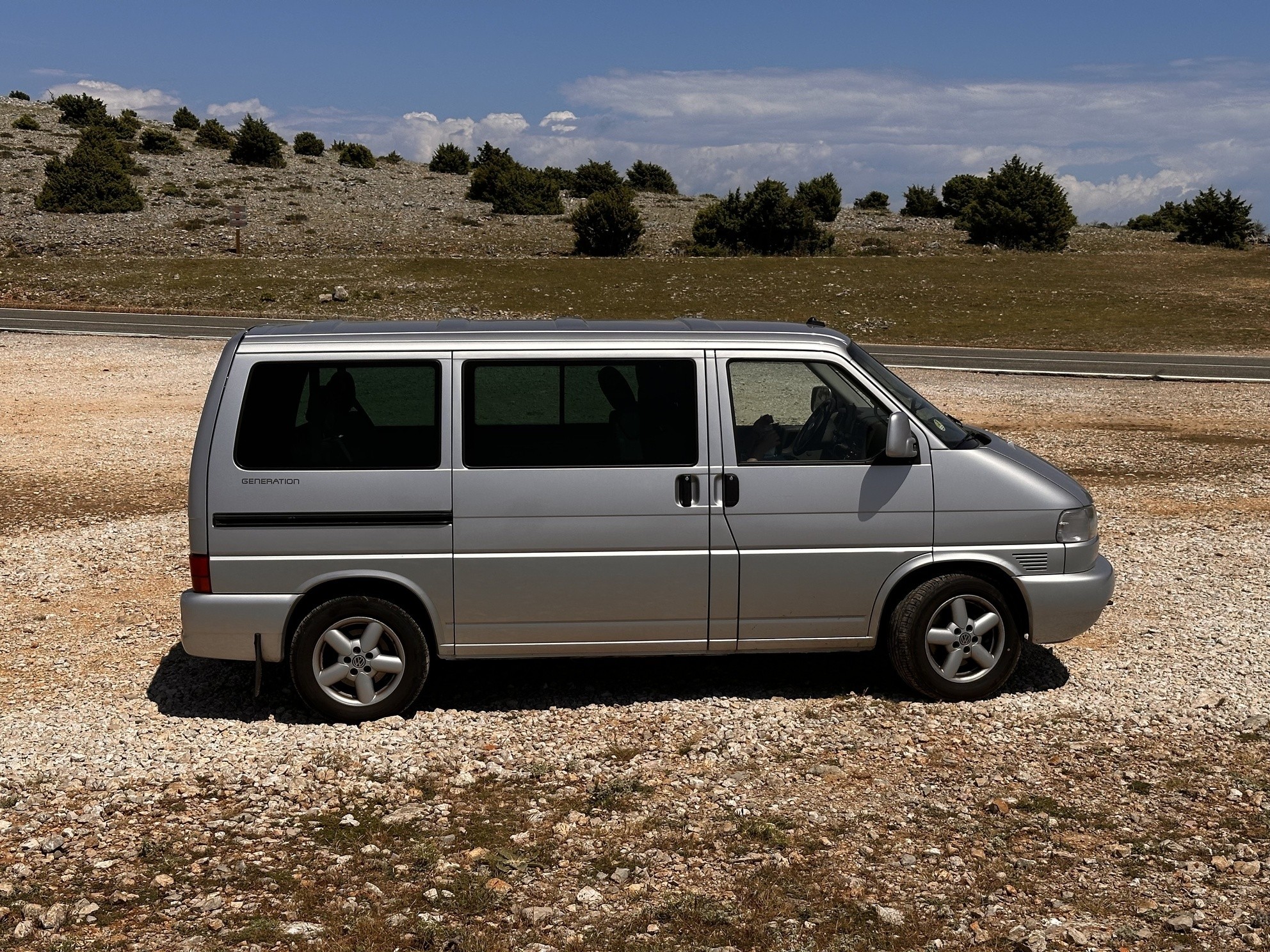}%
    \hspace{0.01\linewidth}%
    \includegraphics[width=0.32666\linewidth]{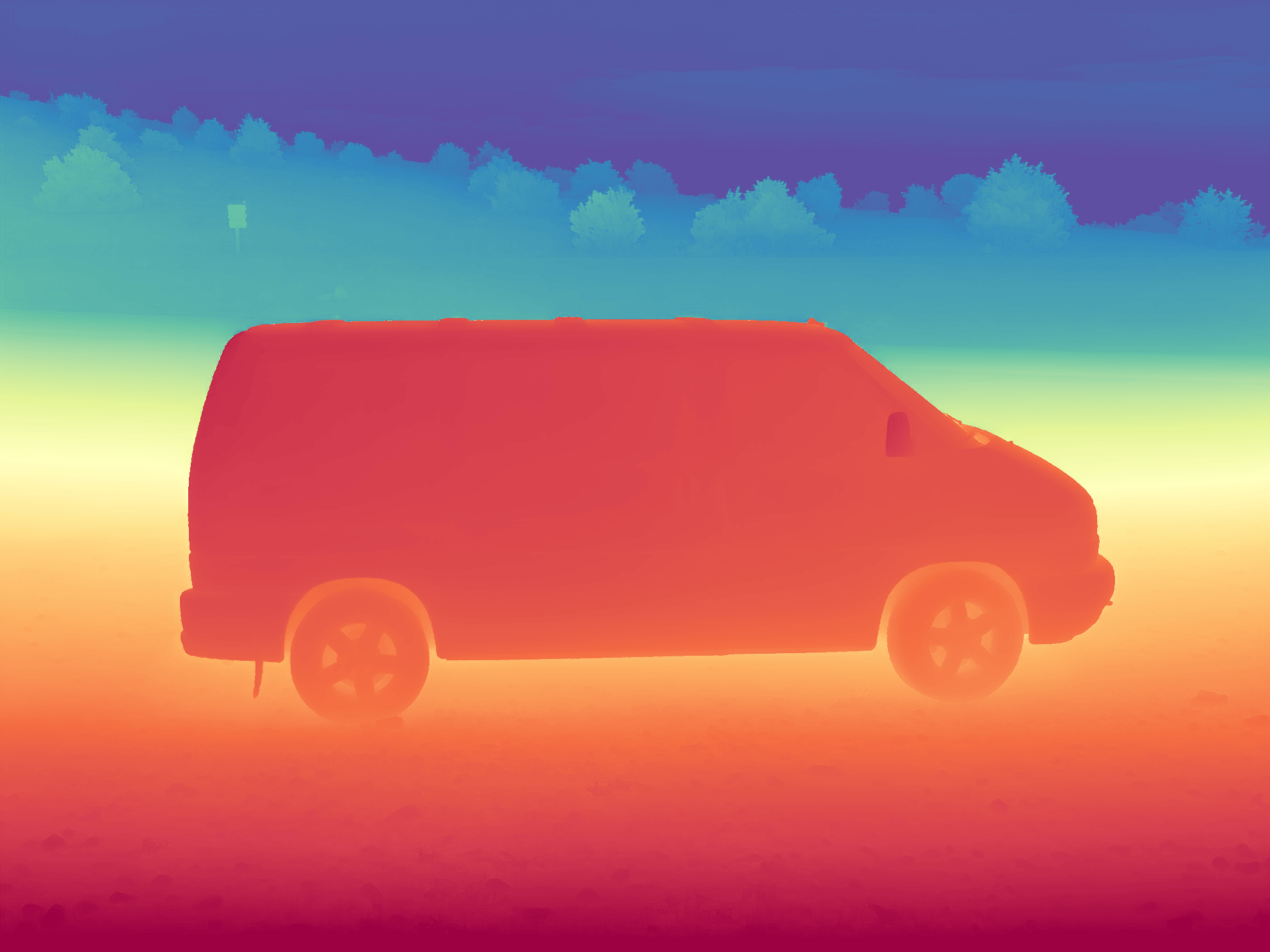}%
    \hspace{0.01\linewidth}%
    \includegraphics[width=0.32666\linewidth]{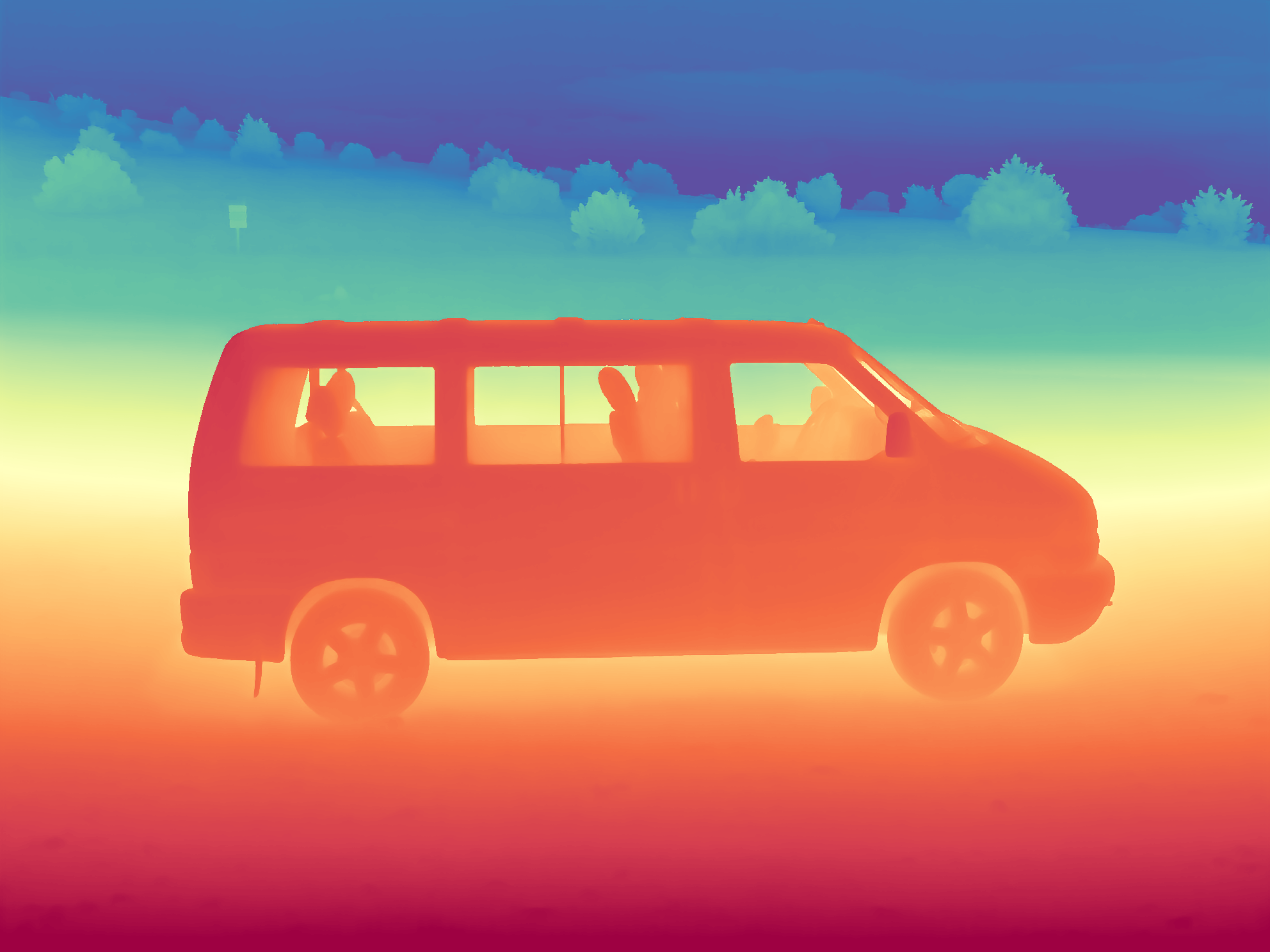}%
  }
  \caption{\textbf{See-through depth qualitative comparison.} From left to right: input RGB, predictions of our base model and of our see-through model.}
  \label{fig:layered_depth_qualitative}
  \Description[See-through depth qualitative comparison]{A two-by-three grid comparing RGB inputs, base-model depth predictions, and see-through-model depth predictions for a van and an atrium. The fine-tuned predictions recover geometry visible behind transparent surfaces.}
\end{figure}

\textbf{Stage-2 Ablation.} 
Finally, we ablate the main components of the second-stage fine-tuning procedure, which is designed to improve boundary quality and reduce flying-pixel artifacts. For these experiments, we use the Soft Edge Error (SEE$_k$) metric on the HyperSim dataset and relate it to the AbsRel across datasets (Fig.~\ref{fig:stage2_abalation}). It shows that Stage-2 SinkLoss fine-tuning substantially improves SEE$_3$, indicating sharper boundary reconstruction, while preserving standard metrics comparable to the original Stage-1 checkpoint.

Compared to the Stage-1 prediction, the SinkLoss-refined model produces fewer flying pixels around object boundaries and high-frequency regions, particularly near thin structures and vegetation. 
The experiment with disabling SinkLoss during VAE decoder finetuning in Stage 2 confirms its importance for recognition of thin details (Fig.~\ref{fig:sinkloss_impact}). %
The resulting depth maps contain cleaner discontinuities and more coherent local detail, showing that the second stage improves boundary fidelity.

\textbf{Latency and Memory Requirements.} We additionally measure inference latency and peak GPU memory for all models using the same protocol on a single 32GB GPU.

\begin{table*}[t]
  \caption{\textbf{Comparison of zero-shot surface normal estimators on NYUv2, ScanNet, iBims-1 and Sintel.} 
  We report MeanErr ($\downarrow$) and $11.25^\circ$ ($\uparrow$). Methods trained on more than 5M images are shown in gray. \textbf{Bold} indicates the best result and \underline{underlining} the second-best result for each metric. %
  }
  \label{tab:normal}
  \centering
  \resizebox{1.0\linewidth}{!}
    {
  \begin{tabular}{@{}lccccccccc}
  \toprule
  \multirow{2}{*}{Method}
  & Training
  & \multicolumn{2}{c}{NYUv2 (Indoor)} & \multicolumn{2}{c}{ScanNet (Indoor)} 
  & \multicolumn{2}{c}{iBims-1 (Indoor)} & \multicolumn{2}{c}{Sintel (Outdoor)}
  
  \\
   \cmidrule(lr){3-4} \cmidrule(lr){5-6} \cmidrule(lr){7-8}\cmidrule(lr){9-10}
   &Data$\downarrow$
   & MeanErr$\downarrow$ & $11.25^\circ$$\uparrow$  
   & MeanErr$\downarrow$ & $11.25^\circ$$\uparrow$ 
   & MeanErr$\downarrow$ & $11.25^\circ$$\uparrow$ 
   & MeanErr$\downarrow$ & $11.25^\circ$$\uparrow$ 
  
   \\
  \midrule

\nocompete{\textbf{MoGe-2} \scriptsize{ (NeurIPS 2025)~\cite{wang2025moge2}}} & \nocompete{8.9M} & \nocompete{14.7} & \nocompete{62.3} & \nocompete{12.8} & \nocompete{68.4} & \nocompete{14.7} & \nocompete{70.4} & \nocompete{29.3} & \nocompete{24.8} \\
\midrule
\textbf{DSINE} \scriptsize{ (CVPR 2024)~\cite{bae2024dsine}} & 160K & \cellthird{16.4} & 59.6 & 16.2 & 61.0 & 17.1 & 67.4 & 34.9 & 21.5 \\
\textbf{GeoWizard} \scriptsize{ (ECCV 2024)~\cite{fu2024geowizard}} & 280K & 18.9 & 50.7 & 17.4 & 53.8 & 19.3 & 63.0 & 40.3 & 12.3 \\
\textbf{StableNormal} \scriptsize{ (SIGGRAPH Asia 2024)~\cite{ye2024stablenormal}} & 250K & 18.6 & 53.5 & 17.1 & 57.4 & 18.2 & 65.0 & 36.7 & 14.1 \\
\textbf{Diffusion-E2E-FT} \scriptsize{ (WACV 2025)~\cite{garcia2025fine}} & 74K & 16.5 & \cellthird{60.4} & {14.7} & 66.1 & 16.1 & 69.7 & 33.5 & \cellthird{22.3} \\
\textbf{GenPercept} \scriptsize{ (ICLR 2025)~\cite{xu2025matters}} & 74K & 18.2 & 56.3 & 17.7 & 58.3 & 18.2 & 64.0 & 37.6 & 16.2 \\
\textbf{Lotus-G} \scriptsize{ (ICLR 2025)~\cite{he2025lotus1}} & 59K & 16.5 & 59.4 & 15.1 & 63.9 & 17.2 & 66.2 & 33.6 & 21.0 \\
\textbf{Lotus-D} \scriptsize{ (ICLR 2025)~\cite{he2025lotus1}} & 59K & \cellsecond{\underline{16.2}} & 59.8 & 14.7 & 64.0 & 17.1 & 66.4 & 32.3 & \cellsecond{\underline{22.4}} \\
\textbf{Marigold Normals V1.1} \scriptsize{ (TPAMI 2025)~\cite{ke2024repurposing,ke2025marigold}} & 77K & \cellfirst{\textbf{{16.1}}} & \cellsecond{\underline{60.5}} & 14.5 & 66.1 & 16.3 & 68.5 & - & - \\
\textbf{Lotus-2} \scriptsize{ (arXiv 2025)~\cite{he2025lotus2}} & 59K & 16.9 & 59.0 & \cellthird{14.2} & \cellthird{66.8} & \cellsecond{\underline{15.4}} & \cellthird{70.4} & \cellsecond{\underline{30.3}} & \cellfirst{\textbf{27.6}} \\
\textbf{FE2E} \scriptsize{ (CVPR 2026)~\cite{wang2026fe2e}} & 71K & \cellsecond{\underline{16.2}} & 59.6 & \cellfirst{\textbf{13.8}} & \cellsecond{\underline{67.2}} & \cellfirst{\textbf{15.1}} & \cellsecond{\underline{70.6}} & \cellthird{31.2} & \cellthird{22.3} \\
\midrule
 \textbf{\methodname (normals, ours)} & 59K & 16.6 & \cellfirst{\textbf{61.2}} & \cellsecond{\underline{14.1}} & \cellfirst{\textbf{67.4}} & \cellthird{15.9} & \cellfirst{\textbf{70.9}} & \cellfirst{\textbf{28.7}} & \cellfirst{\textbf{27.6}} \\
 \bottomrule
  \end{tabular}
    }
\end{table*}

The results are reported in Tab.~\ref{tab:latency_comparison}: compared with diffusion-based methods, \methodname provides a favorable balance between latency, memory, and resolution. Its QLoRA-quantized base weights and single-step formulation avoid the iterative sampling and extra input tokens used by many diffusion models. At $1024 \times 1024$, it is faster than Lotus-2 and FE2E, although slower than Lotus-2 without the detail sharpener, and uses less memory than both. At $2048 \times 2048$, it remains feasible on a single 32GB GPU, while Pixel-Perfect Depth, Lotus-2, and FE2E run out of memory. \methodname is therefore not the fastest model overall, but it offers substantially better resolution scalability than prior diffusion-based estimators.

\section{Other Dense Regression Tasks}

\subsection{Depth Completion}\label{sec:dc}

We address metric depth completion by fitting a new LoRA to a frozen affine-invariant depth prior, supervised at test time by the sparse depth measurements. Marigold-SSD~\cite{marigoldssd} adapts in weight space but trains offline; Marigold-DC~\cite{marigolddc} stays at test time but needs $50$-step latent guidance; CAPA~\cite{capa} does both, but on the metric-native MoGe-2.

We freeze everything and attach a second, zero-initialized \mbox{rank-16} LoRA to the last $12$ out of the $60$ transformer blocks.
In each forward pass, the model predicts affine-invariant log-depth, which is mapped to metric depth by two learned scalars $a$ and $b$ (scale and shift), initialized by least squares against the sparse measurements.
We minimize a combined $L_1$ and $L_2$ residual at the sparse measurements for $100$ Adam iterations, with learning rates of $10^{-3}$ for the LoRA and $3{\times}10^{-2}$ for $(a, b)$. 
This yields $21.2$M trainable parameters in the adapter, plus the two scalars. 
We leverage \methodname ability to run inference at higher resolutions than at training, and run it on a Lanczos-upscaled input, while keeping supervision and evaluation on the native grid: the sparse points are never resampled, and the prediction is bilinearly resized back to native resolution before applying the loss. Tiling raises the effective resolution further and additionally reduces the metric depth range each affine fit must cover. Concretely, we use $3{\times}3$ tiles at $4{\times}$ the tile resolution with $15\%$ overlap for iBims-1, and $2{\times}2$ tiles with $25\%$ overlap at $1.5{\times}$ and $2{\times}$ for KITTI-DC~\cite{uhrig2017sparsity} and DDAD~\cite{guizilini20203dpacking}.
NYUv2 is left untiled because its $500$ measurements are too sparse to subdivide, leaving each tile without enough anchors for a stable affine fit. Tiling costs $4$--$14{\times}$ the single-pass runtime. 
Despite its purely affine-invariant prior, it outperforms both Marigold-SSD~\cite{marigoldssd} and Marigold-DC~\cite{marigolddc} by a wide margin and achieves the lowest RMSE on all four benchmarks (Tab.~\ref{table:dc}).

\subsection{See-Through Depth}\label{sec:seethrough} 

We additionally investigate see-through depth estimation, where the goal is to predict the geometry visible behind transparent surfaces such as glass. We observed that the base \methodname model is biased toward closer depth values in such regions: this bias comes from annotations in HyperSim, where transparent surfaces are annotated at the glass surface rather than at the geometry behind it. Depending on the final application, we may want to learn to estimate depth behind such surfaces: to this end, we additionally fine-tune our model for 30K steps using the final depth layer, $l8$, of LayeredDepth-Syn~\cite{wen2025seeing}. As shown in Tab.~\ref{tab:layereddepth-l8-results}, this reduces AbsRel from $13.7$ to $8.2$ and improves $\delta_1$ from $84.0$ to $92.7$ on the validation split, demonstrating more accurate depth prediction behind glass. The qualitative improvement is shown in Fig.~\ref{fig:layered_depth_qualitative}.

\subsection{Surface Normal Estimation}\label{sec:normals}
\begin{figure}[t]
  \centering
  \makebox[\linewidth][l]{%
    \includegraphics[width=0.32000\linewidth]{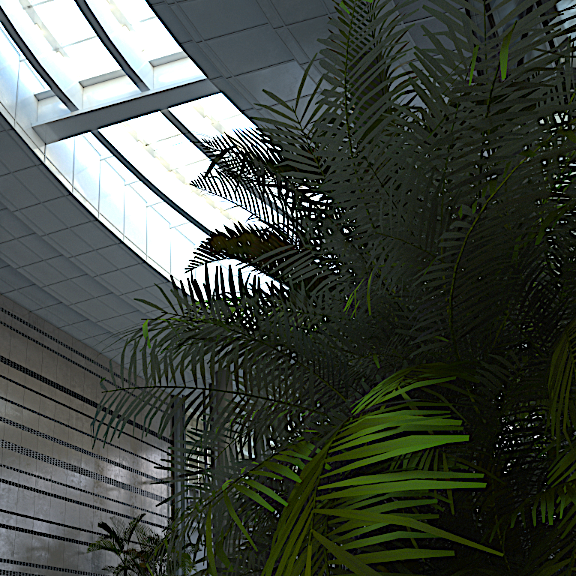}%
    \hspace{0.01\linewidth}%
    \includegraphics[width=0.32000\linewidth]{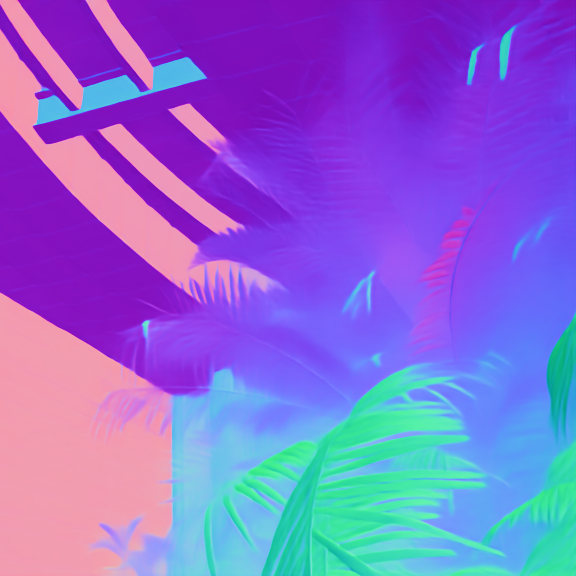}%
    \hspace{0.03\linewidth}%
    \includegraphics[width=0.32000\linewidth]{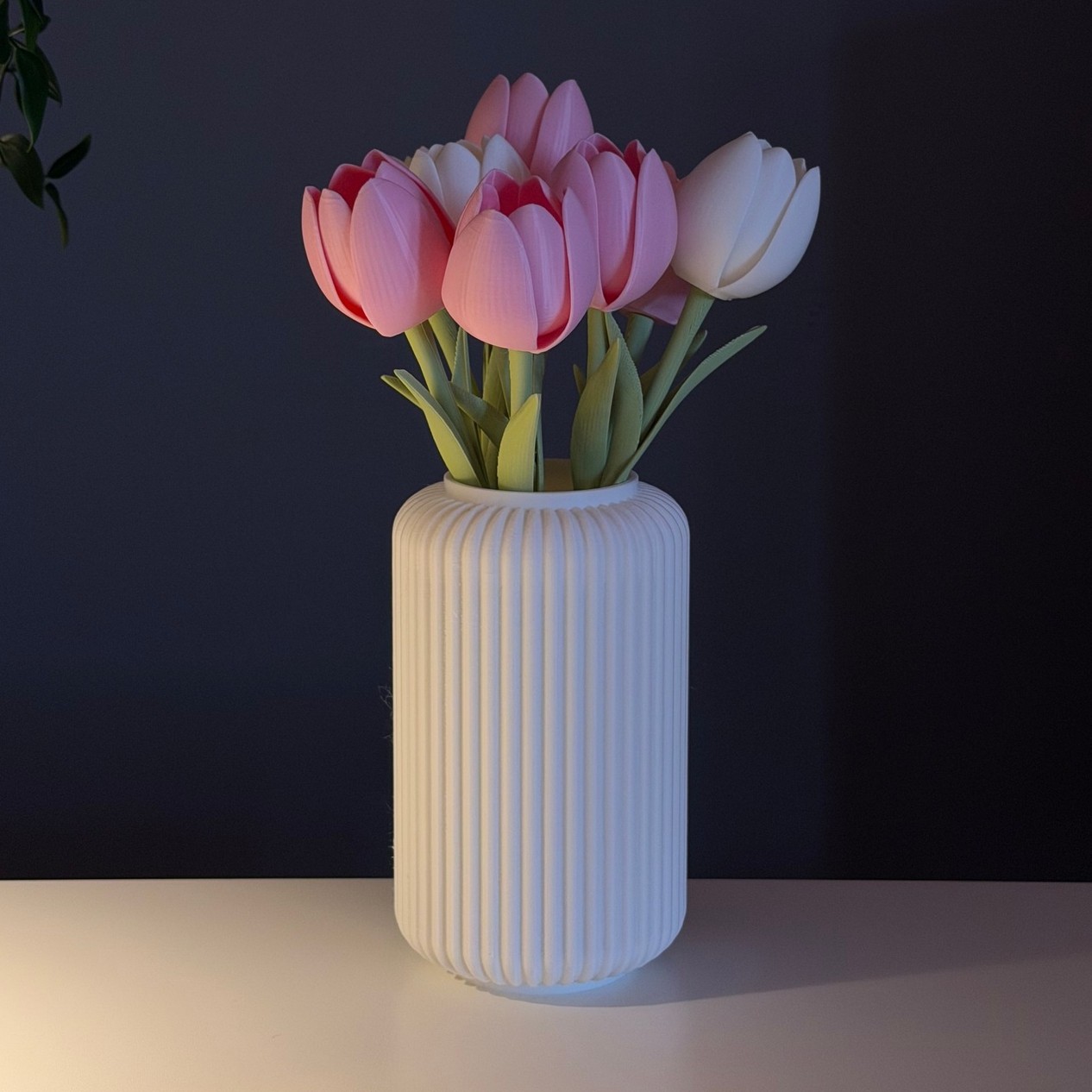}%
  }\\[1pt]
  \makebox[\linewidth][l]{%
    \includegraphics[width=0.32000\linewidth]{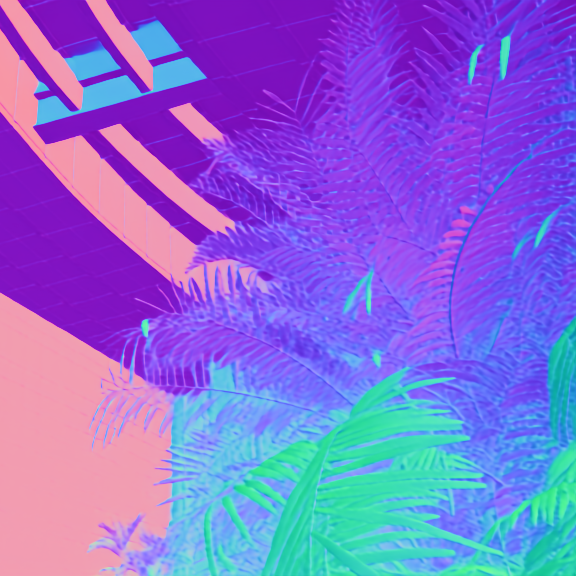}%
    \hspace{0.01\linewidth}%
    \includegraphics[width=0.32000\linewidth]{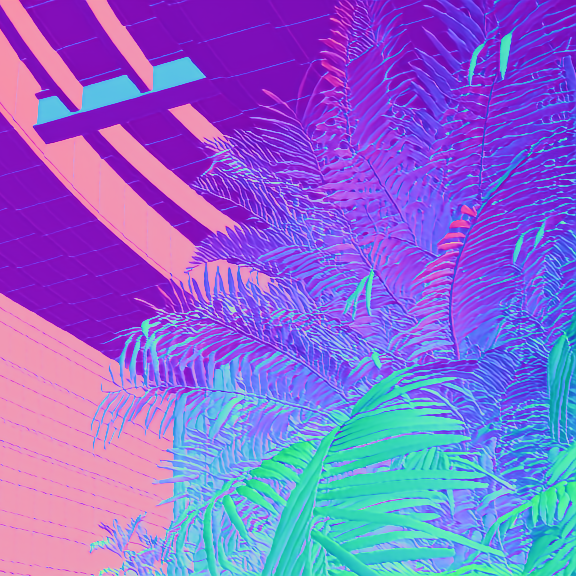}%
    \hspace{0.03\linewidth}%
    \includegraphics[width=0.32000\linewidth]{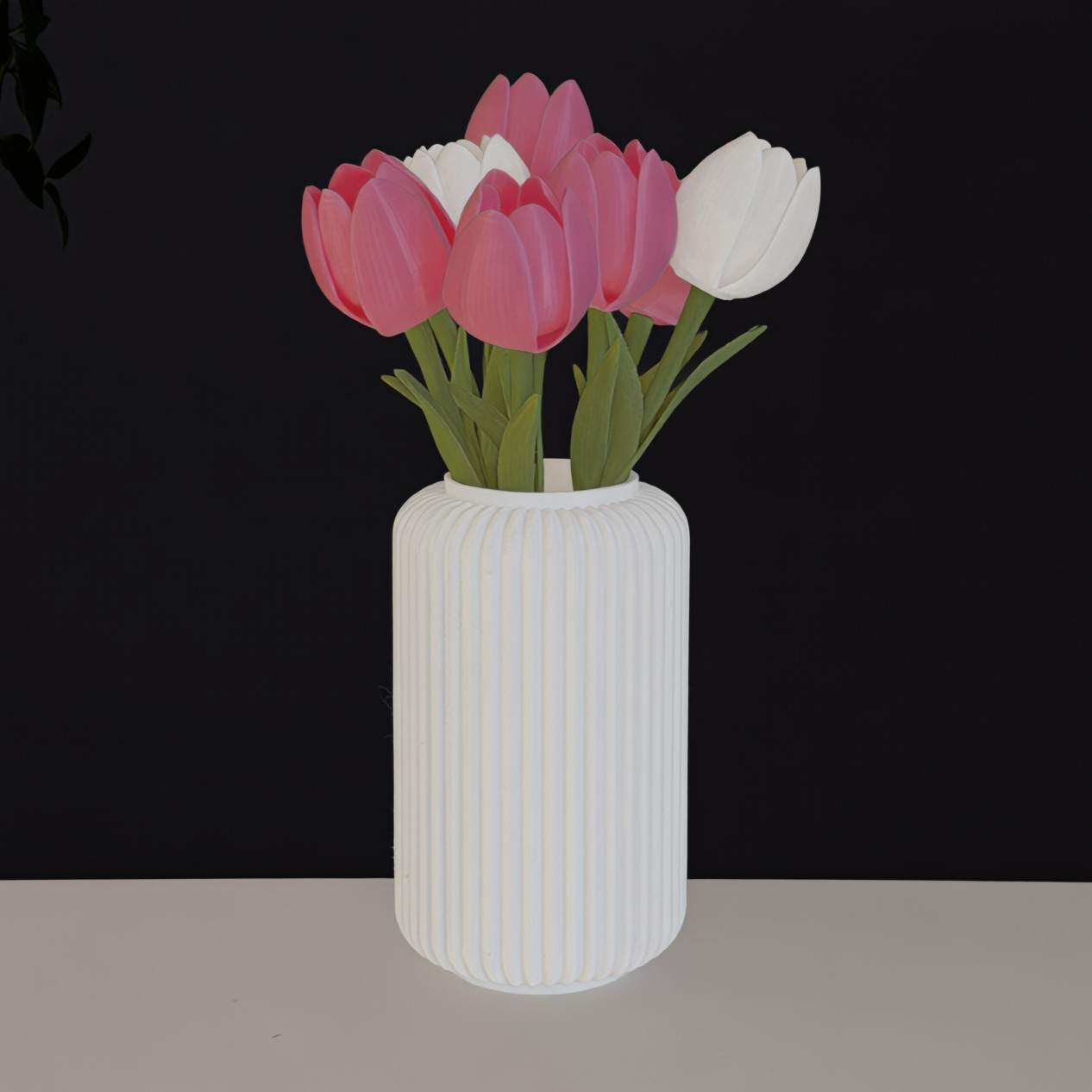}%
  }
  \raggedright
  \caption{\textbf{Qualitative results: surface normals and albedo.}
  Left: Input, angular $\mathcal{L}_1$, $\mathcal{L}_1$ + iREPA, and SinkLoss + iREPA. Right: Input and albedo prediction. Flowers image is courtesy of Rüveyda Akkaya via Pexels.
  }
  \label{fig:normals_qualitative}
  \Description[Qualitative surface-normal and albedo comparison]{A two-by-three grid showing four surface-normal panels in the first two columns and the RGB input with its corresponding albedo prediction in the last column.}
\end{figure}

Tab.~\ref{tab:normal} compares \methodname with recent discriminative and diffu\-sion-based baselines on NYUv2~\cite{silberman2012nyu}, ScanNet~\cite{dai2017scannet}, iBims-1~\cite{koch2018evaluation}, and Sintel~\cite{butler2012naturalistic}. 
We retrain \methodname by replacing the pixel-space term of Eq.~\ref{eq:stage1} with an angular loss between predicted and ground-truth normals, keeping iREPA, and adding SinkLoss on top.
Sinkhorn transport weights are computed from the $\mathcal{L}_1$ distance between predictions and ground truth.
Fig.~\ref{fig:normals_qualitative} (left) and the middle rows of Fig.~\ref{fig:appendix:qualitative_comparison:b} illustrate the effect of the proposed losses, highlighting the much higher fidelity of details enabled by the joint use of iREPA and SinkLoss.

To measure the impact of this latter loss quantitatively, we adapt the soft edge error from depth to normals. The resulting Soft Angular Edge Error (SAEE) measures angular discrepancies near HyperSim boundaries while allowing predictions to match nearby ground-truth orientations rather than requiring exact pixel correspondence. Tab.~\ref{tab:hypersim_normals_see} shows that SinkLoss consistently improves SAEE.

\begin{table}[t]
  \centering
  \small
\caption{\textbf{Impact of SinkLoss on surface normal estimation.} Mean angular error (MeanErr) is averaged over four test datasets, while the Soft Angular Edge Error (SAEE) metrics are computed on the HyperSim test set.}
  \label{tab:hypersim_normals_see}
  \resizebox{0.7\columnwidth}{!}{%
  \begin{tabular}{@{}ccccc@{}}
  \toprule
  $\mathcal{L}_\mathrm{SinkLoss}$ & MeanErr $\downarrow$ & SAEE$_3$ $\downarrow$ & SAEE$_5$ $\downarrow$ & SAEE$_7$ $\downarrow$  \\
  \midrule
  \xmark & \bf 18.53 & 11.25 & 9.42 & 8.53 \\
  \cmark & 18.84 & \bf 10.76 & \bf 8.86 & \bf 7.99  \\
  \bottomrule
  \end{tabular}%
  }
\end{table}

\begin{table}[t]
\centering
\caption{\textbf{Albedo estimation on the HyperSim test set.} We report PSNR ($\uparrow$), SSIM ($\uparrow$), and LPIPS ($\downarrow$).}
\label{tab:albedo}
\resizebox{0.95\columnwidth}{!}{%
\begin{tabular}{
@{}
l@{\hspace{0.9em}}
c@{\hspace{0.6em}}
c@{\hspace{0.6em}}
c@{\hspace{0.6em}}
@{}
}

\toprule

\multirow{2}{*}{Method} & 
\multicolumn{3}{c}{Albedo}
\\ 

& 
PSNR $\uparrow$ & 
SSIM $\uparrow$ & 
LPIPS $\downarrow$\\

\midrule

\textbf{IID-in-the-wild} \scriptsize{(TOG 2024) ~\cite{careaga2024colorful}} & 
\underline{19.28} & 
\textbf{0.819} & 
0.260
\\ 

\textbf{RGB$\leftrightarrow$X} \scriptsize{(SIGGRAPH 2024) ~\cite{zeng2024rgb}} & 
17.43 & 
0.795 & 
\underline{0.200}
\\ 

\textbf{Marigold IID-Lighting V1.1} \scriptsize{(TPAMI 2025) \cite{ke2025marigold}} & 
18.21 & 
0.771 & 
0.218
\\

\midrule
\textbf{\methodname (albedo, ours)} &
 \textbf{20.78} &
 \underline{0.811} &
 \textbf{0.195}
\\  

\bottomrule

\end{tabular}}
\end{table}

\subsection{Albedo Estimation}\label{sec:albedo}

We additionally evaluate \methodname on single-image albedo estimation on the HyperSim test set. 
Starting from the base Qwen model, we fine-tune it on the HyperSim training split for 30K steps using an $\mathcal{L}_1$ reconstruction loss together with iREPA computed directly on the ground-truth albedo values. 
As shown in Tab.~\ref{tab:albedo}, the resulting model achieves the best PSNR and LPIPS among the compared methods, while remaining competitive in SSIM. 
A qualitative example is shown in Fig. \ref{fig:normals_qualitative} (right) and the last row of Fig.~\ref{fig:appendix:qualitative_comparison:b}.

\section{Conclusion}\label{sec:conclusion}

We presented \methodname, a practical, diffusion-based monocular depth estimator that jointly pursues quantitative accuracy and perceptual quality. 
Derived from Qwen-Image-Edit through a 2-stage fine-tuning process steered by iREPA regularization and the novel SinkLoss, our model establishes a new state of the art in monocular depth estimation while retaining the high-frequency, fine details that previous models tend to suppress. 
\methodname{} is thus an accessible tool in the hands of practitioners, unlocking high-quality computational photography applications on top of depth estimation.

Beyond relative depth, the same recipe extends to other dense regression tasks: metric depth completion, see-through depth, surface normals, and albedo estimation. 
The large image-editing backbone precludes real-time use; regions with reflections, motion, or defocus blur remain ambiguous, inviting uncertainty-aware or multi-layer extensions.  
More broadly, \methodname{} shows that state-of-the-art perception can be distilled
from open generative models in days, a recipe we hope others will build on.

\bibliographystyle{ACM-Reference-Format}
\bibliography{main}


\begin{thebibliography}{98}


\ifx \showCODEN    \undefined \def \showCODEN     #1{\unskip}     \fi
\ifx \showISBNx    \undefined \def \showISBNx     #1{\unskip}     \fi
\ifx \showISBNxiii \undefined \def \showISBNxiii  #1{\unskip}     \fi
\ifx \showISSN     \undefined \def \showISSN      #1{\unskip}     \fi
\ifx \showLCCN     \undefined \def \showLCCN      #1{\unskip}     \fi
\ifx \shownote     \undefined \def \shownote      #1{#1}          \fi
\ifx \showarticletitle \undefined \def \showarticletitle #1{#1}   \fi
\ifx \showURL      \undefined \def \showURL       {\relax}        \fi
\providecommand\bibfield[2]{#2}
\providecommand\bibinfo[2]{#2}
\providecommand\natexlab[1]{#1}
\providecommand\showeprint[2][]{arXiv:#2}

\bibitem[Bae and Davison(2024)]%
        {bae2024dsine}
\bibfield{author}{\bibinfo{person}{Gwangbin Bae} {and} \bibinfo{person}{Andrew~J. Davison}.} \bibinfo{year}{2024}\natexlab{}.
\newblock \showarticletitle{Rethinking Inductive Biases for Surface Normal Estimation}. In \bibinfo{booktitle}{\emph{IEEE/CVF Conference on Computer Vision and Pattern Recognition (CVPR)}}.
\newblock


\bibitem[{BFL.ai}(2024)]%
        {bfl2024flux}
\bibfield{author}{\bibinfo{person}{{BFL.ai}}.} \bibinfo{year}{2024}\natexlab{}.
\newblock \bibinfo{booktitle}{\emph{BFL.ai Announces the FLUX.1 Suite of Models}}.
\newblock
\urldef\tempurl%
\url{https://bfl.ai/announcements/24-08-01-bfl}
\showURL{%
\tempurl}


\bibitem[{BFL.ai}(2026)]%
        {bfl2024flux2}
\bibfield{author}{\bibinfo{person}{{BFL.ai}}.} \bibinfo{year}{2026}\natexlab{}.
\newblock \bibinfo{booktitle}{\emph{FLUX.2 [klein]: Towards Interactive Visual Intelligence}}.
\newblock
\urldef\tempurl%
\url{https://bfl.ai/blog/flux2-klein-towards-interactive-visual-intelligence}
\showURL{%
\tempurl}


\bibitem[Bochkovskii et~al\mbox{.}(2025)]%
        {depth-pro}
\bibfield{author}{\bibinfo{person}{Aleksei Bochkovskii}, \bibinfo{person}{Amaël Delaunoy}, \bibinfo{person}{Hugo Germain}, \bibinfo{person}{Marcel Santos}, \bibinfo{person}{Yichao Zhou}, \bibinfo{person}{Stephan~R. Richter}, {and} \bibinfo{person}{Vladlen Koltun}.} \bibinfo{year}{2025}\natexlab{}.
\newblock \showarticletitle{Depth Pro: Sharp Monocular Metric Depth in Less Than a Second}. In \bibinfo{booktitle}{\emph{ICLR}}.
\newblock
\urldef\tempurl%
\url{https://arxiv.org/abs/2410.02073}
\showURL{%
\tempurl}


\bibitem[Butler et~al\mbox{.}(2012)]%
        {butler2012naturalistic}
\bibfield{author}{\bibinfo{person}{Daniel~J. Butler}, \bibinfo{person}{Jonas Wulff}, \bibinfo{person}{Garrett~B. Stanley}, {and} \bibinfo{person}{Michael~J. Black}.} \bibinfo{year}{2012}\natexlab{}.
\newblock \showarticletitle{A naturalistic open source movie for optical flow evaluation}. In \bibinfo{booktitle}{\emph{Proceedings of the 12th European Conference on Computer Vision - Volume Part VI}} (Florence, Italy) \emph{(\bibinfo{series}{ECCV'12})}. \bibinfo{publisher}{Springer-Verlag}, \bibinfo{address}{Berlin, Heidelberg}, \bibinfo{pages}{611–625}.
\newblock
\showISBNx{9783642337826}
\href{https://doi.org/10.1007/978-3-642-33783-3_44}{doi:\nolinkurl{10.1007/978-3-642-33783-3_44}}


\bibitem[Careaga and Aksoy(2024)]%
        {careaga2024colorful}
\bibfield{author}{\bibinfo{person}{Chris Careaga} {and} \bibinfo{person}{Yağız Aksoy}.} \bibinfo{year}{2024}\natexlab{}.
\newblock \showarticletitle{Colorful Diffuse Intrinsic Image Decomposition in the Wild}.
\newblock \bibinfo{journal}{\emph{ACM Transactions on Graphics}} \bibinfo{volume}{43}, \bibinfo{number}{6} (\bibinfo{date}{Nov.} \bibinfo{year}{2024}), \bibinfo{pages}{1–12}.
\newblock
\showISSN{1557-7368}
\href{https://doi.org/10.1145/3687984}{doi:\nolinkurl{10.1145/3687984}}


\bibitem[Chen et~al\mbox{.}(2019)]%
        {chen2019over_smoothing}
\bibfield{author}{\bibinfo{person}{Chuangrong Chen}, \bibinfo{person}{Xiaozhi Chen}, {and} \bibinfo{person}{Hui Cheng}.} \bibinfo{year}{2019}\natexlab{}.
\newblock \showarticletitle{On the over-smoothing problem of cnn based disparity estimation}. In \bibinfo{booktitle}{\emph{Proceedings of the IEEE/CVF International Conference on Computer Vision}}. \bibinfo{pages}{8997--9005}.
\newblock


\bibitem[Chen et~al\mbox{.}(2024)]%
        {chen2024pixart}
\bibfield{author}{\bibinfo{person}{Junsong Chen}, \bibinfo{person}{Jincheng Yu}, \bibinfo{person}{Chongjian Ge}, \bibinfo{person}{Lewei Yao}, \bibinfo{person}{Enze Xie}, \bibinfo{person}{Zhongdao Wang}, \bibinfo{person}{James Kwok}, \bibinfo{person}{Ping Luo}, \bibinfo{person}{Huchuan Lu}, {and} \bibinfo{person}{Zhenguo Li}.} \bibinfo{year}{2024}\natexlab{}.
\newblock \showarticletitle{Pixart-\(\alpha\): Fast training of diffusion transformer for photorealistic text-to-image synthesis}. In \bibinfo{booktitle}{\emph{International conference on learning representations}}, Vol.~\bibinfo{volume}{2024}. \bibinfo{pages}{57611--57640}.
\newblock


\bibitem[Chen et~al\mbox{.}(2025)]%
        {video_depth_anything}
\bibfield{author}{\bibinfo{person}{Sili Chen}, \bibinfo{person}{Hengkai Guo}, \bibinfo{person}{Shengnan Zhu}, \bibinfo{person}{Feihu Zhang}, \bibinfo{person}{Zilong Huang}, \bibinfo{person}{Jiashi Feng}, {and} \bibinfo{person}{Bingyi Kang}.} \bibinfo{year}{2025}\natexlab{}.
\newblock \showarticletitle{Video Depth Anything: Consistent Depth Estimation for Super-Long Videos}. In \bibinfo{booktitle}{\emph{Proceedings of the IEEE/CVF Conference on Computer Vision and Pattern Recognition (CVPR)}}. \bibinfo{pages}{22831--22840}.
\newblock


\bibitem[Cuturi(2013)]%
        {cuturi2013sinkhorn}
\bibfield{author}{\bibinfo{person}{Marco Cuturi}.} \bibinfo{year}{2013}\natexlab{}.
\newblock \showarticletitle{Sinkhorn distances: Lightspeed computation of optimal transport}.
\newblock \bibinfo{journal}{\emph{Advances in neural information processing systems}}  \bibinfo{volume}{26} (\bibinfo{year}{2013}).
\newblock


\bibitem[Dai et~al\mbox{.}(2017)]%
        {dai2017scannet}
\bibfield{author}{\bibinfo{person}{Angela Dai}, \bibinfo{person}{Angel~X. Chang}, \bibinfo{person}{Manolis Savva}, \bibinfo{person}{Maciej Halber}, \bibinfo{person}{Thomas Funkhouser}, {and} \bibinfo{person}{Matthias Nie{\ss}ner}.} \bibinfo{year}{2017}\natexlab{}.
\newblock \showarticletitle{{ScanNet}: Richly-Annotated {3D} Reconstructions of Indoor Scenes}. In \bibinfo{booktitle}{\emph{Proceedings of the IEEE Conference on Computer Vision and Pattern Recognition}}. \bibinfo{pages}{2432--2443}.
\newblock


\bibitem[Deng et~al\mbox{.}(2022)]%
        {kangle2021dsnerf}
\bibfield{author}{\bibinfo{person}{Kangle Deng}, \bibinfo{person}{Andrew Liu}, \bibinfo{person}{Jun-Yan Zhu}, {and} \bibinfo{person}{Deva Ramanan}.} \bibinfo{year}{2022}\natexlab{}.
\newblock \showarticletitle{Depth-supervised {NeRF}: Fewer Views and Faster Training for Free}. In \bibinfo{booktitle}{\emph{Proceedings of the IEEE/CVF Conference on Computer Vision and Pattern Recognition (CVPR)}}.
\newblock


\bibitem[Dettmers et~al\mbox{.}(2023)]%
        {dettmers2023qlora}
\bibfield{author}{\bibinfo{person}{Tim Dettmers}, \bibinfo{person}{Artidoro Pagnoni}, \bibinfo{person}{Ari Holtzman}, {and} \bibinfo{person}{Luke Zettlemoyer}.} \bibinfo{year}{2023}\natexlab{}.
\newblock \showarticletitle{Qlora: Efficient finetuning of quantized llms}.
\newblock \bibinfo{journal}{\emph{Advances in neural information processing systems}}  \bibinfo{volume}{36} (\bibinfo{year}{2023}), \bibinfo{pages}{10088--10115}.
\newblock


\bibitem[Dosovitskiy et~al\mbox{.}(2021)]%
        {dosovitskiyimage}
\bibfield{author}{\bibinfo{person}{Alexey Dosovitskiy}, \bibinfo{person}{Lucas Beyer}, \bibinfo{person}{Alexander Kolesnikov}, \bibinfo{person}{Dirk Weissenborn}, \bibinfo{person}{Xiaohua Zhai}, \bibinfo{person}{Thomas Unterthiner}, \bibinfo{person}{Mostafa Dehghani}, \bibinfo{person}{Matthias Minderer}, \bibinfo{person}{Georg Heigold}, \bibinfo{person}{Sylvain Gelly}, {et~al\mbox{.}}} \bibinfo{year}{2021}\natexlab{}.
\newblock \showarticletitle{An Image is Worth 16x16 Words: Transformers for Image Recognition at Scale}. In \bibinfo{booktitle}{\emph{International Conference on Learning Representations}}.
\newblock


\bibitem[Eftekhar et~al\mbox{.}(2021)]%
        {eftekhar2021omnidata}
\bibfield{author}{\bibinfo{person}{Ainaz Eftekhar}, \bibinfo{person}{Alexander Sax}, \bibinfo{person}{Jitendra Malik}, {and} \bibinfo{person}{Amir Zamir}.} \bibinfo{year}{2021}\natexlab{}.
\newblock \showarticletitle{Omnidata: A scalable pipeline for making multi-task mid-level vision datasets from 3d scans}. In \bibinfo{booktitle}{\emph{Proceedings of the IEEE/CVF International Conference on Computer Vision}}. \bibinfo{pages}{10786--10796}.
\newblock


\bibitem[Eigen et~al\mbox{.}(2014)]%
        {eigen2014depth}
\bibfield{author}{\bibinfo{person}{David Eigen}, \bibinfo{person}{Christian Puhrsch}, {and} \bibinfo{person}{Rob Fergus}.} \bibinfo{year}{2014}\natexlab{}.
\newblock \showarticletitle{Depth map prediction from a single image using a multi-scale deep network}.
\newblock \bibinfo{journal}{\emph{Advances in neural information processing systems}}  \bibinfo{volume}{27} (\bibinfo{year}{2014}).
\newblock


\bibitem[Esser et~al\mbox{.}(2024)]%
        {esser2024scaling}
\bibfield{author}{\bibinfo{person}{Patrick Esser}, \bibinfo{person}{Sumith Kulal}, \bibinfo{person}{Andreas Blattmann}, \bibinfo{person}{Rahim Entezari}, \bibinfo{person}{Jonas M{\"u}ller}, \bibinfo{person}{Harry Saini}, \bibinfo{person}{Yam Levi}, \bibinfo{person}{Dominik Lorenz}, \bibinfo{person}{Axel Sauer}, \bibinfo{person}{Frederic Boesel}, {et~al\mbox{.}}} \bibinfo{year}{2024}\natexlab{}.
\newblock \showarticletitle{Scaling rectified flow transformers for high-resolution image synthesis}. In \bibinfo{booktitle}{\emph{Forty-first international conference on machine learning}}.
\newblock


\bibitem[Fu et~al\mbox{.}(2018)]%
        {fu2018deep}
\bibfield{author}{\bibinfo{person}{Huan Fu}, \bibinfo{person}{Mingming Gong}, \bibinfo{person}{Chaohui Wang}, \bibinfo{person}{Kayhan Batmanghelich}, {and} \bibinfo{person}{Dacheng Tao}.} \bibinfo{year}{2018}\natexlab{}.
\newblock \showarticletitle{Deep ordinal regression network for monocular depth estimation}. In \bibinfo{booktitle}{\emph{Proceedings of the IEEE conference on computer vision and pattern recognition}}. \bibinfo{pages}{2002--2011}.
\newblock


\bibitem[Fu et~al\mbox{.}(2024)]%
        {fu2024geowizard}
\bibfield{author}{\bibinfo{person}{Xiao Fu}, \bibinfo{person}{Wei Yin}, \bibinfo{person}{Mu Hu}, \bibinfo{person}{Kaixuan Wang}, \bibinfo{person}{Yuexin Ma}, \bibinfo{person}{Ping Tan}, \bibinfo{person}{Shaojie Shen}, \bibinfo{person}{Dahua Lin}, {and} \bibinfo{person}{{Xiaoxiao Long. Geowizard: Unleashing the diffusion priors for 3d geometry estimation from a single image}}.} \bibinfo{year}{2024}\natexlab{}.
\newblock In \bibinfo{booktitle}{\emph{European Conference on Computer Vision}}. Springer, \bibinfo{pages}{241--258}.
\newblock


\bibitem[Gabeur et~al\mbox{.}(2026)]%
        {gabeur2026image}
\bibfield{author}{\bibinfo{person}{Valentin Gabeur}, \bibinfo{person}{Shangbang Long}, \bibinfo{person}{Songyou Peng}, \bibinfo{person}{Paul Voigtlaender}, \bibinfo{person}{Shuyang Sun}, \bibinfo{person}{Yanan Bao}, \bibinfo{person}{Karen Truong}, \bibinfo{person}{Zhicheng Wang}, \bibinfo{person}{Wenlei Zhou}, \bibinfo{person}{Jonathan~T Barron}, {et~al\mbox{.}}} \bibinfo{year}{2026}\natexlab{}.
\newblock \showarticletitle{Image Generators are Generalist Vision Learners}.
\newblock \bibinfo{journal}{\emph{arXiv preprint arXiv:2604.20329}} (\bibinfo{year}{2026}).
\newblock


\bibitem[Gaidon et~al\mbox{.}(2016)]%
        {gaidon2016virtual}
\bibfield{author}{\bibinfo{person}{Adrien Gaidon}, \bibinfo{person}{Qiao Wang}, \bibinfo{person}{Yohann Cabon}, {and} \bibinfo{person}{Eleonora Vig}.} \bibinfo{year}{2016}\natexlab{}.
\newblock \showarticletitle{Virtual Worlds as Proxy for Multi-Object Tracking Analysis}. In \bibinfo{booktitle}{\emph{Proceedings of the IEEE Conference on Computer Vision and Pattern Recognition (CVPR)}}.
\newblock


\bibitem[Ganesan et~al\mbox{.}(2026)]%
        {ganesan2026unidac}
\bibfield{author}{\bibinfo{person}{Girish~Chandar Ganesan}, \bibinfo{person}{Yuliang Guo}, \bibinfo{person}{Liu Ren}, {and} \bibinfo{person}{Xiaoming Liu}.} \bibinfo{year}{2026}\natexlab{}.
\newblock \showarticletitle{UniDAC: Universal Metric Depth Estimation for Any Camera}. In \bibinfo{booktitle}{\emph{Proceedings of the IEEE/CVF Conference on Computer Vision and Pattern Recognition}}. \bibinfo{pages}{26953--26963}.
\newblock


\bibitem[Geiger et~al\mbox{.}(2012)]%
        {geiger2012kitti}
\bibfield{author}{\bibinfo{person}{Andreas Geiger}, \bibinfo{person}{Philip Lenz}, {and} \bibinfo{person}{Raquel Urtasun}.} \bibinfo{year}{2012}\natexlab{}.
\newblock \showarticletitle{Are We Ready for Autonomous Driving? The {KITTI} Vision Benchmark Suite}. In \bibinfo{booktitle}{\emph{Proceedings of the IEEE Conference on Computer Vision and Pattern Recognition}}. \bibinfo{pages}{3354--3361}.
\newblock


\bibitem[Godard et~al\mbox{.}(2017)]%
        {monodepth17}
\bibfield{author}{\bibinfo{person}{Cl{\'{e}}ment Godard}, \bibinfo{person}{Oisin {Mac Aodha}}, {and} \bibinfo{person}{Gabriel~J. Brostow}.} \bibinfo{year}{2017}\natexlab{}.
\newblock \showarticletitle{Unsupervised Monocular Depth Estimation with Left-Right Consistency}. In \bibinfo{booktitle}{\emph{CVPR}}.
\newblock


\bibitem[Godard et~al\mbox{.}(2019)]%
        {monodepth2}
\bibfield{author}{\bibinfo{person}{Cl{\'{e}}ment Godard}, \bibinfo{person}{Oisin {Mac Aodha}}, \bibinfo{person}{Michael Firman}, {and} \bibinfo{person}{Gabriel~J. Brostow}.} \bibinfo{year}{2019}\natexlab{}.
\newblock \showarticletitle{Digging into Self-Supervised Monocular Depth Prediction}. In \bibinfo{booktitle}{\emph{The International Conference on Computer Vision (ICCV)}}.
\newblock


\bibitem[Gregorek et~al\mbox{.}(2026)]%
        {marigoldssd}
\bibfield{author}{\bibinfo{person}{Jakub Gregorek}, \bibinfo{person}{Paraskevas Pegios}, \bibinfo{person}{Nando Metzger}, \bibinfo{person}{Konrad Schindler}, \bibinfo{person}{Theodora Kontogianni}, {and} \bibinfo{person}{Lazaros Nalpantidis}.} \bibinfo{year}{2026}\natexlab{}.
\newblock \showarticletitle{Need for Speed: Zero-Shot Depth Completion with Single-Step Diffusion}. In \bibinfo{booktitle}{\emph{Proceedings of the IEEE/CVF Conference on Computer Vision and Pattern Recognition (CVPR) Workshops}}. \bibinfo{pages}{1861--1872}.
\newblock


\bibitem[Gui et~al\mbox{.}(2025)]%
        {gui2025depthfm}
\bibfield{author}{\bibinfo{person}{Ming Gui}, \bibinfo{person}{Johannes Schusterbauer}, \bibinfo{person}{Ulrich Prestel}, \bibinfo{person}{Pingchuan Ma}, \bibinfo{person}{Dmytro Kotovenko}, \bibinfo{person}{Olga Grebenkova}, \bibinfo{person}{Stefan~Andreas Baumann}, \bibinfo{person}{Vincent~Tao Hu}, {and} \bibinfo{person}{Bj{\"o}rn Ommer}.} \bibinfo{year}{2025}\natexlab{}.
\newblock \showarticletitle{DepthFM: Fast Generative Monocular Depth Estimation with Flow Matching}. In \bibinfo{booktitle}{\emph{Proceedings of the AAAI Conference on Artificial Intelligence}}, Vol.~\bibinfo{volume}{39}. \bibinfo{pages}{3203--3211}.
\newblock


\bibitem[Guizilini et~al\mbox{.}(2020)]%
        {guizilini20203dpacking}
\bibfield{author}{\bibinfo{person}{Vitor Guizilini}, \bibinfo{person}{Rares Ambrus}, \bibinfo{person}{Sudeep Pillai}, \bibinfo{person}{Allan Raventos}, {and} \bibinfo{person}{Adrien Gaidon}.} \bibinfo{year}{2020}\natexlab{}.
\newblock \showarticletitle{3D Packing for Self-Supervised Monocular Depth Estimation}. In \bibinfo{booktitle}{\emph{Proceedings of the IEEE/CVF Conference on Computer Vision and Pattern Recognition (CVPR)}}.
\newblock


\bibitem[He et~al\mbox{.}(2025a)]%
        {he2025lotus2}
\bibfield{author}{\bibinfo{person}{Jing He}, \bibinfo{person}{Haodong Li}, \bibinfo{person}{Mingzhi Sheng}, {and} \bibinfo{person}{Ying-Cong Chen}.} \bibinfo{year}{2025}\natexlab{a}.
\newblock \showarticletitle{Lotus-2: Advancing Geometric Dense Prediction with Powerful Image Generative Model}.
\newblock \bibinfo{journal}{\emph{arXiv preprint arXiv:2512.01030}} (\bibinfo{year}{2025}).
\newblock


\bibitem[He et~al\mbox{.}(2025b)]%
        {he2025lotus1}
\bibfield{author}{\bibinfo{person}{Jing He}, \bibinfo{person}{Haodong Li}, \bibinfo{person}{Wei Yin}, \bibinfo{person}{Yixun Liang}, \bibinfo{person}{Leheng Li}, \bibinfo{person}{Kaiqiang Zhou}, \bibinfo{person}{Hongbo Zhang}, \bibinfo{person}{Bingbing Liu}, {and} \bibinfo{person}{YingCong Chen}.} \bibinfo{year}{2025}\natexlab{b}.
\newblock \showarticletitle{Lotus: Diffusion-based visual foundation model for high-quality dense prediction}. In \bibinfo{booktitle}{\emph{International Conference on Learning Representations}}, Vol.~\bibinfo{volume}{2025}. \bibinfo{pages}{89454--89467}.
\newblock


\bibitem[Hu et~al\mbox{.}(2022)]%
        {hu2022lora}
\bibfield{author}{\bibinfo{person}{Edward~J Hu}, \bibinfo{person}{yelong shen}, \bibinfo{person}{Phillip Wallis}, \bibinfo{person}{Zeyuan Allen-Zhu}, \bibinfo{person}{Yuanzhi Li}, \bibinfo{person}{Shean Wang}, \bibinfo{person}{Lu Wang}, {and} \bibinfo{person}{Weizhu Chen}.} \bibinfo{year}{2022}\natexlab{}.
\newblock \showarticletitle{Lo{RA}: Low-Rank Adaptation of Large Language Models}. In \bibinfo{booktitle}{\emph{International Conference on Learning Representations}}.
\newblock
\urldef\tempurl%
\url{https://openreview.net/forum?id=nZeVKeeFYf9}
\showURL{%
\tempurl}


\bibitem[Hu(2024)]%
        {hu2024animate}
\bibfield{author}{\bibinfo{person}{Li Hu}.} \bibinfo{year}{2024}\natexlab{}.
\newblock \showarticletitle{Animate anyone: Consistent and controllable image-to-video synthesis for character animation}. In \bibinfo{booktitle}{\emph{Proceedings of the IEEE/CVF Conference on Computer Vision and Pattern Recognition}}. \bibinfo{pages}{8153--8163}.
\newblock


\bibitem[Huang et~al\mbox{.}(2024)]%
        {huang20242d}
\bibfield{author}{\bibinfo{person}{Binbin Huang}, \bibinfo{person}{Zehao Yu}, \bibinfo{person}{Anpei Chen}, \bibinfo{person}{Andreas Geiger}, {and} \bibinfo{person}{Shenghua Gao}.} \bibinfo{year}{2024}\natexlab{}.
\newblock \showarticletitle{2d gaussian splatting for geometrically accurate radiance fields}. In \bibinfo{booktitle}{\emph{ACM SIGGRAPH 2024 conference papers}}. \bibinfo{pages}{1--11}.
\newblock


\bibitem[Jiang et~al\mbox{.}(2026)]%
        {jiang2026dimer}
\bibfield{author}{\bibinfo{person}{Lutao Jiang}, \bibinfo{person}{Jiantao Lin}, \bibinfo{person}{Kanghao Chen}, \bibinfo{person}{Wenhang Ge}, \bibinfo{person}{Xin Yang}, \bibinfo{person}{Yifan Jiang}, \bibinfo{person}{Yuanhuiyi Lyu}, \bibinfo{person}{Xu Zheng}, \bibinfo{person}{LI JING}, \bibinfo{person}{Yinchuan Li}, {and} \bibinfo{person}{Ying-Cong Chen}.} \bibinfo{year}{2026}\natexlab{}.
\newblock \showarticletitle{DiMeR: Disentangled Mesh Reconstruction Model with Normal-only Geometry Training}. In \bibinfo{booktitle}{\emph{The Fourteenth International Conference on Learning Representations}}.
\newblock
\urldef\tempurl%
\url{https://openreview.net/forum?id=fK2pCgoavb}
\showURL{%
\tempurl}


\bibitem[Ke et~al\mbox{.}(2024)]%
        {ke2024repurposing}
\bibfield{author}{\bibinfo{person}{Bingxin Ke}, \bibinfo{person}{Anton Obukhov}, \bibinfo{person}{Shengyu Huang}, \bibinfo{person}{Nando Metzger}, \bibinfo{person}{Rodrigo~Caye Daudt}, {and} \bibinfo{person}{Konrad Schindler}.} \bibinfo{year}{2024}\natexlab{}.
\newblock \showarticletitle{Repurposing diffusion-based image generators for monocular depth estimation}. In \bibinfo{booktitle}{\emph{Proceedings of the IEEE/CVF conference on computer vision and pattern recognition}}. \bibinfo{pages}{9492--9502}.
\newblock


\bibitem[Ke et~al\mbox{.}(2025)]%
        {ke2025marigold}
\bibfield{author}{\bibinfo{person}{Bingxin Ke}, \bibinfo{person}{Kevin Qu}, \bibinfo{person}{Tianfu Wang}, \bibinfo{person}{Nando Metzger}, \bibinfo{person}{Shengyu Huang}, \bibinfo{person}{Bo Li}, \bibinfo{person}{Anton Obukhov}, {and} \bibinfo{person}{Konrad Schindler}.} \bibinfo{year}{2025}\natexlab{}.
\newblock \showarticletitle{Marigold: Affordable adaptation of diffusion-based image generators for image analysis}.
\newblock \bibinfo{journal}{\emph{IEEE Transactions on Pattern Analysis and Machine Intelligence}} (\bibinfo{year}{2025}).
\newblock


\bibitem[Ke et~al\mbox{.}(2026)]%
        {capa}
\bibfield{author}{\bibinfo{person}{Bingxin Ke}, \bibinfo{person}{Qunjie Zhou}, \bibinfo{person}{Jiahui Huang}, \bibinfo{person}{Xuanchi Ren}, \bibinfo{person}{Tianchang Shen}, \bibinfo{person}{Konrad Schindler}, \bibinfo{person}{Laura Leal-Taixé}, {and} \bibinfo{person}{Shengyu Huang}.} \bibinfo{year}{2026}\natexlab{}.
\newblock \bibinfo{title}{Depth Completion as Parameter-Efficient Test-Time Adaptation}.
\newblock
\showeprint{arXiv:2602.14751}


\bibitem[Koch et~al\mbox{.}(2018)]%
        {koch2018evaluation}
\bibfield{author}{\bibinfo{person}{Tobias Koch}, \bibinfo{person}{Lukas Liebel}, \bibinfo{person}{Friedrich Fraundorfer}, {and} \bibinfo{person}{Marco Korner}.} \bibinfo{year}{2018}\natexlab{}.
\newblock \showarticletitle{Evaluation of CNN-based Single-Image Depth Estimation Methods}. In \bibinfo{booktitle}{\emph{Proceedings of the European Conference on Computer Vision (ECCV) Workshops}}.
\newblock


\bibitem[Krishnan et~al\mbox{.}(2025)]%
        {Krishnan_2025_ICCV}
\bibfield{author}{\bibinfo{person}{Akshay Krishnan}, \bibinfo{person}{Xinchen Yan}, \bibinfo{person}{Vincent Casser}, {and} \bibinfo{person}{Abhijit Kundu}.} \bibinfo{year}{2025}\natexlab{}.
\newblock \showarticletitle{Orchid: Image Latent Diffusion for Joint Appearance and Geometry Generation}. In \bibinfo{booktitle}{\emph{Proceedings of the IEEE/CVF International Conference on Computer Vision (ICCV)}}. \bibinfo{pages}{28217--28227}.
\newblock


\bibitem[Le et~al\mbox{.}(2025)]%
        {Le_2025_CVPR}
\bibfield{author}{\bibinfo{person}{Duong~H. Le}, \bibinfo{person}{Tuan Pham}, \bibinfo{person}{Sangho Lee}, \bibinfo{person}{Christopher Clark}, \bibinfo{person}{Aniruddha Kembhavi}, \bibinfo{person}{Stephan Mandt}, \bibinfo{person}{Ranjay Krishna}, {and} \bibinfo{person}{Jiasen Lu}.} \bibinfo{year}{2025}\natexlab{}.
\newblock \showarticletitle{One Diffusion to Generate Them All}. In \bibinfo{booktitle}{\emph{Proceedings of the IEEE/CVF Conference on Computer Vision and Pattern Recognition (CVPR)}}. \bibinfo{pages}{2671--2682}.
\newblock


\bibitem[Lee et~al\mbox{.}(2019)]%
        {lee2019big}
\bibfield{author}{\bibinfo{person}{Jin~Han Lee}, \bibinfo{person}{Myung-Kyu Han}, \bibinfo{person}{Dong~Wook Ko}, {and} \bibinfo{person}{Il~Hong Suh}.} \bibinfo{year}{2019}\natexlab{}.
\newblock \showarticletitle{From Big to Small: Multi-Scale Local Planar Guidance for Monocular Depth Estimation}.
\newblock \bibinfo{journal}{\emph{arXiv e-prints}} (\bibinfo{year}{2019}), \bibinfo{pages}{arXiv--1907}.
\newblock


\bibitem[Lin et~al\mbox{.}(2026)]%
        {lin2025depth}
\bibfield{author}{\bibinfo{person}{Haotong Lin}, \bibinfo{person}{Sili Chen}, \bibinfo{person}{Jun~Hao Liew}, \bibinfo{person}{Donny~Y. Chen}, \bibinfo{person}{Zhenyu Li}, \bibinfo{person}{Yang Zhao}, \bibinfo{person}{Sida Peng}, \bibinfo{person}{Hengkai Guo}, \bibinfo{person}{Xiaowei Zhou}, \bibinfo{person}{Guang Shi}, \bibinfo{person}{Jiashi Feng}, {and} \bibinfo{person}{Bingyi Kang}.} \bibinfo{year}{2026}\natexlab{}.
\newblock \showarticletitle{Depth Anything 3: Recovering the Visual Space from Any Views}.
\newblock  (\bibinfo{year}{2026}).
\newblock
\urldef\tempurl%
\url{https://openreview.net/forum?id=yirunib8l8}
\showURL{%
\tempurl}


\bibitem[Liu et~al\mbox{.}(2026)]%
        {liu2026opensourceimageeditingmodels}
\bibfield{author}{\bibinfo{person}{Wei Liu}, \bibinfo{person}{Jiaxin Lin}, {and} \bibinfo{person}{Rui Chen}.} \bibinfo{year}{2026}\natexlab{}.
\newblock \bibinfo{title}{Open-Source Image Editing Models Are Zero-Shot Vision Learners}.
\newblock
\showeprint[arxiv]{2605.04566}~[cs.CV]
\urldef\tempurl%
\url{https://arxiv.org/abs/2605.04566}
\showURL{%
\tempurl}


\bibitem[Long et~al\mbox{.}(2024)]%
        {long2024wonder3d}
\bibfield{author}{\bibinfo{person}{Xiaoxiao Long}, \bibinfo{person}{Yuan-Chen Guo}, \bibinfo{person}{Cheng Lin}, \bibinfo{person}{Yuan Liu}, \bibinfo{person}{Zhiyang Dou}, \bibinfo{person}{Lingjie Liu}, \bibinfo{person}{Yuexin Ma}, \bibinfo{person}{Song-Hai Zhang}, \bibinfo{person}{Marc Habermann}, \bibinfo{person}{Christian Theobalt}, {et~al\mbox{.}}} \bibinfo{year}{2024}\natexlab{}.
\newblock \showarticletitle{Wonder3d: Single image to 3d using cross-domain diffusion}. In \bibinfo{booktitle}{\emph{Proceedings of the IEEE/CVF conference on computer vision and pattern recognition}}. \bibinfo{pages}{9970--9980}.
\newblock


\bibitem[Martin~Garcia et~al\mbox{.}(2025)]%
        {garcia2025fine}
\bibfield{author}{\bibinfo{person}{Gonzalo Martin~Garcia}, \bibinfo{person}{Karim Knaebel}, \bibinfo{person}{Christian Schmidt}, \bibinfo{person}{Daan de Geus}, \bibinfo{person}{Alexander Hermans}, {and} \bibinfo{person}{Bastian Leibe}.} \bibinfo{year}{2025}\natexlab{}.
\newblock \showarticletitle{Fine-Tuning Image-Conditional Diffusion Models is Easier than You Think}. In \bibinfo{booktitle}{\emph{Proceedings of the IEEE/CVF Winter Conference on Applications of Computer Vision (WACV)}}.
\newblock


\bibitem[Oquab et~al\mbox{.}(2024)]%
        {oquabdinov2}
\bibfield{author}{\bibinfo{person}{Maxime Oquab}, \bibinfo{person}{Timoth{\'e}e Darcet}, \bibinfo{person}{Th{\'e}o Moutakanni}, \bibinfo{person}{Huy~V Vo}, \bibinfo{person}{Marc Szafraniec}, \bibinfo{person}{Vasil Khalidov}, \bibinfo{person}{Pierre Fernandez}, \bibinfo{person}{Daniel HAZIZA}, \bibinfo{person}{Francisco Massa}, \bibinfo{person}{Alaaeldin El-Nouby}, {et~al\mbox{.}}} \bibinfo{year}{2024}\natexlab{}.
\newblock \showarticletitle{DINOv2: Learning Robust Visual Features without Supervision}.
\newblock \bibinfo{journal}{\emph{Transactions on Machine Learning Research}} (\bibinfo{year}{2024}).
\newblock


\bibitem[Peebles and Xie(2023)]%
        {peebles2023scalable}
\bibfield{author}{\bibinfo{person}{William Peebles} {and} \bibinfo{person}{Saining Xie}.} \bibinfo{year}{2023}\natexlab{}.
\newblock \showarticletitle{Scalable diffusion models with transformers}. In \bibinfo{booktitle}{\emph{Proceedings of the IEEE/CVF international conference on computer vision}}. \bibinfo{pages}{4195--4205}.
\newblock


\bibitem[Peng et~al\mbox{.}(2022)]%
        {Peng2022BokehMe}
\bibfield{author}{\bibinfo{person}{Juewen Peng}, \bibinfo{person}{Zhiguo Cao}, \bibinfo{person}{Xianrui Luo}, \bibinfo{person}{Hao Lu}, \bibinfo{person}{Ke Xian}, {and} \bibinfo{person}{Jianming Zhang}.} \bibinfo{year}{2022}\natexlab{}.
\newblock \showarticletitle{BokehMe: When Neural Rendering Meets Classical Rendering}. In \bibinfo{booktitle}{\emph{Proceedings of the IEEE/CVF International Conference on Computer Vision and Pattern Recognition (CVPR)}}.
\newblock


\bibitem[Pham et~al\mbox{.}(2025)]%
        {pham2025sharpdepth}
\bibfield{author}{\bibinfo{person}{Duc-Hai Pham}, \bibinfo{person}{Tung Do}, \bibinfo{person}{Phong Nguyen}, \bibinfo{person}{Binh-Son Hua}, \bibinfo{person}{Khoi Nguyen}, {and} \bibinfo{person}{Rang Nguyen}.} \bibinfo{year}{2025}\natexlab{}.
\newblock \showarticletitle{Sharpdepth: Sharpening metric depth predictions using diffusion distillation}. In \bibinfo{booktitle}{\emph{Proceedings of the Computer Vision and Pattern Recognition Conference}}. \bibinfo{pages}{17060--17069}.
\newblock


\bibitem[Piccinelli et~al\mbox{.}(2025)]%
        {piccinelli2025unik3d}
\bibfield{author}{\bibinfo{person}{Luigi Piccinelli}, \bibinfo{person}{Christos Sakaridis}, \bibinfo{person}{Mattia Segu}, \bibinfo{person}{Yung-Hsu Yang}, \bibinfo{person}{Siyuan Li}, \bibinfo{person}{Wim Abbeloos}, {and} \bibinfo{person}{Luc Van~Gool}.} \bibinfo{year}{2025}\natexlab{}.
\newblock \showarticletitle{{U}ni{K3D}: Universal Camera Monocular 3D Estimation}. In \bibinfo{booktitle}{\emph{IEEE/CVF Conference on Computer Vision and Pattern Recognition (CVPR)}}.
\newblock


\bibitem[Piccinelli et~al\mbox{.}(2026)]%
        {piccinelli2026velodepth}
\bibfield{author}{\bibinfo{person}{Luigi Piccinelli}, \bibinfo{person}{Thiemo Wandel}, \bibinfo{person}{Christos Sakaridis}, \bibinfo{person}{Wim Abbeloos}, {and} \bibinfo{person}{Luc Van~Gool}.} \bibinfo{year}{2026}\natexlab{}.
\newblock \showarticletitle{Video Depth Propagation}. In \bibinfo{booktitle}{\emph{Proceedings of the International Conference on 3D Vision (3DV)}}.
\newblock


\bibitem[Piccinelli et~al\mbox{.}(2024)]%
        {piccinelli2024unidepth}
\bibfield{author}{\bibinfo{person}{Luigi Piccinelli}, \bibinfo{person}{Yung-Hsu Yang}, \bibinfo{person}{Christos Sakaridis}, \bibinfo{person}{Mattia Segu}, \bibinfo{person}{Siyuan Li}, \bibinfo{person}{Luc Van~Gool}, {and} \bibinfo{person}{Fisher Yu}.} \bibinfo{year}{2024}\natexlab{}.
\newblock \showarticletitle{{U}ni{D}epth: Universal Monocular Metric Depth Estimation}. In \bibinfo{booktitle}{\emph{Proceedings of the IEEE/CVF Conference on Computer Vision and Pattern Recognition (CVPR)}}. \bibinfo{publisher}{IEEE}, \bibinfo{address}{Seattle, WA, USA}.
\newblock


\bibitem[Poggi et~al\mbox{.}(2020)]%
        {Poggi_CVPR_2020}
\bibfield{author}{\bibinfo{person}{Matteo Poggi}, \bibinfo{person}{Filippo Aleotti}, \bibinfo{person}{Fabio Tosi}, {and} \bibinfo{person}{Stefano Mattoccia}.} \bibinfo{year}{2020}\natexlab{}.
\newblock \showarticletitle{On the uncertainty of self-supervised monocular depth estimation}. In \bibinfo{booktitle}{\emph{IEEE/CVF Conference on Computer Vision and Pattern Recognition (CVPR)}}.
\newblock


\bibitem[Ranftl et~al\mbox{.}(2021)]%
        {ranftl2021vision}
\bibfield{author}{\bibinfo{person}{Ren{\'e} Ranftl}, \bibinfo{person}{Alexey Bochkovskiy}, {and} \bibinfo{person}{Vladlen Koltun}.} \bibinfo{year}{2021}\natexlab{}.
\newblock \showarticletitle{Vision transformers for dense prediction}. In \bibinfo{booktitle}{\emph{Proceedings of the IEEE/CVF international conference on computer vision}}. \bibinfo{pages}{12179--12188}.
\newblock


\bibitem[Ranftl et~al\mbox{.}(2020)]%
        {ranftl2020towards}
\bibfield{author}{\bibinfo{person}{Ren{\'e} Ranftl}, \bibinfo{person}{Katrin Lasinger}, \bibinfo{person}{David Hafner}, \bibinfo{person}{Konrad Schindler}, {and} \bibinfo{person}{Vladlen Koltun}.} \bibinfo{year}{2020}\natexlab{}.
\newblock \showarticletitle{Towards robust monocular depth estimation: Mixing datasets for zero-shot cross-dataset transfer}.
\newblock \bibinfo{journal}{\emph{IEEE transactions on pattern analysis and machine intelligence}} \bibinfo{volume}{44}, \bibinfo{number}{3} (\bibinfo{year}{2020}), \bibinfo{pages}{1623--1637}.
\newblock


\bibitem[Roberts et~al\mbox{.}(2021)]%
        {roberts2021hypersim}
\bibfield{author}{\bibinfo{person}{Mike Roberts}, \bibinfo{person}{Jason Ramapuram}, \bibinfo{person}{Anurag Ranjan}, \bibinfo{person}{Atulit Kumar}, \bibinfo{person}{Miguel~Angel Bautista}, \bibinfo{person}{Nathan Paczan}, \bibinfo{person}{Russ Webb}, {and} \bibinfo{person}{Joshua~M. Susskind}.} \bibinfo{year}{2021}\natexlab{}.
\newblock \showarticletitle{Hypersim: A Photorealistic Synthetic Dataset for Holistic Indoor Scene Understanding}. In \bibinfo{booktitle}{\emph{Proceedings of the IEEE/CVF International Conference on Computer Vision (ICCV)}}.
\newblock


\bibitem[Rombach et~al\mbox{.}(2022)]%
        {rombach2022high}
\bibfield{author}{\bibinfo{person}{Robin Rombach}, \bibinfo{person}{Andreas Blattmann}, \bibinfo{person}{Dominik Lorenz}, \bibinfo{person}{Patrick Esser}, {and} \bibinfo{person}{Bj{\"o}rn Ommer}.} \bibinfo{year}{2022}\natexlab{}.
\newblock \showarticletitle{High-resolution image synthesis with latent diffusion models}. In \bibinfo{booktitle}{\emph{Proceedings of the IEEE/CVF conference on computer vision and pattern recognition}}. \bibinfo{pages}{10684--10695}.
\newblock


\bibitem[Ronneberger et~al\mbox{.}(2015)]%
        {ronneberger2015u}
\bibfield{author}{\bibinfo{person}{Olaf Ronneberger}, \bibinfo{person}{Philipp Fischer}, {and} \bibinfo{person}{Thomas Brox}.} \bibinfo{year}{2015}\natexlab{}.
\newblock \showarticletitle{U-net: Convolutional networks for biomedical image segmentation}. In \bibinfo{booktitle}{\emph{International Conference on Medical image computing and computer-assisted intervention}}. Springer, \bibinfo{pages}{234--241}.
\newblock


\bibitem[Safadoust et~al\mbox{.}(2024)]%
        {safadoust2024BMVC}
\bibfield{author}{\bibinfo{person}{Sadra Safadoust}, \bibinfo{person}{Fabio Tosi}, \bibinfo{person}{Fatma G{\"u}ney}, {and} \bibinfo{person}{Matteo Poggi}.} \bibinfo{year}{2024}\natexlab{}.
\newblock \showarticletitle{Self-Evolving Depth-Supervised 3D Gaussian Splatting from Rendered Stereo Pairs}. In \bibinfo{booktitle}{\emph{British Machine Vision Conference (BMVC)}}.
\newblock


\bibitem[Sch{\"o}ps et~al\mbox{.}(2017)]%
        {schoeps2017eth3d}
\bibfield{author}{\bibinfo{person}{Thomas Sch{\"o}ps}, \bibinfo{person}{Johannes~L. Sch{\"o}nberger}, \bibinfo{person}{Silvano Galliani}, \bibinfo{person}{Torsten Sattler}, \bibinfo{person}{Konrad Schindler}, \bibinfo{person}{Marc Pollefeys}, {and} \bibinfo{person}{Andreas Geiger}.} \bibinfo{year}{2017}\natexlab{}.
\newblock \showarticletitle{A Multi-View Stereo Benchmark with High-Resolution Images and Multi-Camera Videos}. In \bibinfo{booktitle}{\emph{Proceedings of the IEEE Conference on Computer Vision and Pattern Recognition}}. \bibinfo{pages}{3260--3269}.
\newblock


\bibitem[Silberman et~al\mbox{.}(2012)]%
        {silberman2012nyu}
\bibfield{author}{\bibinfo{person}{Nathan Silberman}, \bibinfo{person}{Derek Hoiem}, \bibinfo{person}{Pushmeet Kohli}, {and} \bibinfo{person}{Rob Fergus}.} \bibinfo{year}{2012}\natexlab{}.
\newblock \showarticletitle{Indoor Segmentation and Support Inference from {RGBD} Images}. In \bibinfo{booktitle}{\emph{European Conference on Computer Vision}}. \bibinfo{publisher}{Springer}, \bibinfo{pages}{746--760}.
\newblock


\bibitem[Sim{\'e}oni et~al\mbox{.}(2026)]%
        {simeoni2025dinov3}
\bibfield{author}{\bibinfo{person}{Oriane Sim{\'e}oni}, \bibinfo{person}{Huy~V. Vo}, \bibinfo{person}{Maximilian Seitzer}, \bibinfo{person}{Federico Baldassarre}, \bibinfo{person}{Maxime Oquab}, \bibinfo{person}{Cijo Jose}, \bibinfo{person}{Vasil Khalidov}, \bibinfo{person}{Marc Szafraniec}, \bibinfo{person}{Seung~Eun Yi}, \bibinfo{person}{Michael Ramamonjisoa}, \bibinfo{person}{Francisco Massa}, \bibinfo{person}{Daniel HAZIZA}, \bibinfo{person}{Luca Wehrstedt}, \bibinfo{person}{Jianyuan Wang}, \bibinfo{person}{Timoth{\'e}e Darcet}, \bibinfo{person}{Th{\'e}o Moutakanni}, \bibinfo{person}{Leonel Sentana}, \bibinfo{person}{Claire Roberts}, \bibinfo{person}{Andrea Vedaldi}, \bibinfo{person}{Jamie Tolan}, \bibinfo{person}{John Brandt}, \bibinfo{person}{Camille Couprie}, \bibinfo{person}{Julien Mairal}, \bibinfo{person}{Herve Jegou}, \bibinfo{person}{Patrick Labatut}, {and} \bibinfo{person}{Piotr Bojanowski}.} \bibinfo{year}{2026}\natexlab{}.
\newblock \showarticletitle{{DINO}v3}.
\newblock \bibinfo{journal}{\emph{Transactions on Machine Learning Research}} (\bibinfo{year}{2026}).
\newblock
\showISSN{2835-8856}
\urldef\tempurl%
\url{https://openreview.net/forum?id=2NlGyqNjns}
\showURL{%
\tempurl}
\newblock
\shownote{Featured Certification}.


\bibitem[Singh et~al\mbox{.}(2026)]%
        {singh2025irepa}
\bibfield{author}{\bibinfo{person}{Jaskirat Singh}, \bibinfo{person}{Xingjian Leng}, \bibinfo{person}{Zongze Wu}, \bibinfo{person}{Liang Zheng}, \bibinfo{person}{Richard Zhang}, \bibinfo{person}{Eli Shechtman}, {and} \bibinfo{person}{Saining Xie}.} \bibinfo{year}{2026}\natexlab{}.
\newblock \showarticletitle{What matters for Representation Alignment: Global Information or Spatial Structure?}. In \bibinfo{booktitle}{\emph{The Fourteenth International Conference on Learning Representations}}.
\newblock
\urldef\tempurl%
\url{https://openreview.net/forum?id=y0UxFtXqXf}
\showURL{%
\tempurl}


\bibitem[Sinkhorn and Knopp(1967)]%
        {sinkhorn1967concerning}
\bibfield{author}{\bibinfo{person}{Richard Sinkhorn} {and} \bibinfo{person}{Paul Knopp}.} \bibinfo{year}{1967}\natexlab{}.
\newblock \showarticletitle{Concerning nonnegative matrices and doubly stochastic matrices}.
\newblock \bibinfo{journal}{\emph{Pacific J. Math.}} \bibinfo{volume}{21}, \bibinfo{number}{2} (\bibinfo{year}{1967}), \bibinfo{pages}{343--348}.
\newblock


\bibitem[Song et~al\mbox{.}(2026)]%
        {song2026depthmaster}
\bibfield{author}{\bibinfo{person}{Ziyang Song}, \bibinfo{person}{Zerong Wang}, \bibinfo{person}{Bo Li}, \bibinfo{person}{Hao Zhang}, \bibinfo{person}{Ruijie Zhu}, \bibinfo{person}{Li Liu}, \bibinfo{person}{Peng-Tao Jiang}, {and} \bibinfo{person}{Tianzhu Zhang}.} \bibinfo{year}{2026}\natexlab{}.
\newblock \showarticletitle{Depthmaster: Taming diffusion models for monocular depth estimation}.
\newblock \bibinfo{journal}{\emph{IEEE Transactions on Circuits and Systems for Video Technology}} (\bibinfo{year}{2026}).
\newblock


\bibitem[Tosi et~al\mbox{.}(2019)]%
        {tosi2019learning}
\bibfield{author}{\bibinfo{person}{Fabio Tosi}, \bibinfo{person}{Filippo Aleotti}, \bibinfo{person}{Matteo Poggi}, {and} \bibinfo{person}{Stefano Mattoccia}.} \bibinfo{year}{2019}\natexlab{}.
\newblock \showarticletitle{Learning monocular depth estimation infusing traditional stereo knowledge}. In \bibinfo{booktitle}{\emph{Proceedings of the IEEE/CVF conference on computer vision and pattern recognition}}. \bibinfo{pages}{9799--9809}.
\newblock


\bibitem[Tosi et~al\mbox{.}(2021)]%
        {tosi2021smd}
\bibfield{author}{\bibinfo{person}{Fabio Tosi}, \bibinfo{person}{Yiyi Liao}, \bibinfo{person}{Carolin Schmitt}, {and} \bibinfo{person}{Andreas Geiger}.} \bibinfo{year}{2021}\natexlab{}.
\newblock \showarticletitle{Smd-nets: Stereo mixture density networks}. In \bibinfo{booktitle}{\emph{Proceedings of the IEEE/CVF conference on computer vision and pattern recognition}}. \bibinfo{pages}{8942--8952}.
\newblock


\bibitem[Uhrig et~al\mbox{.}(2017)]%
        {uhrig2017sparsity}
\bibfield{author}{\bibinfo{person}{Jonas Uhrig}, \bibinfo{person}{Nick Schneider}, \bibinfo{person}{Lukas Schneider}, \bibinfo{person}{Uwe Franke}, \bibinfo{person}{Thomas Brox}, {and} \bibinfo{person}{Andreas Geiger}.} \bibinfo{year}{2017}\natexlab{}.
\newblock \showarticletitle{Sparsity Invariant CNNs}. In \bibinfo{booktitle}{\emph{IEEE International Conference on 3D Vision (3DV)}}.
\newblock
\urldef\tempurl%
\url{http://lmb.informatik.uni-freiburg.de/Publications/2017/UB17a}
\showURL{%
\tempurl}


\bibitem[Vasiljevic et~al\mbox{.}(2019)]%
        {vasiljevic2019diode}
\bibfield{author}{\bibinfo{person}{Igor Vasiljevic}, \bibinfo{person}{Nicholas~I. Kolkin}, \bibinfo{person}{Shanyi Zhang}, \bibinfo{person}{Ruotian Luo}, \bibinfo{person}{Haochen Wang}, \bibinfo{person}{Falcon~Z. Dai}, \bibinfo{person}{Andrea~F. Daniele}, \bibinfo{person}{Mohammadreza Mostajabi}, \bibinfo{person}{Steven Basart}, \bibinfo{person}{Matthew~R. Walter}, {and} \bibinfo{person}{Gregory Shakhnarovich}.} \bibinfo{year}{2019}\natexlab{}.
\newblock \showarticletitle{{DIODE}: A Dense Indoor and Outdoor {DEpth} Dataset}.
\newblock \bibinfo{journal}{\emph{arXiv preprint arXiv:1908.00463}} (\bibinfo{year}{2019}).
\newblock


\bibitem[Viola et~al\mbox{.}(2025)]%
        {marigolddc}
\bibfield{author}{\bibinfo{person}{Massimiliano Viola}, \bibinfo{person}{Kevin Qu}, \bibinfo{person}{Nando Metzger}, \bibinfo{person}{Bingxin Ke}, \bibinfo{person}{Alexander Becker}, \bibinfo{person}{Konrad Schindler}, {and} \bibinfo{person}{Anton Obukhov}.} \bibinfo{year}{2025}\natexlab{}.
\newblock \showarticletitle{Marigold-DC: Zero-Shot Monocular Depth Completion with Guided Diffusion}. In \bibinfo{booktitle}{\emph{Proceedings of the IEEE/CVF International Conference on Computer Vision (ICCV)}}.
\newblock


\bibitem[Wang et~al\mbox{.}(2026a)]%
        {wang2026fe2e}
\bibfield{author}{\bibinfo{person}{Jiyuan Wang}, \bibinfo{person}{Chunyu Lin}, \bibinfo{person}{Lei Sun}, \bibinfo{person}{Rongying Liu}, \bibinfo{person}{Lang Nie}, \bibinfo{person}{Mingxing Li}, \bibinfo{person}{Kang Liao}, {and} \bibinfo{person}{Xiangxiang Chu}.} \bibinfo{year}{2026}\natexlab{a}.
\newblock \showarticletitle{From Editor to Dense Geometry Estimator}. In \bibinfo{booktitle}{\emph{Proceedings of the IEEE/CVF Conference on Computer Vision and Pattern Recognition}}.
\newblock


\bibitem[Wang et~al\mbox{.}(2025a)]%
        {wang2025moge}
\bibfield{author}{\bibinfo{person}{Ruicheng Wang}, \bibinfo{person}{Sicheng Xu}, \bibinfo{person}{Cassie Dai}, \bibinfo{person}{Jianfeng Xiang}, \bibinfo{person}{Yu Deng}, \bibinfo{person}{Xin Tong}, {and} \bibinfo{person}{Jiaolong Yang}.} \bibinfo{year}{2025}\natexlab{a}.
\newblock \showarticletitle{Moge: Unlocking accurate monocular geometry estimation for open-domain images with optimal training supervision}. In \bibinfo{booktitle}{\emph{Proceedings of the Computer Vision and Pattern Recognition Conference}}. \bibinfo{pages}{5261--5271}.
\newblock


\bibitem[Wang et~al\mbox{.}(2025b)]%
        {wang2025moge2}
\bibfield{author}{\bibinfo{person}{Ruicheng Wang}, \bibinfo{person}{Sicheng Xu}, \bibinfo{person}{Yue Dong}, \bibinfo{person}{Yu Deng}, \bibinfo{person}{Jianfeng Xiang}, \bibinfo{person}{Zelong Lv}, \bibinfo{person}{Guangzhong Sun}, \bibinfo{person}{Xin Tong}, {and} \bibinfo{person}{Jiaolong Yang}.} \bibinfo{year}{2025}\natexlab{b}.
\newblock \showarticletitle{MoGe-2: Accurate Monocular Geometry with Metric Scale and Sharp Details}. In \bibinfo{booktitle}{\emph{The Thirty-ninth Annual Conference on Neural Information Processing Systems}}.
\newblock
\urldef\tempurl%
\url{https://openreview.net/forum?id=16mDq7m2OK}
\showURL{%
\tempurl}


\bibitem[Wang et~al\mbox{.}(2026b)]%
        {wang2025pi}
\bibfield{author}{\bibinfo{person}{Yifan Wang}, \bibinfo{person}{Jianjun Zhou}, \bibinfo{person}{Haoyi Zhu}, \bibinfo{person}{Wenzheng Chang}, \bibinfo{person}{Yang Zhou}, \bibinfo{person}{Zizun Li}, \bibinfo{person}{Junyi Chen}, \bibinfo{person}{Jiangmiao Pang}, \bibinfo{person}{Chunhua Shen}, {and} \bibinfo{person}{Tong He}.} \bibinfo{year}{2026}\natexlab{b}.
\newblock \showarticletitle{{$\pi^3$}: Scalable Permutation-Equivariant Visual Geometry Learning}. In \bibinfo{booktitle}{\emph{International Conference on Learning Representations (ICLR)}}.
\newblock


\bibitem[Wen et~al\mbox{.}(2025)]%
        {wen2025seeing}
\bibfield{author}{\bibinfo{person}{Hongyu Wen}, \bibinfo{person}{Yiming Zuo}, \bibinfo{person}{Venkat Subramanian}, \bibinfo{person}{Patrick Chen}, {and} \bibinfo{person}{Jia Deng}.} \bibinfo{year}{2025}\natexlab{}.
\newblock \showarticletitle{Seeing and Seeing Through the Glass: Real and Synthetic Data for Multi-Layer Depth Estimation}. In \bibinfo{booktitle}{\emph{Proceedings of the IEEE/CVF International Conference on Computer Vision (ICCV)}}. \bibinfo{pages}{6715--6725}.
\newblock


\bibitem[Wu et~al\mbox{.}(2025)]%
        {wu2025qwen}
\bibfield{author}{\bibinfo{person}{Chenfei Wu}, \bibinfo{person}{Jiahao Li}, \bibinfo{person}{Jingren Zhou}, \bibinfo{person}{Junyang Lin}, \bibinfo{person}{Kaiyuan Gao}, \bibinfo{person}{Kun Yan}, \bibinfo{person}{Sheng-ming Yin}, \bibinfo{person}{Shuai Bai}, \bibinfo{person}{Xiao Xu}, \bibinfo{person}{Yilei Chen}, {et~al\mbox{.}}} \bibinfo{year}{2025}\natexlab{}.
\newblock \showarticletitle{Qwen-image technical report}.
\newblock \bibinfo{journal}{\emph{arXiv preprint arXiv:2508.02324}} (\bibinfo{year}{2025}).
\newblock


\bibitem[Xu et~al\mbox{.}(2025a)]%
        {xu2026pixel}
\bibfield{author}{\bibinfo{person}{Gangwei Xu}, \bibinfo{person}{Haotong Lin}, \bibinfo{person}{Hongcheng Luo}, \bibinfo{person}{Xianqi Wang}, \bibinfo{person}{Jingfeng Yao}, \bibinfo{person}{Lianghui Zhu}, \bibinfo{person}{Yuechuan Pu}, \bibinfo{person}{Cheng Chi}, \bibinfo{person}{Haiyang Sun}, \bibinfo{person}{Bing Wang}, \bibinfo{person}{Guang Chen}, \bibinfo{person}{Hangjun Ye}, \bibinfo{person}{Sida Peng}, {and} \bibinfo{person}{Xin Yang}.} \bibinfo{year}{2025}\natexlab{a}.
\newblock \showarticletitle{Pixel-perfect depth with semantics-prompted diffusion transformers}.
\newblock \bibinfo{journal}{\emph{Advances in Neural Information Processing Systems}}  \bibinfo{volume}{38} (\bibinfo{year}{2025}), \bibinfo{pages}{174731--174755}.
\newblock


\bibitem[Xu et~al\mbox{.}(2025b)]%
        {xu2025matters}
\bibfield{author}{\bibinfo{person}{Guangkai Xu}, \bibinfo{person}{Mingyu Liu}, \bibinfo{person}{Chengxiang Fan}, \bibinfo{person}{Kangyang Xie}, \bibinfo{person}{Zhiyue Zhao}, \bibinfo{person}{Hao Chen}, \bibinfo{person}{Chunhua Shen}, {et~al\mbox{.}}} \bibinfo{year}{2025}\natexlab{b}.
\newblock \showarticletitle{What matters when repurposing diffusion models for general dense perception tasks?}. In \bibinfo{booktitle}{\emph{International Conference on Learning Representations}}, Vol.~\bibinfo{volume}{2025}. \bibinfo{pages}{6786--6799}.
\newblock


\bibitem[Yang et~al\mbox{.}(2021)]%
        {yang2021S3Net}
\bibfield{author}{\bibinfo{person}{Hao-Hsiang Yang}, \bibinfo{person}{Wei-Ting Chen}, {and} \bibinfo{person}{Sy-Yen Kuo}.} \bibinfo{year}{2021}\natexlab{}.
\newblock \showarticletitle{{S3N}et: A Single Stream Structure for Depth Guided Image Relighting}. In \bibinfo{booktitle}{\emph{Proceedings of the IEEE Conference on Computer Vision and Pattern Recognition Workshops (CVPRW)}}.
\newblock


\bibitem[Yang et~al\mbox{.}(2024a)]%
        {yang2024depth}
\bibfield{author}{\bibinfo{person}{Lihe Yang}, \bibinfo{person}{Bingyi Kang}, \bibinfo{person}{Zilong Huang}, \bibinfo{person}{Xiaogang Xu}, \bibinfo{person}{Jiashi Feng}, {and} \bibinfo{person}{Hengshuang Zhao}.} \bibinfo{year}{2024}\natexlab{a}.
\newblock \showarticletitle{Depth anything: Unleashing the power of large-scale unlabeled data}. In \bibinfo{booktitle}{\emph{Proceedings of the IEEE/CVF conference on computer vision and pattern recognition}}. \bibinfo{pages}{10371--10381}.
\newblock


\bibitem[Yang et~al\mbox{.}(2024b)]%
        {yang2024depth2}
\bibfield{author}{\bibinfo{person}{Lihe Yang}, \bibinfo{person}{Bingyi Kang}, \bibinfo{person}{Zilong Huang}, \bibinfo{person}{Zhen Zhao}, \bibinfo{person}{Xiaogang Xu}, \bibinfo{person}{Jiashi Feng}, {and} \bibinfo{person}{Hengshuang Zhao}.} \bibinfo{year}{2024}\natexlab{b}.
\newblock \showarticletitle{Depth anything v2}.
\newblock \bibinfo{journal}{\emph{Advances in Neural Information Processing Systems}}  \bibinfo{volume}{37} (\bibinfo{year}{2024}), \bibinfo{pages}{21875--21911}.
\newblock


\bibitem[Yang et~al\mbox{.}(2022)]%
        {yang2022monocular}
\bibfield{author}{\bibinfo{person}{Xin Yang}, \bibinfo{person}{Qingling Chang}, \bibinfo{person}{Xinlin Liu}, {and} \bibinfo{person}{Yan Cui}.} \bibinfo{year}{2022}\natexlab{}.
\newblock \showarticletitle{Monocular depth estimation with sharp boundary}. In \bibinfo{booktitle}{\emph{2022 8th International Conference on Virtual Reality (ICVR)}}. IEEE, \bibinfo{pages}{384--391}.
\newblock


\bibitem[Ye et~al\mbox{.}(2024)]%
        {ye2024stablenormal}
\bibfield{author}{\bibinfo{person}{Chongjie Ye}, \bibinfo{person}{Lingteng Qiu}, \bibinfo{person}{Xiaodong Gu}, \bibinfo{person}{Qi Zuo}, \bibinfo{person}{Yushuang Wu}, \bibinfo{person}{Zilong Dong}, \bibinfo{person}{Liefeng Bo}, \bibinfo{person}{Yuliang Xiu}, {and} \bibinfo{person}{Xiaoguang Han}.} \bibinfo{year}{2024}\natexlab{}.
\newblock \showarticletitle{{StableNormal}: Reducing Diffusion Variance for Stable and Sharp Normal}.
\newblock \bibinfo{journal}{\emph{ACM Transactions on Graphics}} (\bibinfo{year}{2024}).
\newblock


\bibitem[Yin et~al\mbox{.}(2020)]%
        {yin2020diversedepth}
\bibfield{author}{\bibinfo{person}{Wei Yin}, \bibinfo{person}{Xinlong Wang}, \bibinfo{person}{Chunhua Shen}, \bibinfo{person}{Yifan Liu}, \bibinfo{person}{Zhi Tian}, \bibinfo{person}{Songcen Xu}, \bibinfo{person}{Changming Sun}, {and} \bibinfo{person}{Dou Renyin}.} \bibinfo{year}{2020}\natexlab{}.
\newblock \showarticletitle{DiverseDepth: Affine-invariant Depth Prediction Using Diverse Data}.
\newblock \bibinfo{journal}{\emph{arXiv preprint arXiv:2002.00569}} (\bibinfo{year}{2020}).
\newblock


\bibitem[Yin et~al\mbox{.}(2021)]%
        {yin2021learning}
\bibfield{author}{\bibinfo{person}{Wei Yin}, \bibinfo{person}{Jianming Zhang}, \bibinfo{person}{Oliver Wang}, \bibinfo{person}{Simon Niklaus}, \bibinfo{person}{Long Mai}, \bibinfo{person}{Simon Chen}, {and} \bibinfo{person}{Chunhua Shen}.} \bibinfo{year}{2021}\natexlab{}.
\newblock \showarticletitle{Learning to Recover 3D Scene Shape from a Single Image}. In \bibinfo{booktitle}{\emph{Proceedings of the IEEE/CVF Conference on Computer Vision and Pattern Recognition}}.
\newblock


\bibitem[Yu et~al\mbox{.}(2026a)]%
        {yu2026infinidepth}
\bibfield{author}{\bibinfo{person}{Hao Yu}, \bibinfo{person}{Haotong Lin}, \bibinfo{person}{Jiawei Wang}, \bibinfo{person}{Jiaxin Li}, \bibinfo{person}{Yida Wang}, \bibinfo{person}{Xueyang Zhang}, \bibinfo{person}{Yue Wang}, \bibinfo{person}{Xiaowei Zhou}, \bibinfo{person}{Ruizhen Hu}, {and} \bibinfo{person}{Sida Peng}.} \bibinfo{year}{2026}\natexlab{a}.
\newblock \showarticletitle{InfiniDepth: Arbitrary-Resolution and Fine-Grained Depth Estimation with Neural Implicit Fields}. In \bibinfo{booktitle}{\emph{Proceedings of the IEEE/CVF Conference on Computer Vision and Pattern Recognition}}.
\newblock


\bibitem[Yu et~al\mbox{.}(2025)]%
        {repa}
\bibfield{author}{\bibinfo{person}{Sihyun Yu}, \bibinfo{person}{Sangkyung Kwak}, \bibinfo{person}{Huiwon Jang}, \bibinfo{person}{Jongheon Jeong}, \bibinfo{person}{Jonathan Huang}, \bibinfo{person}{Jinwoo Shin}, {and} \bibinfo{person}{Saining Xie}.} \bibinfo{year}{2025}\natexlab{}.
\newblock \showarticletitle{Representation Alignment for Generation: Training Diffusion Transformers Is Easier Than You Think}. In \bibinfo{booktitle}{\emph{International Conference on Learning Representations}}.
\newblock


\bibitem[Yu et~al\mbox{.}(2026b)]%
        {LDCM}
\bibfield{author}{\bibinfo{person}{Zhu Yu}, \bibinfo{person}{Zhengyi Zhao}, \bibinfo{person}{Runmin Zhang}, \bibinfo{person}{Lingteng Qiu}, \bibinfo{person}{Kejie Qiu}, \bibinfo{person}{Yisheng He}, \bibinfo{person}{Siyu Zhu}, \bibinfo{person}{Zilong Dong}, \bibinfo{person}{Si-Yuan Cao}, {and} \bibinfo{person}{Hui-Liang Shen}.} \bibinfo{year}{2026}\natexlab{b}.
\newblock \showarticletitle{Large Depth Completion Model from Sparse Observations}. In \bibinfo{booktitle}{\emph{ICLR}}.
\newblock


\bibitem[Yuan et~al\mbox{.}(2022)]%
        {yuan2022neural}
\bibfield{author}{\bibinfo{person}{Weihao Yuan}, \bibinfo{person}{Xiaodong Gu}, \bibinfo{person}{Zuozhuo Dai}, \bibinfo{person}{Siyu Zhu}, {and} \bibinfo{person}{Ping Tan}.} \bibinfo{year}{2022}\natexlab{}.
\newblock \showarticletitle{Neural window fully-connected crfs for monocular depth estimation}. In \bibinfo{booktitle}{\emph{Proceedings of the IEEE/CVF conference on computer vision and pattern recognition}}. \bibinfo{pages}{3916--3925}.
\newblock


\bibitem[Zakarin et~al\mbox{.}(2026)]%
        {Zakarin_2026_CVPR}
\bibfield{author}{\bibinfo{person}{Daniyar Zakarin}, \bibinfo{person}{Thiemo Wandel}, \bibinfo{person}{Anton Obukhov}, {and} \bibinfo{person}{Dengxin Dai}.} \bibinfo{year}{2026}\natexlab{}.
\newblock \showarticletitle{Reflection Removal through Efficient Adaptation of Diffusion Transformers}. In \bibinfo{booktitle}{\emph{Proceedings of the IEEE/CVF Conference on Computer Vision and Pattern Recognition (CVPR) Workshops}}. \bibinfo{pages}{2776--2785}.
\newblock


\bibitem[Zeng et~al\mbox{.}(2024)]%
        {zeng2024rgb}
\bibfield{author}{\bibinfo{person}{Zheng Zeng}, \bibinfo{person}{Valentin Deschaintre}, \bibinfo{person}{Iliyan Georgiev}, \bibinfo{person}{Yannick Hold-Geoffroy}, \bibinfo{person}{Yiwei Hu}, \bibinfo{person}{Fujun Luan}, \bibinfo{person}{Ling-Qi Yan}, {and} \bibinfo{person}{Miloš Hašan}.} \bibinfo{year}{2024}\natexlab{}.
\newblock \showarticletitle{RGB↔X: Image decomposition and synthesis using material- and lighting-aware diffusion models}. In \bibinfo{booktitle}{\emph{Special Interest Group on Computer Graphics and Interactive Techniques Conference Conference Papers}} \emph{(\bibinfo{series}{SIGGRAPH ’24})}. \bibinfo{publisher}{ACM}, \bibinfo{pages}{1–11}.
\newblock
\href{https://doi.org/10.1145/3641519.3657445}{doi:\nolinkurl{10.1145/3641519.3657445}}


\bibitem[Zhang et~al\mbox{.}(2022)]%
        {zhang2022hierarchical}
\bibfield{author}{\bibinfo{person}{Chi Zhang}, \bibinfo{person}{Wei Yin}, \bibinfo{person}{Zhibin Wang}, \bibinfo{person}{Gang Yu}, \bibinfo{person}{Bin Fu}, {and} \bibinfo{person}{Chunhua Shen}.} \bibinfo{year}{2022}\natexlab{}.
\newblock \showarticletitle{Hierarchical Normalization for Robust Monocular Depth Estimation}. In \bibinfo{booktitle}{\emph{Advances in Neural Information Processing Systems}}, Vol.~\bibinfo{volume}{35}.
\newblock


\bibitem[Zhang et~al\mbox{.}(2023)]%
        {zhang2023adding}
\bibfield{author}{\bibinfo{person}{Lvmin Zhang}, \bibinfo{person}{Anyi Rao}, {and} \bibinfo{person}{Maneesh Agrawala}.} \bibinfo{year}{2023}\natexlab{}.
\newblock \showarticletitle{Adding conditional control to text-to-image diffusion models}. In \bibinfo{booktitle}{\emph{Proceedings of the IEEE/CVF international conference on computer vision}}. \bibinfo{pages}{3836--3847}.
\newblock


\bibitem[Zhang et~al\mbox{.}(2018)]%
        {zhang2018unreasonable}
\bibfield{author}{\bibinfo{person}{Richard Zhang}, \bibinfo{person}{Phillip Isola}, \bibinfo{person}{Alexei~A Efros}, \bibinfo{person}{Eli Shechtman}, {and} \bibinfo{person}{Oliver Wang}.} \bibinfo{year}{2018}\natexlab{}.
\newblock \showarticletitle{The unreasonable effectiveness of deep features as a perceptual metric}. In \bibinfo{booktitle}{\emph{2018 IEEE/CVF conference on computer vision and pattern recognition}}. IEEE, \bibinfo{pages}{586--595}.
\newblock


\bibitem[Zhang et~al\mbox{.}(2024)]%
        {zhang2024betterdepth}
\bibfield{author}{\bibinfo{person}{Xiang Zhang}, \bibinfo{person}{Bingxin Ke}, \bibinfo{person}{Hayko Riemenschneider}, \bibinfo{person}{Nando Metzger}, \bibinfo{person}{Anton Obukhov}, \bibinfo{person}{Markus Gross}, \bibinfo{person}{Konrad Schindler}, {and} \bibinfo{person}{Christopher Schroers}.} \bibinfo{year}{2024}\natexlab{}.
\newblock \showarticletitle{BetterDepth: Plug-and-Play Diffusion Refiner for Zero-Shot Monocular Depth Estimation}. In \bibinfo{booktitle}{\emph{The Thirty-eighth Annual Conference on Neural Information Processing Systems}}.
\newblock
\urldef\tempurl%
\url{https://openreview.net/forum?id=35WwZhkush}
\showURL{%
\tempurl}


\bibitem[Zhao et~al\mbox{.}(2023)]%
        {zhao2023gasmono}
\bibfield{author}{\bibinfo{person}{Chaoqiang Zhao}, \bibinfo{person}{Matteo Poggi}, \bibinfo{person}{Fabio Tosi}, \bibinfo{person}{Lei Zhou}, \bibinfo{person}{Qiyu Sun}, \bibinfo{person}{Yang Tang}, {and} \bibinfo{person}{Stefano Mattoccia}.} \bibinfo{year}{2023}\natexlab{}.
\newblock \showarticletitle{Gasmono: Geometry-aided self-supervised monocular depth estimation for indoor scenes}. In \bibinfo{booktitle}{\emph{Proceedings of the IEEE/CVF international conference on computer vision}}. \bibinfo{pages}{16209--16220}.
\newblock


\bibitem[Zhao et~al\mbox{.}(2025)]%
        {zhaodiception}
\bibfield{author}{\bibinfo{person}{Canyu Zhao}, \bibinfo{person}{Yanlong Sun}, \bibinfo{person}{Mingyu Liu}, \bibinfo{person}{Huanyi Zheng}, \bibinfo{person}{Muzhi Zhu}, \bibinfo{person}{Zhiyue Zhao}, \bibinfo{person}{Hao Chen}, \bibinfo{person}{Tong He}, {and} \bibinfo{person}{Chunhua Shen}.} \bibinfo{year}{2025}\natexlab{}.
\newblock \showarticletitle{DICEPTION: A Generalist Diffusion Model for Visual Perceptual Tasks}. In \bibinfo{booktitle}{\emph{The Thirty-ninth Annual Conference on Neural Information Processing Systems}}.
\newblock


\bibitem[Zhao et~al\mbox{.}(2022)]%
        {zhao2022monovit}
\bibfield{author}{\bibinfo{person}{Chaoqiang Zhao}, \bibinfo{person}{Youmin Zhang}, \bibinfo{person}{Matteo Poggi}, \bibinfo{person}{Fabio Tosi}, \bibinfo{person}{Xianda Guo}, \bibinfo{person}{Zheng Zhu}, \bibinfo{person}{Guan Huang}, \bibinfo{person}{Yang Tang}, {and} \bibinfo{person}{Stefano Mattoccia}.} \bibinfo{year}{2022}\natexlab{}.
\newblock \showarticletitle{Monovit: Self-supervised monocular depth estimation with a vision transformer}. In \bibinfo{booktitle}{\emph{2022 international conference on 3D vision (3DV)}}. IEEE, \bibinfo{pages}{668--678}.
\newblock


\end{thebibliography}

\end{document}